\documentclass[runningheads]{llncs}
\usepackage{graphicx}
\usepackage{comment}
\usepackage{amsmath,amssymb} 
\usepackage{color}
\usepackage{epsfig}
\usepackage{microtype}
\usepackage{float}
\usepackage{bm}
\usepackage{epstopdf}
\usepackage{enumitem}
\usepackage{comment}

\usepackage{multirow}
\usepackage[table,xcdraw]{xcolor}
\usepackage{multirow}
\usepackage[table,xcdraw]{xcolor}
\usepackage[normalem]{ulem}
\useunder{\uline}{\ul}{}
\usepackage{mathtools}
\usepackage{wrapfig}
\usepackage{hyperref}
\hypersetup{
    colorlinks=true,
    linkcolor=blue,
    filecolor=magenta,      
    urlcolor=blue,
}

\def\eg{\emph{e.g.}} 
\def\ie{\emph{i.e.}}

\def\etal{\emph{et al.}}

\usepackage[width=122mm,left=12mm,paperwidth=146mm,height=193mm,top=12mm,paperheight=217mm]{geometry}
\begin{document}
\pagestyle{headings}
\mainmatter

\title{Pseudo RGB-D for Self-Improving Monocular SLAM and Depth Prediction} 

\titlerunning{Pseudo RGB-D for Self-Improving Monocular SLAM and Depth Prediction}

\author{Lokender Tiwari\inst{1} \and
Pan Ji\inst{2} \and
Quoc-Huy Tran\inst{2} \and
Bingbing Zhuang\inst{2} \and
Saket Anand\inst{1} \and
Manmohan Chandraker\inst{2,3} }

\authorrunning{L. Tiwari et al.}
\institute{$^{1}$IIIT-Delhi ~~~ $^{2}$NEC Laboratories America, Inc.~~~$^{3}$University of California, San Diego}

\maketitle


\begin{abstract}
Classical monocular Simultaneous Localization And Mapping (SLAM) and the recently emerging convolutional neural networks (CNNs) for monocular depth prediction represent two largely disjoint approaches towards building a 3D map of the surrounding environment. In this paper, we demonstrate that the coupling of these two by leveraging the strengths of each mitigates the other’s shortcomings. Specifically, we propose a joint narrow and wide baseline based self-improving framework, where on the one hand the CNN-predicted depth is leveraged to perform \emph{pseudo RGB-D} feature-based SLAM, leading to better accuracy and robustness than the monocular RGB SLAM baseline. On the other hand, the bundle-adjusted 3D scene structures and camera poses from the more principled geometric SLAM are injected back into the depth network through novel wide baseline losses proposed for improving the depth prediction network, which then continues to contribute towards better pose and 3D structure estimation in the next iteration. We emphasize that our framework only requires \textit{ unlabeled monocular} videos in both training and inference stages, and yet is able to outperform state-of-the-art \textit{self-supervised monocular} and \textit{stereo} depth prediction networks (\eg,~Monodepth2) and feature-based monocular SLAM system (\ie,~ORB-SLAM). Extensive experiments on KITTI and TUM RGB-D datasets verify the superiority of our self-improving geometry-CNN framework.
\keywords{self-supervised learning, self-improving, single-view depth prediction, monocular SLAM }
\end{abstract}

\section{Introduction}
\label{sec:introduction}
One of the most reliable cues towards 3D perception from a monocular camera arises from camera motion that induces multiple-view geometric constraints~\cite{hartley2003multiple} wherein the 3D scene structure is encoded. Over the years, Simultaneous Localization And Mapping (SLAM)~\cite{davison2007monoslam,klein2007parallel,newcombe2011dtam} has been long studied to simultaneously recover the 3D scene structure of the surrounding and estimate the ego-motion of the agent. With the advent of Convolutional Neural Networks (CNNs), unsupervised learning of single-view depth estimation~\cite{garg2016unsupervised,godard2017unsupervised,zhou2017unsupervised} has emerged as a promising alternative to the traditional geometric approaches. Such methods rely on CNNs to extract meaningful depth cues (\eg, shading, texture, and semantics) from a single image, yielding very promising results. 

Despite the general maturity of monocular geometric SLAM~\cite{engel2014lsd,mur2015orb,engel2017direct} and the rapid advances in unsupervised monocular depth prediction approaches~\cite{mahjourian2018unsupervised} \cite{wang2018learning} \cite{yin2018geonet} \cite{bian2019unsupervised} \cite{godard2019digging}     \cite{Sheng_2019_ICCV}, they both still have their own limitations. \\
\textbf{Monocular SLAM.}  Traditional monocular SLAM has well-known limitations in robustness and accuracy as compared to those leveraging active depth sensors, \eg, RGB-D SLAM~\cite{mur2017orb}. This performance issue is due to the inherent scale ambiguity of depth recovery from monocular cameras, which causes the so-called scale drift in both the camera trajectory and 3D scene depth, and thus lowers robustness and accuracy of conventional monocular SLAM. In addition, the triangulation-based depth estimation employed by traditional SLAM methods is degenerate under pure rotational camera motion~\cite{hartley2003multiple}. \\ 
\noindent \textbf{Unsupervised Monocular Depth Prediction.} Most of the unsupervised and self-supervised methods~\cite{zhou2017unsupervised,godard2017unsupervised,godard2019digging,bian2019unsupervised} formulate single image depth estimation as a novel-view synthesis problem, with appearance based photometric losses being central to the training strategy. Usually, these models train two networks, one each for \emph{pose} and \emph{depth}. As photometric losses largely rely on the brightness consistency assumption, nearly all existing self-supervised approaches operate in a narrow-baseline setting optimizing the loss over a snippet of 2-5 consecutive frames. Consequently, models like MondoDepth2~\cite{godard2019digging}, work very well for close range points, but generate inaccurate depth estimates for points that are farther away (\eg,~see $0^{th}$ iteration in Fig. \ref{fig:self-loop-analysis}). While it is well known that a wide-baseline yields better depth estimates for points at larger depth, a straightforward extension of existing CNN based approaches is inadequate for the following two reasons. A wide baseline in a video sequence implies a larger temporal window, which in most practical scenarios will violate the brightness consistency assumption, rendering the photometric loss ineffective. Secondly, larger temporal windows (wider baselines) would also imply more occluded regions that behave as outliers. Unless these aspects are effectively handled, training of CNN based depth and pose networks in the wide baseline setting will lead to inaccuracies and biases. 

In view of the limitations in both monocular geometric SLAM and unsupervised monocular depth estimation approaches, a particularly interesting question to ask is whether these two approaches can complement each other (see Sec. \ref{sec:self-loop-analysis}) and mitigate the issues discussed above. Our work makes contributions towards answering this question. Specifically, we propose a \emph{self-supervised, self-improving} framework of these two tasks, which is shown to improve the robustness and accuracy on each of them. 

While the performance gap between geometric SLAM and self-supervised learning-based SLAM methods is still large, incorporating depth information drastically improves the robustness of geometric SLAM methods (\eg, see RGB-D SLAM vs. RGB SLAM on the KITTI Odometry leaderboard~\cite{geiger2012we}). Inspired by this success of RGB-D SLAM, we postulate the use of an unsupervised CNN-based depth estimation model as a \emph{pseudo depth sensor}, which allows us to design our self-supervised approach, pseudo RGB-D SLAM (pRGBD-SLAM) that only uses monocular cameras and yet achieves significant improvements in robustness and accuracy as compared to RGB SLAM.

\begin{figure}[H]
  \begin{center}
    	\includegraphics[width=0.45\linewidth]{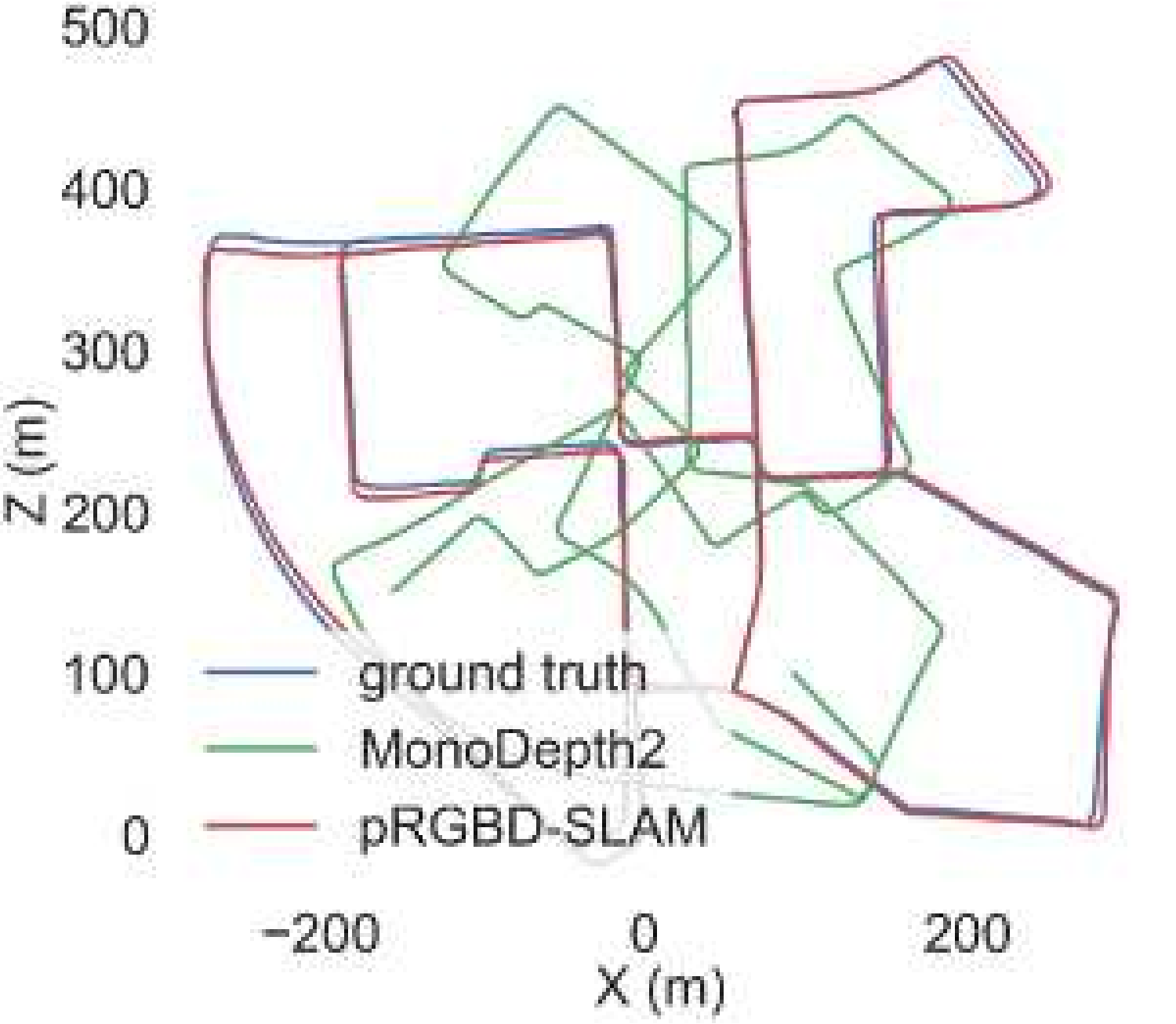}
  \end{center}
  \vspace{-0.5cm}
    	\caption{MonoDepth2~\cite{godard2019digging} pose network camera poses vs. pseudo RGBD-SLAM camera poses where depth(D) is from CNN. The pose network from \cite{godard2019digging} leads to significant drift.}
  \label{fig:pose_md2_vs_prgbd}
  \vspace{-0.5cm}
\end{figure}
Our fusion of geometric SLAM and CNN-based monocular depth estimation turns out to be symbiotic and this complementary nature sets the basis of our self-improving framework. To improve the depth predictions, we make use of two main modifications in the training strategy. First, we eschew the learning based pose estimates in favor of geometric SLAM based estimates (an illustrative motivation is shown in Fig. \ref{fig:pose_md2_vs_prgbd}). Second, we make use of common tracked keypoints from neighboring \emph{keyframes} and impose a symmetric depth transfer and a depth consistency loss on the CNN model. These adaptations are based on the observation that both pose estimates and sparse 3D feature point estimates from geometric SLAM are robust, as most techniques typically apply multiple bundle adjustment iterations over wide baseline depth estimates of common keypoints. This simple observation and the subsequent modification is key to our self-improving framework, which can leverage any unsupervised CNN-based depth estimation model and a modern monocular SLAM method. In this paper, we test our framework, with ORBSLAM \cite{mur2017orb} as the geometric SLAM method and MonoDepth2 \cite{godard2019digging} as the CNN-based model. We show that our self-improving framework outperforms previously proposed self-supervised approaches that utilizes monocular, stereo, and monocular-plus-stereo cues for self-supervision (see Tab. \ref{tab:depth-raw-test-quan}) and a strong feature based RGB-SLAM baseline (see Tab. \ref{tab:pose-odometry-test-quan}).
The framework runs in a simple alternating update fashion: first, we use depth maps from the CNN-based depth network and run pRGBD-SLAM; second, we inject the outputs of pRGBD-SLAM, \ie, the relative camera poses and common tracked keypoints and keyframes to fine-tune the depth network parameters to improve the depth prediction; then, we repeat the process until we see no improvement. Our  specific contributions are summarized here:
\begin{itemize}
    \item We propose a self-improving strategy to inject into depth prediction networks the supervision from SLAM outputs, which stem from more generally applicable geometric principles.
    \item We introduce two wide baseline losses, \ie, the symmetric depth transfer loss and the depth consistency loss on common tracked points, and propose a joint narrow and wide baseline based depth prediction learning setup, where appearance based losses are computed on narrow baselines and purely geometric losses on wide baselines (non-consecutive temporally distant keyframes).
      \item Through extensive experiments on KITTI~\cite{geiger2012we} and TUM RGB-D~\cite{sturm2012benchmark}, our framework is shown to outperform  both monocular SLAM system (\ie, ORB-SLAM~\cite{mur2015orb}) and the state-of-the-art unsupervised single-view depth prediction network (\ie, Monodepth2~\cite{godard2019digging}).
\end{itemize}

\section{Related Work}
\label{sec:relatedwork}
\noindent \textbf{Monocular SLAM.} Visual SLAM has a long history of research in the computer vision community. Due to its well-understood underlying geometry, various geometric approaches have been proposed in the literature, ranging from the classical MonoSLAM~\cite{davison2007monoslam}, PTAM~\cite{klein2007parallel}, DTAM~\cite{newcombe2011dtam} to the more recent LSD-SLAM~\cite{engel2014lsd}, ORB-SLAM~\cite{mur2015orb} and DSO~\cite{engel2017direct}. More recently, in view of the successful application of deep learning in a wide variety of areas, researchers have also started to exploit deep learning approaches for SLAM, in the hope that it can improve certain components of geometric approaches or even serve as a complete alternative. Our work makes further contributions along this line of research.

\noindent \textbf{Monocular Depth Prediction.} Inspired by the pioneering work by Eigen et al.~\cite{eigen2014depth} on learning single-view depth estimation, a vast amount of learning methods~\cite{bloesch2018codeslam,liu2015learning,fu2018deep} emerge along this line of research. The earlier works often require ground truth depths for fully-supervised training. However, per-pixel depth ground truth is generally hard or prohibitively costly to obtain. Therefore, many self-supervised methods that make use of geometric constraints as supervision signals are proposed. Godard et al.~\cite{godard2017unsupervised}, relies on the photo-consistency between the left-right cameras of a calibrated stereo. Zhou et al.~\cite{zhou2017unsupervised} learn monocular depth prediction as well as ego-motion estimation, thereby permitting unsupervised learning with only a monocular camera. This pipeline has inspired a large amount of follow-up works that utilize various additional heuristics, including 3D geometric constraints on point clouds~\cite{mahjourian2018unsupervised}, direct visual odometry~\cite{wang2018learning}, joint learning with optical flow~\cite{yin2018geonet}, scale consistency~\cite{bian2019unsupervised}, and others~\cite{godard2019digging,Sheng_2019_ICCV,chen2019self,zou2020learning}.

\noindent \textbf{Using Depth to Improve Monocular SLAM.}  Approaches~\cite{tateno2017cnn,yin2017scale,yang2018deep,loo2019cnn} leveraging CNN-based depth estimates to tackle issues in monocular SLAM have been proposed. CNN-SLAM~\cite{tateno2017cnn} uses learned depth maps to initialize keyframes' depth maps in LSD-SLAM~\cite{engel2014lsd} and refines them via a filtering framework. Yin et al.~\cite{yin2017scale} use a combination of CNNs and conditional random fields to recover scale from the depth predictions and iteratively refine ego-motion and depth estimates. DVSO~\cite{yang2018deep} trains a single CNN to predict both the left and right disparity maps, forming a virtual stereo pair. The CNN is trained with photo-consistency between stereo images and consistency with depths estimated by Stereo DSO~\cite{wang2017stereo}. More recently, CNN-SVO~\cite{loo2019cnn} uses depths learned from stereo images to initialize depths of keypoints  and reduce their corresponding uncertainties in SVO~\cite{forster2014svo}. In contrast to our self-supervised approach, \cite{tateno2017cnn,yin2017scale} use \emph{ground truth} depths for training depth networks while \cite{yang2018deep,loo2019cnn} need \emph{stereo} images.

\noindent \textbf{Using SLAM to Improve Monocular Depth Prediction.} Depth estimates from geometric SLAM have been leveraged for training monocular depth estimation networks in recent works~\cite{klodt2018supervising,andraghetti2019enhancing}. In~\cite{andraghetti2019enhancing}, sparse depth maps by Stereo ORB-SLAM~\cite{mur2017orb} are first converted into dense ones via an auto-encoder, which are then integrated into geometric constraints for training the depth network. \cite{klodt2018supervising} employ depths and poses by ORB-SLAM~\cite{mur2015orb} as supervision signals for training the depth and pose networks respectively. This approach only considers five consecutive frames, thus restricting its operation in the narrow-baseline setting.

\newcommand{\cent}{$\mathcal{I}_c$}
\newcommand{\kbef}{$\mathcal{I}_{k1}$}
\newcommand{\kaft}{$\mathcal{I}_{k2}$}
\newcommand{\cbef}{$\mathcal{I}_{c\text{-}1}$}
\newcommand{\caft}{$\mathcal{I}_{c+1}$}
\newcommand{\pkbef}{$\textbf{p}^{i}_{k1}$}
\newcommand{\pkaft}{$\textbf{p}^{i}_{k2}$}
\newcommand{\pcent}{$\textbf{p}^{i}_{c}$}
\newcommand{\bpcent}{$\textbf{X}^{i}_{c}$}
\newcommand{\bpkaft}{$\textbf{X}^{i}_{k2}$}
\newcommand{\bpkbef}{$\textbf{X}^{i}_{k1}$}
\newcommand{\bpcentfromkbef}{$\textbf{X}^{i}_{c \rightarrow k1}$}
\newcommand{\bpcentfromkaft}{$\textbf{X}^{i}_{c \rightarrow k2}$}
\newcommand{\bpkbeffromkaft}{$\textbf{X}^{i}_{k2 \rightarrow k1}$}
\newcommand{\bpkbeffromcent}{$\textbf{X}^{i}_{c \rightarrow k1}$}
\newcommand{\bpkaftfromkbef}{$\textbf{X}^{i}_{k2 \rightarrow k2}$}
\newcommand{\bpkaftfromcent}{$\textbf{X}^{i}_{c \rightarrow k2}$}
\newcommand{\intK}{$\textbf{K}$}
\newcommand{\depthCNET}{$d^{i}_{c}(\textbf{w})$}
\newcommand{\depthCSLAM}{$d^{i}_{c}(\text{\tiny SLAM})$}
\newcommand{\depthkbefNET}{$d^{i}_{k1}(\textbf{w})$}
\newcommand{\depthkbefSLAM}{$d^{i}_{k1}(\text{\tiny SLAM})$}
\newcommand{\depthkaftNET}{$d^{i}_{k2}(\textbf{w})$}
\newcommand{\depthkaftSLAM}{$d^{i}_{k2}(\text{\tiny SLAM})$}
\newcommand{\trCtoKbef}{$\textbf{T}_{c\rightarrow k1}^{\text{\tiny SLAM}}$}
\newcommand{\trCtoKaft}{$\textbf{T}_{c\rightarrow k2}$}

\newcommand{\trKbeftoC}{$\textbf{T}_{k1\rightarrow c}^{\text{\tiny SLAM}}$}
\newcommand{\trKbeftoKaft}{$\textbf{T}_{k1\rightarrow k2}$}

\newcommand{\trKafttoC}{$\textbf{T}_{k2\rightarrow k1}$}
\newcommand{\trKafttoKbef}{$\textbf{T}_{k2\rightarrow k1}$}

\newcommand{\trCtoCbef}{$\textbf{T}_{c\text{-}1\rightarrow c}^{\text{\tiny SLAM}}$}
\newcommand{\trCtoCaft}{$\textbf{T}_{c+1\rightarrow c}^{\text{\tiny SLAM}}$}
\section{Method: A Self-Improving Framework}
\label{sec:method}

Our self-improving framework leverages the strengths of each, the unsupervised single-image depth estimation and the geometric SLAM approaches, to mitigate the other's shortcomings. On one hand, the depth network typically generates reliable depth estimates for nearby points, which assist in improving the geometric SLAM estimates of poses and sparse 3D points (Sec. \ref{sec:refinepose}). On the other hand, geometric SLAM methods rely on a more holistic view of the scene to generate robust pose estimates as well as identify \emph{persistent} 3D points that are visible across many frames, thus providing an opportunity to perform wide-baseline and reliable sparse depth estimation. Our framework leverages these sparse, but robust estimates to improve the noisier depth estimates of the farther scene points by minimizing a blend of the symmetric transfer and depth consistency losses (Sec. \ref{sec:refinedepth}) and the commonly used appearance based loss. In the following iteration, this improved depth estimate further enhances the capability of geometric SLAM and the cycle continues until the improvements become negligible. Even in the absence of ground truth, our self-improving framework continues to produce better pose and depth estimates.


\begin{figure}[t]
  \begin{center}
    \includegraphics[width=0.95\linewidth]{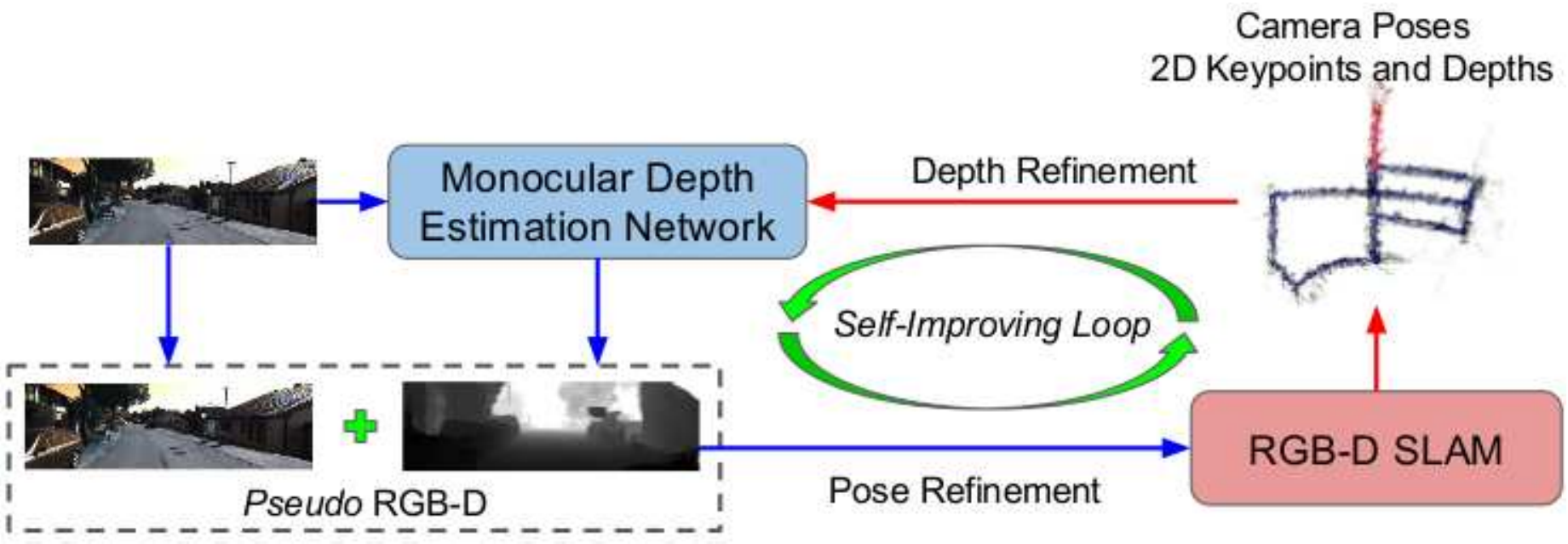}
  \end{center}
  \vspace{-0.5cm}
    	\caption{\textbf{Overview of Our Self-Improving Framework.} It alternates between pose refinement (blue arrows; Sec.~\ref{sec:refinepose}) and depth refinement (red arrows; Sec.~\ref{sec:refinedepth}). } 
	\label{fig:overview}
  \vspace{-0.5cm}
\end{figure}

An overview of the proposed self-improving framework is shown in in Fig.~\ref{fig:overview}, which iterates between improving poses and improving depths. Our pose refinement and depth refinement steps are then detailed in Sec.~\ref{sec:refinepose} and~\ref{sec:refinedepth} respectively. An overview of narrow and wide baseline losses we use for improving the depth network is shown in Fig.~\ref{fig:loss} and details are provided in Sec. \ref{sec:refinedepth}.

\subsection{Pose Refinement}
\label{sec:refinepose}

\textbf{Pseudo RGB-D for Improving Monocular SLAM.} We employ a well explored and widely used geometry-based SLAM system, i.e., the RGB-D version of ORB-SLAM~\cite{mur2017orb}, to process the pseudo RGB-D data, yielding camera poses as well as 3D map points and the associated 2D keypoints. Any other geometric SLAM system that provides these output estimates can also be used in place of ORB-SLAM.
A trivial direct use of pseudo RGB-D data to run RGB-D ORB-SLAM is not possible, because CNN might predict depth at a very different scale compared to depth measurements from real active sensors, \eg, LiDAR. Keeping the above difference in mind, we discuss an important adaptation in order for RGB-D ORB-SLAM to work well in our setting. We first note that RGB-D ORB-SLAM transforms the depth data into disparity on a virtual stereo to reuse the framework of stereo ORB-SLAM. Specifically, considering a keypoint with 2D coordinates $(u_l,v_l)$ (\ie, $u_l$ and $v_l$ denote the horizontal and vertical coordinates respectively) and a CNN-predicted depth $d_l$, the corresponding 2D keypoint coordinates $(u_r,v_r)$ on the virtual rectified right view are $ u_r = u_l - \frac{f_x b}{d_l},~~~v_r = v_l,$ where $f_x$ is the horizontal focal length and $b$ is the virtual stereo baseline.

\noindent \textbf{Adaptation.} In order to have a reasonable range of disparity, we mimic the setup of the KITTI dataset~\cite{geiger2012we} by making the baseline adaptive, $  b = \frac{b^{\text{KITTI}}}{d_{max}^{\text{KITTI}}}*d_{max}$, where $d_{max}$ represents the maximum CNN-predicted depth of the input sequence, and $b^{\text{KITTI}}=0.54$ and $d_{max}^{\text{KITTI}} = 80$ (both in meters) are respectively the actual stereo baseline and empirical maximum depth value of the KITTI dataset. 

We also summarize the overall pipeline of RGB-D ORB-SLAM here. The 3D map is initialized at the very first frame of the sequence due to the availability of depth. After that, the following main tasks are performed: i) track the camera by matching 2D keypoints against the local map, ii) enhance the local map via local bundle adjustment, and iii) detect and close loops for pose-graph optimization and full bundle adjustment to improve camera poses and scene depths. As we will show in Sec.~\ref{sec:pose-eval}, using pseudo RGB-D data leads to better robustness and accuracy as compared to using only RGB data.

\subsection{Depth Refinement}
\label{sec:refinedepth}
\begin{figure}[!t]
      	\centering
\includegraphics[width=0.90\linewidth]{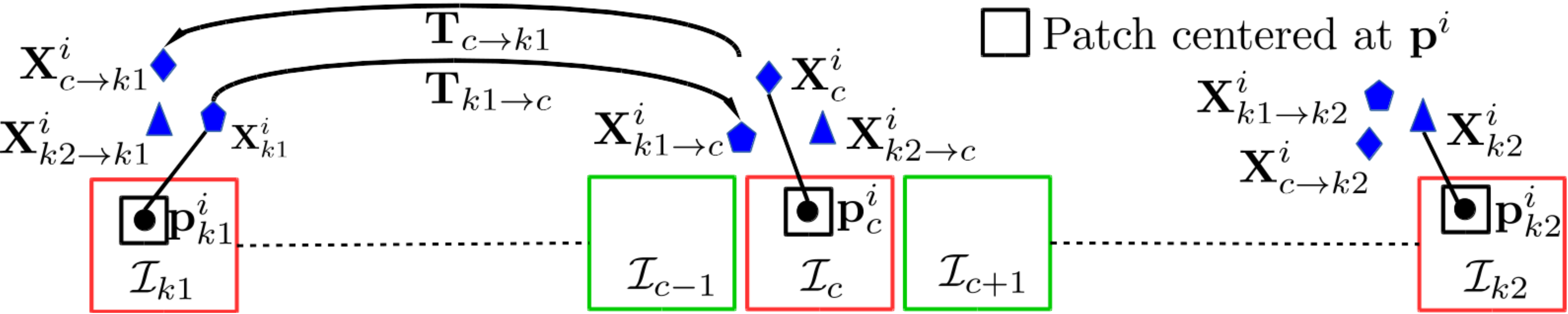}
  	\caption{\textbf{Narrow and Wide Baseline Losses.} Narrow baseline photometric and smoothness losses involve keyframe \cent~and temporally \emph{adjacent} frames \cbef~and \caft, and wide baseline symmetric depth transfer and depth consistency losses involve keyframe \cent~and temporally \emph{farther} keyframes \kbef~and \kaft. Refer to the text below for details. } 
	\label{fig:loss}
\vspace{-0.5cm}
\end{figure}

We start from the pre-trained depth network of Monodepth2~\cite{godard2019digging}, a state-of-the-art monocular depth estimation network, and fine-tune its network parameters with the camera poses, 3D map points and the associated 2D keypoints produced by the above pseudo RGB-D ORB-SLAM (pRGBD-SLAM). In contrast to Monodepth2, which relies only on the narrow baseline photometric reconstruction loss between adjacent frames for short-term consistencies, we propose wide baseline symmetric depth transfer and sparse depth consistency losses to introduce long-term consistencies. Our final loss (Eq.~\eqref{eq:total-loss}) consists of both narrow and wide baseline losses. The narrow baseline losses, \ie, photometric and smoothness losses, involve the current keyframe \cent~and its temporally adjacent frames \cbef~and \caft, while wide baseline losses are computed on the current keyframe \cent~and the two neighboring keyframes \kbef~and \kaft~that are temporally farther than \cbef~and \caft~(see Fig.~\ref{fig:loss}). Next, we introduce the notation and describe the losses in detail.   %

 \noindent \textbf{Notation.} Let $\mathcal{X}$ represent the set of common tracked keypoints visible in all the three keyframes $\mathcal{I}_{k1}$, $\mathcal{I}_{c}$ and $\mathcal{I}_{k2}$ obtained from pRGBD-SLAM. Note that $k1$ and $k2$ are two neighboring keyframes of the current frame $c$ (\ie, $k1 < c < k2$) in which keypoints are visible. 
Let \pkbef$=[p^{i1}_{k1},~p^{i2}_{k1}]$, \pcent$=[p^{i1}_{c},~p^{i2}_{c}]$~and \pkaft$=[p^{i1}_{k2},~p^{i2}_{k2}]$~be the 2D coordinates of the $i^{th}$ common tracked keypoint in the keyframes \kbef, \cent~and \kaft~ respectively, and the associated depth values obtained from pRGBD-SLAM are represented by \depthkbefSLAM,~\depthCSLAM, and \depthkaftSLAM~ respectively. The depth values corresponding to the keypoints \pkbef, \pcent~and \pkaft~can also be obtained from the depth network and are represented by \depthkbefNET,~\depthCNET, and \depthkaftNET~respectively, where $\textbf{w}$ stands for the depth network parameters.

\noindent \textbf{Symmetric Depth Transfer Loss.} Given the camera intrinsic matrix \intK, and the depth value \depthCNET~of the $i^{th}$ keypoint \pcent, the 2D coordinates of the keypoint \pcent~ can be back-projected to its corresponding 3D coordinates as:~\bpcent(\textbf{w})$=\textbf{K}^{-1}[$\pcent$,~1]^{T}$ \depthCNET. Let \trCtoKbef~ represent the relative camera pose of frame $k1$ w.r.t. frame $c$ obtained from pRGBD-SLAM. Using \trCtoKbef,~we can transfer the 3D point \bpcent(\textbf{w}) from frame $c$ to $k1$ as: \bpkbeffromcent(\textbf{w}) =  \trCtoKbef~\bpcent(\textbf{w})$=[x^{i}_{c\rightarrow k1}(\textbf{w}),~y^{i}_{c\rightarrow k1}(\textbf{w}),~d^{i}_{c\rightarrow k1}(\textbf{w})]^{T}$. Here, $d^{i}_{c\rightarrow k1}(\textbf{w})$ is the transferred depth of the $i^{th}$ keypoint from frame $c$ to frame $k1$. Following the above procedure, we can obtain the transferred depth $d^{i}_{k1\rightarrow c}(\textbf{w})$ of the same $i^{th}$ keypoint from frame $k1$ to frame $c$. The symmetric depth transfer loss of the keypoint \pcent~between frame pair $c$ and $k1$, is the sum of absolute errors ($\ell_1$ distance) between the transferred network-predicted depth ${d}^{i}_{c \rightarrow k1}(\textbf{w})$ and the existing network-predicted depth $d^{i}_{k1}(\textbf{w})$ in the target keyframe $k1$,  and vice-versa. Mathematically, it can be written as:
\begin{equation}
\begin{aligned}
\mathcal{T}^{i}_{c \leftrightarrow k1}(\textbf{w}) &{=}|{d}^{i}_{c \rightarrow k1}(\textbf{w}){-}d^{i}_{k1}(\textbf{w})|{+}|{d}^{i}_{k1 \rightarrow c}(\textbf{w}){-}d^{i}_{c}(\textbf{w})|.
\label{eq:sym-depth-transfer}
\end{aligned}
\end{equation} 

Similarly, we can compute the symmetric depth transfer loss of the same $i^{th}$ keypoint between frame pair $c$ and $k2$, \ie, $\mathcal{T}^{i}_{c \leftrightarrow k2}(\textbf{w}) $, and between $k1$ and $k2$, \ie, $\mathcal{T}^{i}_{k1 \leftrightarrow k2}(\textbf{w}) $. We accumulate the total symmetric transfer loss between frame $c$ and $k1$ in $\mathcal{T}_{c \leftrightarrow k1}$, which is the loss of all the common tracked keypoints and the points within the patch of size $5\times5$ centered at the common tracked keypoints. Similarly, we compute the total symmetric depth transfer loss $\mathcal{T}_{c \leftrightarrow k2}$ and $\mathcal{T}_{k1 \leftrightarrow k2}$ between frame pair $(c,k2)$, and $(k1,k2)$ respectively.

\noindent \textbf{Depth Consistency Loss.} The role of the depth consistency loss is to make depth network's prediction consistent with the refined depth values obtained from the pRGBD-SLAM. Note that depth values from pRGBD-SLAM undergo multiple optimization over wide baselines, hence are more accurate and capture long-term consistencies. We inject these long-term consistent depths from pRGBD-SLAM to depth network through the depth consistency loss. The loss for the frame $c$ can be written as follows:
\begin{eqnarray}
    \mathcal{D}_{c} =  \frac{\sum_{i \in \mathcal{X}}|d^{i}_{c}(\textbf{w}) - d^{i}_{c}(\text{\tiny SLAM})|}{|\mathcal{X}|}.
    \label{eq:depth-consis}
\end{eqnarray}

\noindent \textbf{Photometric Reconstruction Loss.} Denote the relative camera pose of frame \cbef~and \caft~w.r.t. current keyframe \cent~obtained from pRGBD-SLAM by \trCtoCbef~and \trCtoCaft~respectively. Using frame \caft,~\trCtoCaft, network-predicted depth map $d_{c}(\textbf{w})$ of the keyframe \cent, and the camera intrinsic \intK, we can synthesize the current frame \cent~\cite{godard2019digging,godard2017unsupervised}. Let the synthesized frame be represented in the functional form as: $\mathcal{I}_{c + 1 \rightarrow c}(d_{c}(\textbf{w}),$~\trCtoCaft,~\intK). Similarly we can synthesize  $\mathcal{I}_{c \text{-} 1 \rightarrow c}(d_{c}(\textbf{w}),$~\trCtoCbef,~\intK) using frame \cbef. The photometric reconstruction error between the synthesized and the original current frame \cite{garg2016unsupervised,godard2017unsupervised,zhou2017unsupervised} is then computed as: 
\begin{equation}
    \mathcal{P}_{c} = pe(\mathcal{I}_{c + 1 \rightarrow c}(d_{c}(\textbf{w}),\textbf{T}_{c+1\rightarrow c}^{\text{\tiny SLAM}},\textbf{K}),\mathcal{I}_{c}) + pe(\mathcal{I}_{c\text{-}1 \rightarrow c}(d_{c}(\textbf{w}),\textbf{T}_{c\text{-}1\rightarrow c}^{\text{\tiny SLAM}},\textbf{K}),\mathcal{I}_{c}),
    \label{eq:photo-smoothness}
\end{equation}
where we follow~\cite{godard2017unsupervised,godard2019digging} to construct the photometric reconstruction error function $pe(\cdot,\cdot)$. Additionally, we adopt the more robust per-pixel minimum error, multi-scale strategy, auto-masking, and depth smoothness loss $\mathcal{S}_c$ from~\cite{godard2019digging}.

Our final loss for fine-tuning the depth network at the depth refinement step is the weighted sum of narrow baseline losses (\ie, photometric ($\mathcal{P}_c$) and smoothness loss ($\mathcal{S}_c$)), and wide baseline losses (\ie, symmetric depth transfer ($\mathcal{T}_{c \leftrightarrow k1}, \mathcal{T}_{c \leftrightarrow k2}, \mathcal{T}_{k1 \leftrightarrow k2}$) and depth consistency loss ($\mathcal{D}_{c}$)):
\begin{equation}
    \mathcal{L} = 
    \alpha \mathcal{P}_c + \beta \mathcal{S}_c  + \gamma \mathcal{D}_{c} + \mu (\mathcal{T}_{c \leftrightarrow k1} + \mathcal{T}_{c \leftrightarrow k2}+\mathcal{T}_{k1 \leftrightarrow k2} ).
      \label{eq:total-loss}
\end{equation}

\section{Experiments}
\label{sec:experiments}
We conduct experiments to evaluate depth refinement and pose refinement steps of our self-improving framework with the state-of-the-arts in self-supervised depth estimation and RGB-SLAM based pose estimation respectively. 

\subsection{Datasets and Evaluation Metrics}
\label{sec:dsets}
\textbf{KITTI Dataset.} Our experiments are mostly performed on the KITTI dataset~\cite{geiger2012we}, which contains outdoor driving sequences for road scene understanding~\cite{song2014robust,dhiman2016continuous}. We further split KITTI experiments into two parts: one focused on depth refinement evaluation and the other on pose refinement. For depth refinement evaluation we train/fine-tune the depth network using the Eigen train split~\cite{eigen2014depth} which contains 28 training sequences and evaluate depth prediction on the Eigen test split~\cite{eigen2014depth} following the baselines~\cite{zhou2017unsupervised,yang2017unsupervised,mahjourian2018unsupervised,yin2018geonet,wang2018learning,zou2018dfnet,yang2018lego,ranjan2019competitive,luo2018every,casser2019depth}. For pose refinement evaluation, we train/fine-tune the depth network using KITTI odometry sequences 00-08 and test on sequences 09-10 and 11-21. Note, for evaluation on sequences 09-10 we use the ground-truth trajectories provided by~\cite{geiger2012we}, while for evaluation on sequences 11-21, since the ground-truth is not available we use the pseudo ground-truth trajectories obtained by running stereo version of ORB-SLAM on these sequences.

\noindent \textbf{TUM RGB-D Dataset.} For completeness and to demonstrate the capability of our self-improving framework on indoor scenes, we evaluate on the TUM RGB-D dataset~\cite{sturm2012benchmark}, which consists of indoor sequences captained by a hand-held camera. We choose \emph{freiburg3} sequences because only they have \emph{undistorted} RGB images and ground truth available to train/fine-tune and evaluate respectively.  We use 6 of 8 \emph{freiburg3} sequences for training/fine-tune and the remaining 2 for evaluation. 

\noindent \textbf{Metrics for Pose Evaluation.} For quantitative pose evaluation, we compute the Root Mean Square Error (\emph{RMSE}), Relative Translation (\emph{Rel Tr}) error, and Relative Rotation (\emph{Rel Rot}) error of the predicted camera trajectory. Since monocular SLAM systems can only recover camera poses up to a global scale, we align the camera trajectory estimated by each method with the ground truth one using the EVO toolbox~\cite{grupp2017evo}. We then use the official evaluation code from the KITTI Odometry benchmark to compute the \emph{Rel Tr} and \emph{Rel Rot} errors for all sub-trajectories with length in $\{100,\dots,800\}$ meters. 

\noindent \textbf{Metrics for Depth Evaluation.} For quantitative depth evaluation, we use the standard metrics, including the Absolute Relative (\emph{Abs Rel}) error, Squared Relative (\emph{Sq Rel}) error, \emph{RMSE}, \emph{RMSE log}, $\delta < 1.25$ (namely \emph{a1}), $\delta < 1.25^2$ (namely \emph{a2}), and $\delta < 1.25^3$ (namely \emph{a3}) as defined in~\cite{eigen2014depth}. Again, since the depths from monocular images can only be estimated up to scale, we align the predicted depth map with the ground truth one using their median depth values. Following~\cite{eigen2014depth} and other baselines, we also clip the depths to 80 meters. \\
\noindent \textbf{Note}. In all tables, best performances are in \textbf{bold} and second bests are {\ul underlined}. 

\subsection{Implementation Details}
We implement our framework based on Monodepth2~\cite{godard2019digging} and ORB-SLAM~\cite{mur2017orb}, \ie, we use the depth network of Monodepth2 and the RGB-D version of ORB-SLAM for depth refinement and pose refinement respectively. We would like to emphasize, that our self-improving strategy is not specific to MonoDepth2 or ORB-SLAM. Any other depth network that allows to incorporate SLAM outputs and any SLAM system that can provide the desired SLAM outputs can be put into the self-improving framework. We set the weight of the smoothness loss term of the final loss (Eq.~\eqref{eq:total-loss}) $\beta=0.001$ similar as in \cite{godard2019digging} and $\alpha$,$\gamma$, and $\mu$ to 1. The ablation study results on disabling different loss terms can be found in Tab. \ref{tab:abl}. A single self-improving loop takes $0.6$ hour on a NVIDIA TITAN Xp 8GB GPU.\\

\noindent \textbf{KITTI Eigen Split/Odometry Experiments.} We pre-train MonoDepth2 using monocular videos of the KITTI Eigen split training set with the hyper-parameters as suggested in MonoDepth2~\cite{godard2019digging}.  We use an input/output resolution of $640 \times 192$ for training/fine-tuning and scale it up to the original resolution while running pRGBD-SLAM. We use same hyperparameters as for KITTI Eigen split to train/fine-tune the depth model on KITTI Odometry train sequences mentioned in Sec. \ref{sec:dsets}. During a self-improving loop, we \emph{discard} pose network of MonoDepth2 and instead use camera poses from pRGBD-SLAM.

\noindent\underline{Outlier Removal.} Before running a depth refinement step, we run an outlier removal step on the SLAM outputs. Specifically, we filter out outlier 3D map points and the associated 2D keypoints that satisfy at least one of the following conditions: i) it is observed in less than 3 keyframes, ii) its reprojection error in the current keyframe $\mathcal{I}_{c}$ is larger than 3 pixels.

\noindent\underline{Camera Intrinsics.} Monodepth2 computes the average camera intrinsics for the KITTI dataset and uses it for the training. However, for our fine-tuning of the depth network, using the average camera intrinsics leads to inferior performance, because we use the camera poses from pRGBD-SLAM, which runs with different camera intrinsics. Therefore, we use different camera intrinsics for different sequences when fine-tuning the depth network.\\
For fine-tuning the depth network pre-trained on KITTI Eigen split training sequences, we run pRGBD-SLAM on all the training sequences, and extract camera poses, 2D keypoints and the associated depths from keyframes. For pRGBD-SLAM(RGB-D ORB-SLAM), we use the default setting of ORB-SLAM, except for the adjusted $b$ described in Sec.~\ref{sec:refinepose}. The same above procedure is followed for depth model pre-trained on KITTI Odometry training sequences. The average number of keyframes used in a self-improving loop is $\sim9K$ and $\sim10K$ for KITTI Eigen split and KITTI Odometry experiments respectively. At each depth refinement step, we fine-tune the depth network parameters with 1 epoch only, using learning rate 1e-6, keeping all the other hyperparameters the same as pre-training. For both KITTI Eigen split and KITTI Odometry experiments we report results after \emph{5 self-improving loops}. \\

\noindent \textbf{TUM RGB-D Experiments.} For TUM RGB-D, we pre-train/fine-tune the depth network on 6 \emph{freiburg3} sequences, and test on 2 \emph{freiburg3} sequences. The average number of keyframes in a self-improving loop is  $\sim3.5K$.  We use an input/output resolution of $480 \times 320$ for pre-training/fine-tuning and scale it up to the original resolution while running pRGBD-SLAM. We report results after \emph{3 self-improving loops}. Other details can be found in the supplementary material.

\subsection{Monocular Depth/Depth Refinement Evaluation}
In the following, we evaluate the performance of our depth estimation on the KITTI Raw Eigen split test set and TUM RGB-D \emph{frieburg3} sequences.\\
\noindent \textbf{Results on KITTI Eigen Split Test Set.} We show the depth evaluation results on the Eigen split test set in Tab.~\ref{tab:depth-raw-test-quan}. From the table, it is evident that our refined depth model (pRGBD-Refined) outperforms all the competing monocular (M) unsupervised methods by non-trivial margins, including MonoDepth2-M re-trained depth model, and even surpasses the unsupervised methods with stereo (S) training, \ie, Monodepth2-S, and combined monocular-stereo (MS) training, \ie, MonoDepth2-MS, in most metrics. Our method also outperforms several ground-truth depth supervised methods~\cite{eigen2014depth,liu2015learning}.
\begin{table}[!t]
       \begin{minipage}{\linewidth}
       	\centering
       	\scriptsize
       	\setlength{\tabcolsep}{1.5pt}
       	\caption{Depth evaluation result on KITTI Eigen split test set. M: self-supervised monocular supervision, and S: self-supervised stereo supervision, D: depth supervision. `-' means the result is not available from the paper. pRGBD-Refined outperforms all the self-supervised monocular methods and several stereo only and combined monocular and stereo methods. Our results are after \emph{5 self-improving loop}s.}
\label{tab:depth-raw-test-quan}
\begin{tabular}{|l|l|c|c|c|c|c|c|c|c|}
\hline
               &         &  & \multicolumn{4}{c}{\cellcolor[HTML]{FFCE93}\textbf{Lower is better}}    & \multicolumn{3}{c|}{\cellcolor[HTML]{CBCEFB}{\color[HTML]{000000} \textbf{Higher is better}}}                                                               \\
 &Method                                 &   Train    & {Abs Rel} & {Sq Rel} & {RMSE}  & {RMSE log} & {a1} & {a2} &{ a3} \\
                                \hline
                                 \parbox[t]{4mm}{\multirow{22}{*}{\rotatebox[origin=c]{90}{ self-supervised}}}&
Yang\cite{yang2017unsupervised}                            & M     & 0.182            & 1.481           & 6.501          & 0.267             & 0.725                              & 0.906                                                     & 0.963                                                     \\
&Mahjourian\cite{mahjourian2018unsupervised}                      & M     & 0.163            & 1.240           & 6.220          & 0.250             & 0.762                              & 0.916                                                     & 0.968                                                     \\
&Klodt\cite{klodt2018supervising}               & M     & 0.166            & 1.490         & 5.998          & -             & 0.778                             &0.919                                                     & 0.966   \\
&DDVO\cite{wang2018learning}                            & M     & 0.151            & 1.257           & 5.583          & 0.228             & 0.810                              & 0.936                                                     & 0.974                                                     \\
&GeoNet\cite{yin2018geonet}                          & M     & 0.149            & 1.060           & 5.567          & 0.226             & 0.796                              & 0.935                                                     & 0.975                                                     \\
&DF-Net\cite{zou2018dfnet}                          & M     & 0.150            & 1.124           & 5.507          & 0.223             & 0.806                              & 0.933                                                     & 0.973                                                     \\
&Ranjan\cite{ranjan2019competitive}                          & M     & 0.148            & 1.149           & 5.464          & 0.226             & 0.815                              & 0.935                                                     & 0.973                                                     \\
&EPC++\cite{luo2018every}                           & M     & 0.141            & 1.029           & 5.350          & 0.216             & 0.816                              & 0.941                                                     & 0.976                                                     \\
&Struct2depth(M)\cite{casser2019depth}                & M     & 0.141            & 1.026           & 5.291          & 0.215             & 0.816                              & 0.945                                                     & 0.979                                               \\
&WBAF~\cite{zhou2020windowed}                            &M &             0.135 &    0.992 &          5.288 &        0.211 &   0.831 &  0.942 &  0.976 \\
&MonoDepth2-M (re-train)~\cite{godard2019digging}              & M     &  {0.117}            &  {0.941}           & {4.889}          &  {0.194}             & {0.873}                              &  {0.957}                                                     &  {0.980}                                                     \\
&MonoDepth2-M (original)~\cite{godard2019digging}     &  M    &          {\ul0.115}  &   {\ul0.903}   &        {\ul4.863}  &      {\ul0.193} &   \textbf{0.877} &  {\ul0.959} &  {\ul0.981} \\
&pRGBD-Refined           & M     & \textbf{0.113}   & \textbf{0.793}  & \textbf{4.655} & \textbf{0.188}    & {\ul0.874}                        & \textbf{0.960}                                            & \textbf{0.983}   \\

\cline{2-10}
&Garg\cite{garg2016unsupervised}                            & S     & 0.152            & 1.226           & 5.849          & 0.246             & 0.784                              & 0.921                                                     & 0.967                                                             \\
&3Net (R50)\cite{poggi2018learning}                      & S     & 0.129            & 0.996           & 5.281          & 0.223             & 0.831                              & 0.939                                                     & 0.974                                                     \\
&Monodepth2-S\cite{godard2019digging}                & S     & {0.109}            & 0.873           & {4.960}         & 0.209             & {0.864}                              & {0.948}                                                    & 0.975    \\
&SuperDepth \cite{pillai2019superdepth}  & S     & 0.112            & 0.875           & {4.958}          & {0.207}            & {0.852}                           & {0.947}                                                     & {0.977}                                                                                        \\
&monoResMatch~\cite{tosi2019learning}            &  S  &             0.111 &   0.867  &         4.714  &       0.199 &   0.864 &  {\ul0.954} &  {\ul0.979}\\
&DepthHints~\cite{watson2019self}                          & S       &       {\ul0.106} &   {\ul0.780}   &        {\ul4.695}   &      {\ul0.193}  &  {\ul0.875} &  \textbf{0.958} &  \textbf{0.980}\\
&DVSO\cite{yang2018deep}               & S     & \textbf{0.097}            & \textbf{0.734}         & \textbf{4.442}         & \textbf{0.187}             & \textbf{0.888}                             &\textbf{0.958}                                                     & \textbf{0.980}    \\
\cline{2-10}
&UnDeepVO~\cite{li2018undeepvo}               & MS     & 0.183            & 1.730         & 6.570          & 0.268             & -                             &-                                                     & -   \\
&EPC++~\cite{luo2018every} & MS     & {\ul0.128}            & {\ul0.935}         & {\ul5.011}          & {\ul0.209}           & {\ul0.831}                             &{\ul0.945}                                                    &  \textbf{0.979}   \\
&Monodepth2-MS\cite{godard2019digging}                & MS     &  \textbf{0.106}            & \textbf{0.818}           & \textbf{4.750}          & \textbf{0.196}             & \textbf{0.874}                              & \textbf{0.957}                                                     & \textbf{0.979}  \\
\hline
\parbox[t]{3mm}{\multirow{9}{*}{\rotatebox[origin=c]{90}{ }}}&Eigen\cite{eigen2014depth}               & D     & 0.203            & 1.548         & 6.307            & 0.282             & 0.702                             & 0.890                                                   & 0.890   \\
&Liu\cite{liu2015learning}               & D     & 0.201            & 1.584         & 6.471          & 0.273             & 0.680                             & 0.898                                                     & 0.967   \\
&Kuznietsov\cite{kuznietsov2017semi}               & DS     & 0.113            & 0.741         & 4.621          & 0.189             & 0.862                             & 0.960                                                     & 0.986   \\
&SVSM FT\cite{luo2018every}               & DS     & {\ul0.094}            & {\ul0.626}        & 4.252          & 0.177             & 0.891                             &0.965                                                     & 0.984   \\
&Guo\cite{guo2018learning}               & DS     & 0.096           & 0.641         & { \ul4.095}          & {\ul0.168}             & {\ul0.892}                            &{\ul0.967}                                                   & {\ul0.986}   \\
&DORN\cite{fu2018deep}               & D     & \textbf{0.072}            & \textbf{0.307}         & \textbf{2.727}          & \textbf{0.120}             &  \textbf{0.932}                             &\textbf{0.984}                                                    & \textbf{0.994}   \\
\hline
\end{tabular}
       \end{minipage} 
\end{table}

\begin{figure*}[h]
\centering
\setlength{\tabcolsep}{2pt}
\begin{tabular}{ccccc}
\includegraphics[width=3.01cm]{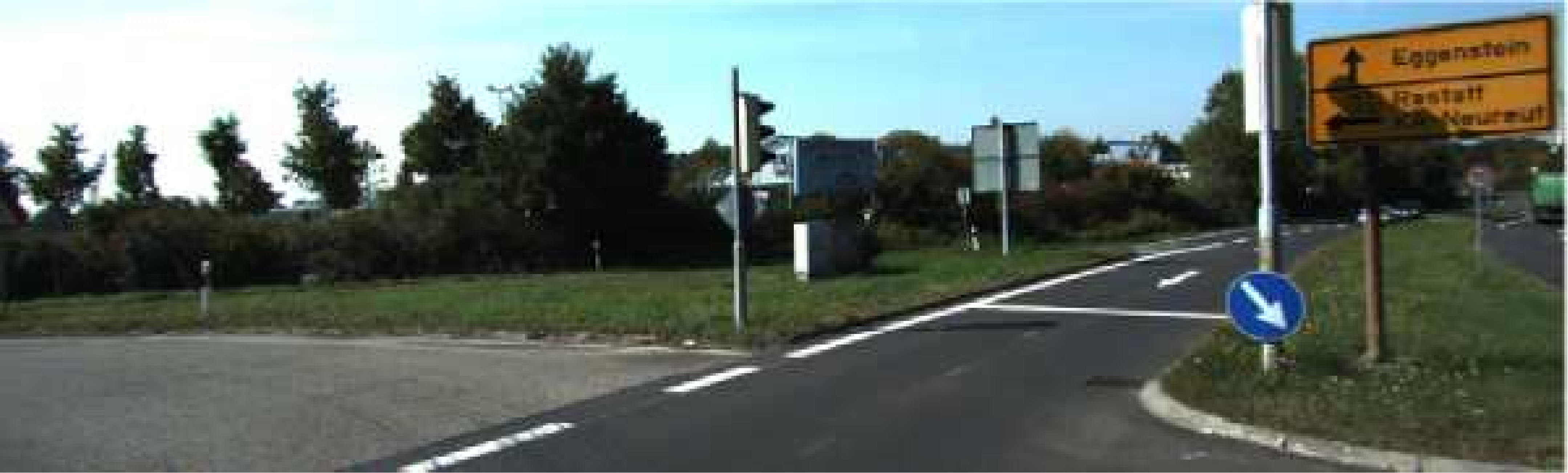} &
\includegraphics[width=3.01cm]{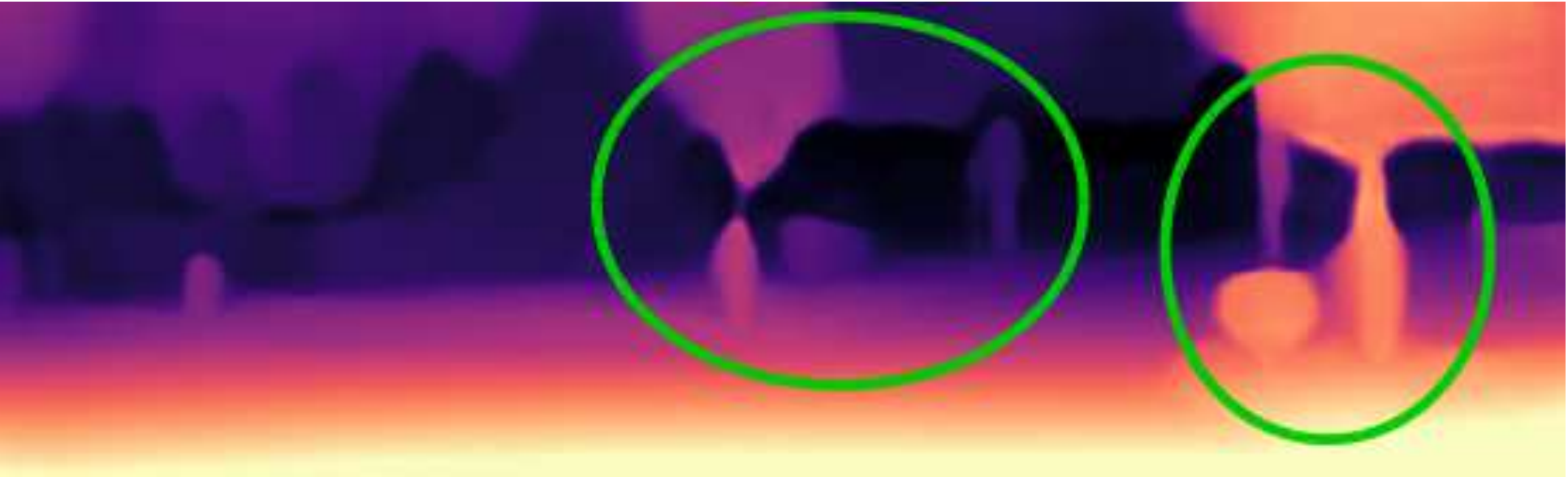} &
\includegraphics[width=3.01cm]{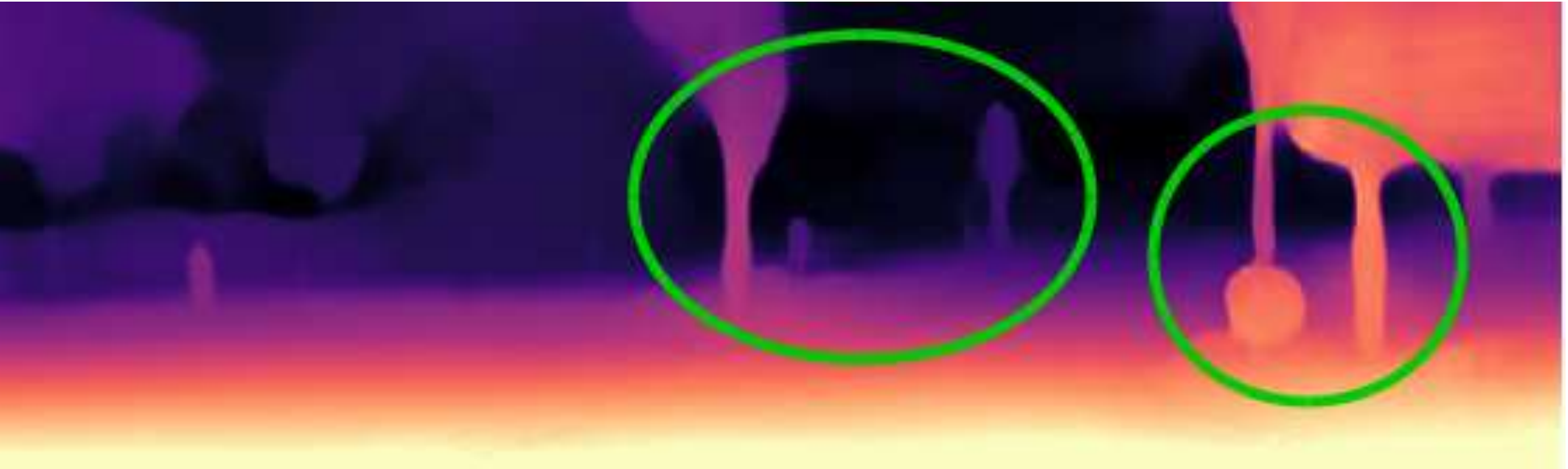} &
\includegraphics[width=3.01cm]{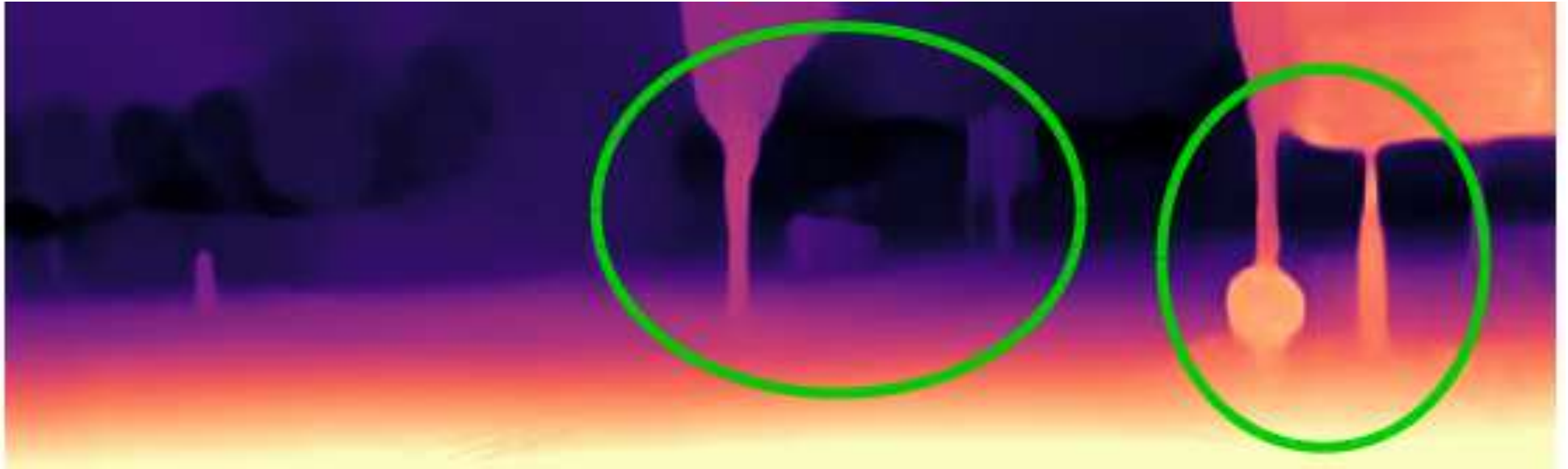} \vspace{-0.1cm}\\ 
\includegraphics[width=3.01cm]{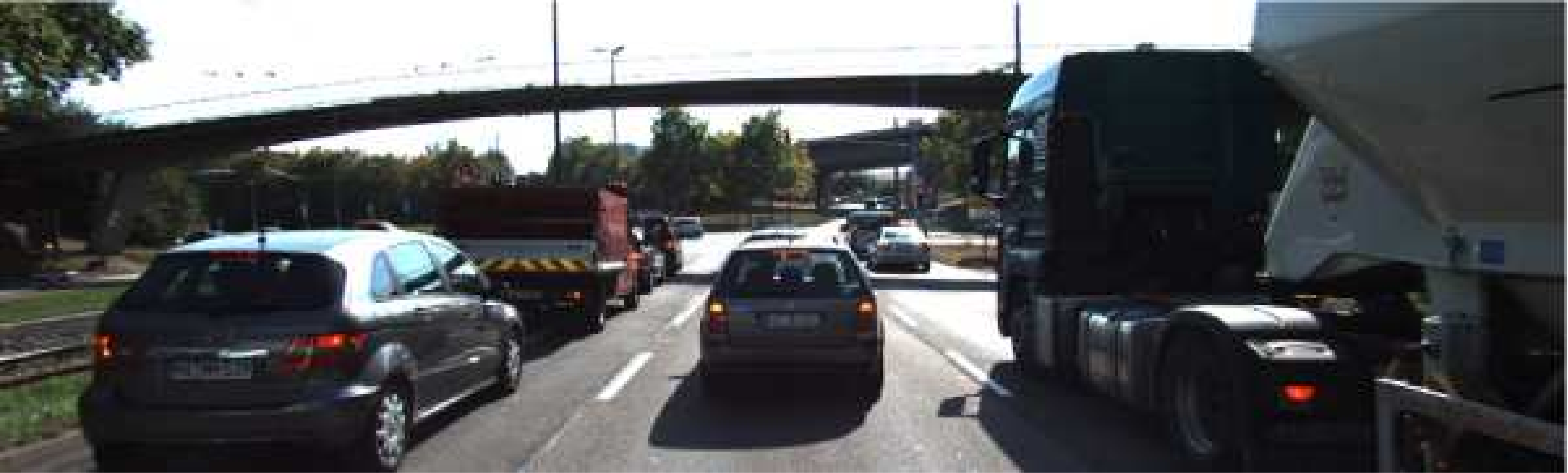} &
\includegraphics[width=3.01cm]{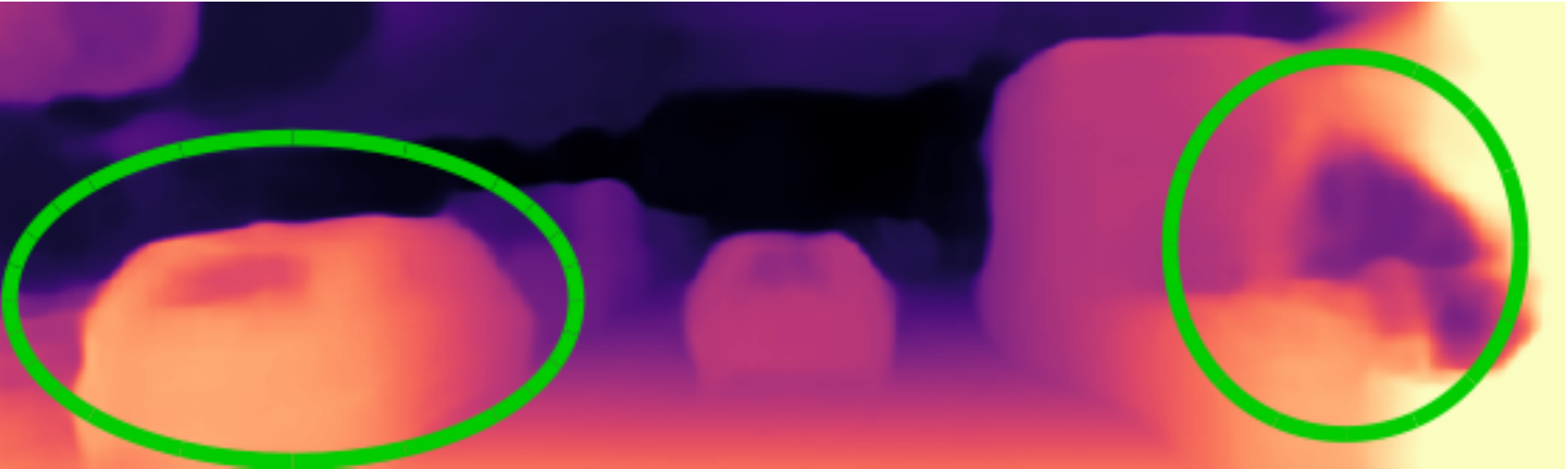} &
\includegraphics[width=3.01cm]{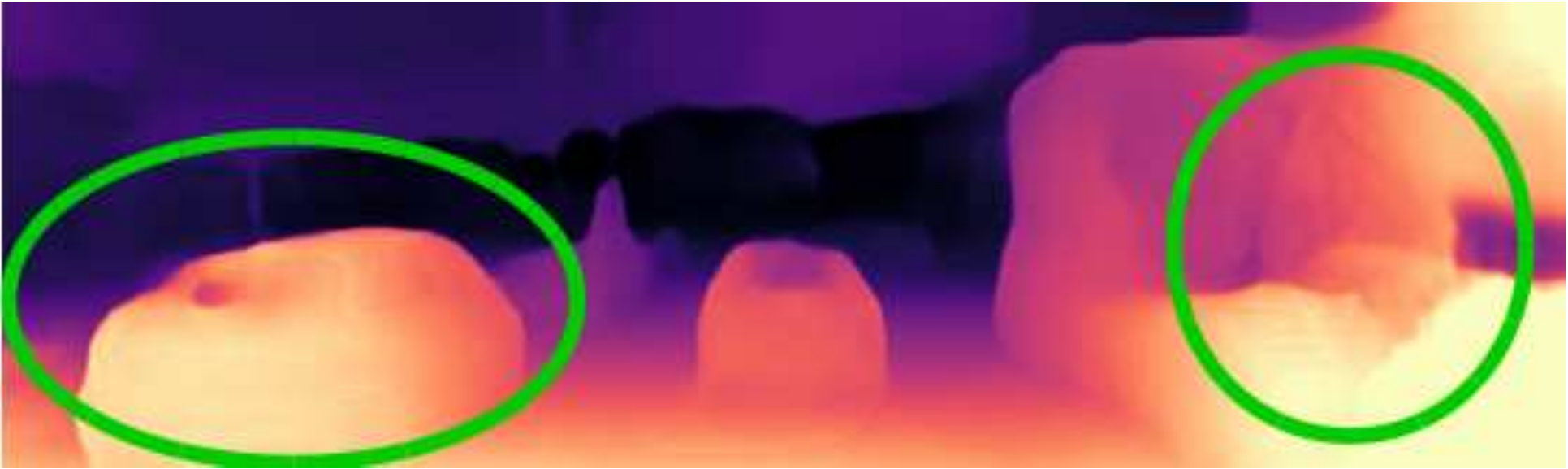} &
\includegraphics[width=3.01cm]{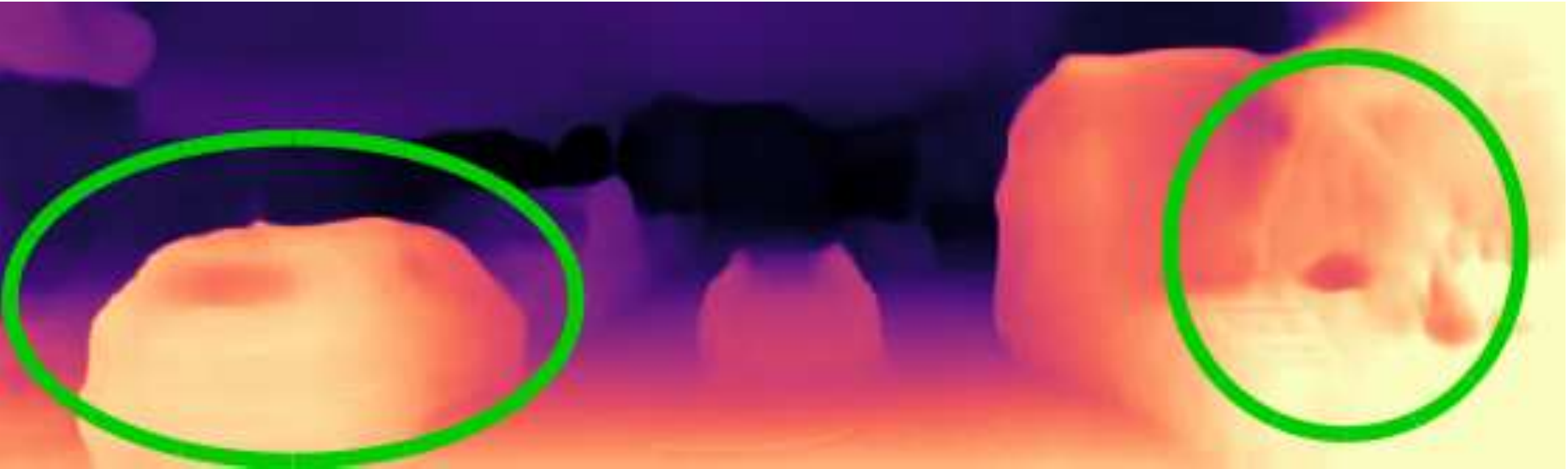} \vspace{-0.1cm} \\

\includegraphics[width=3.01cm]{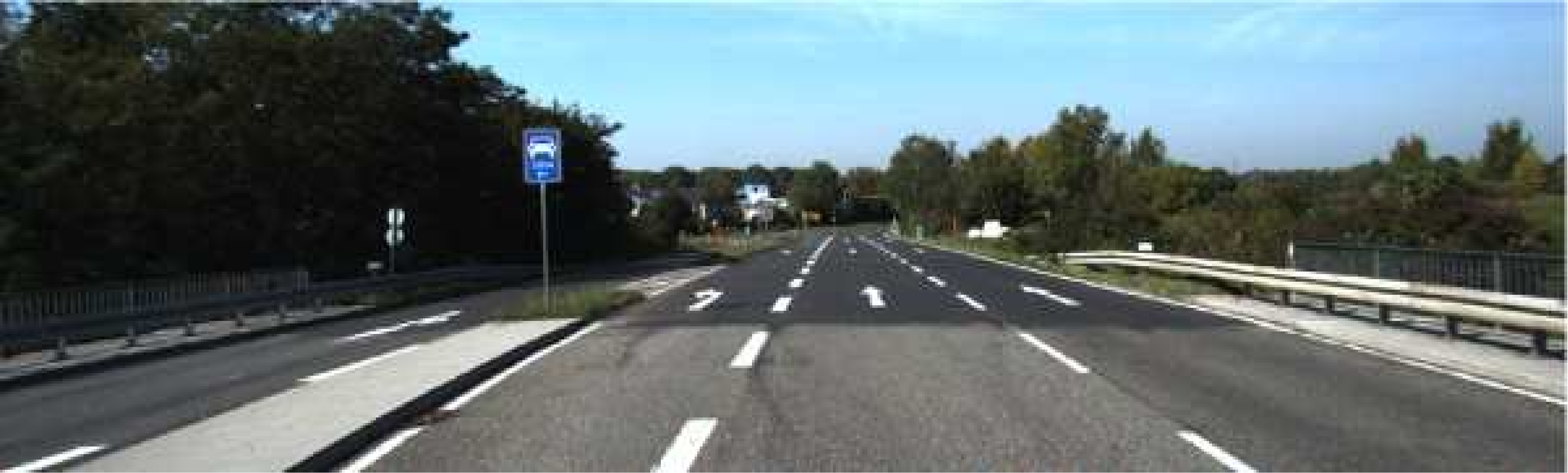} &
\includegraphics[width=3.01cm]{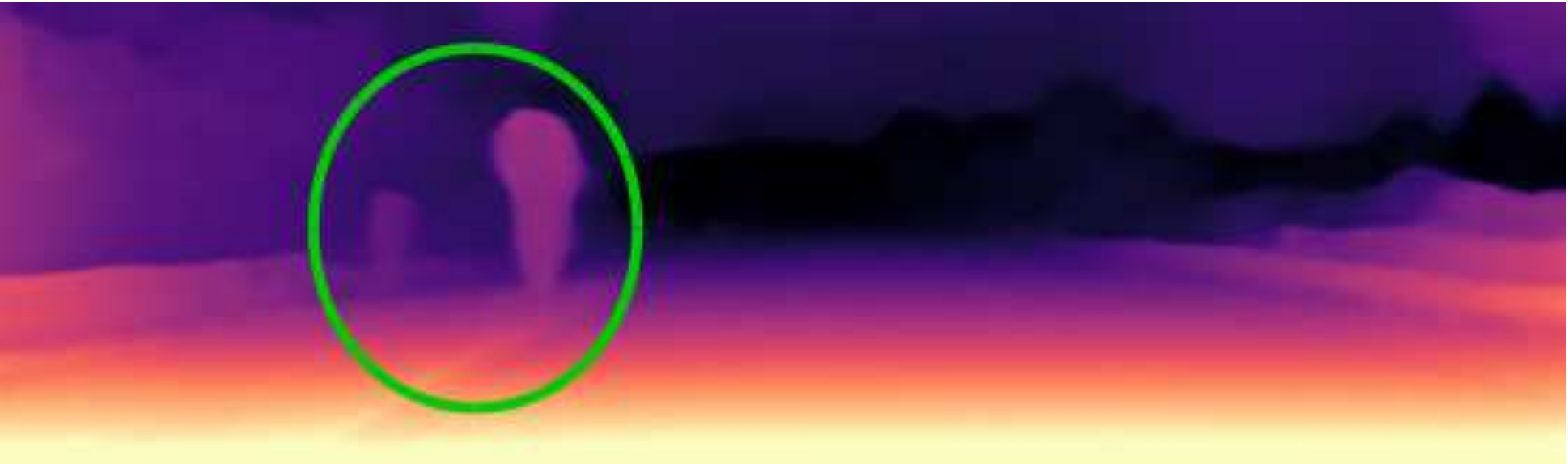} &
\includegraphics[width=3.01cm]{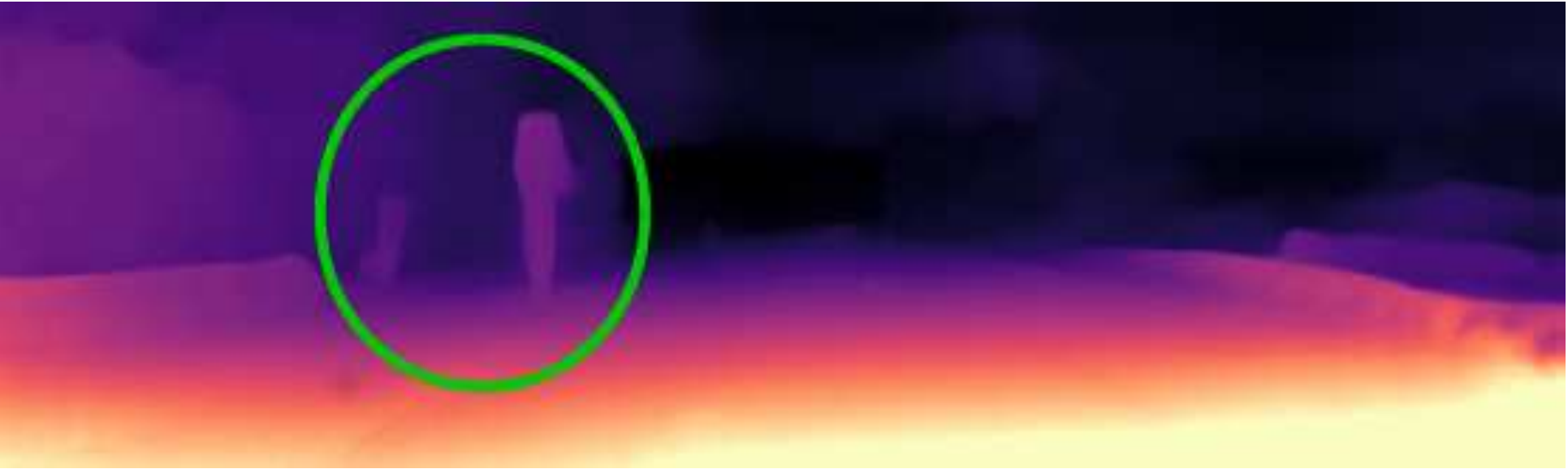}&
\includegraphics[width=3.01cm]{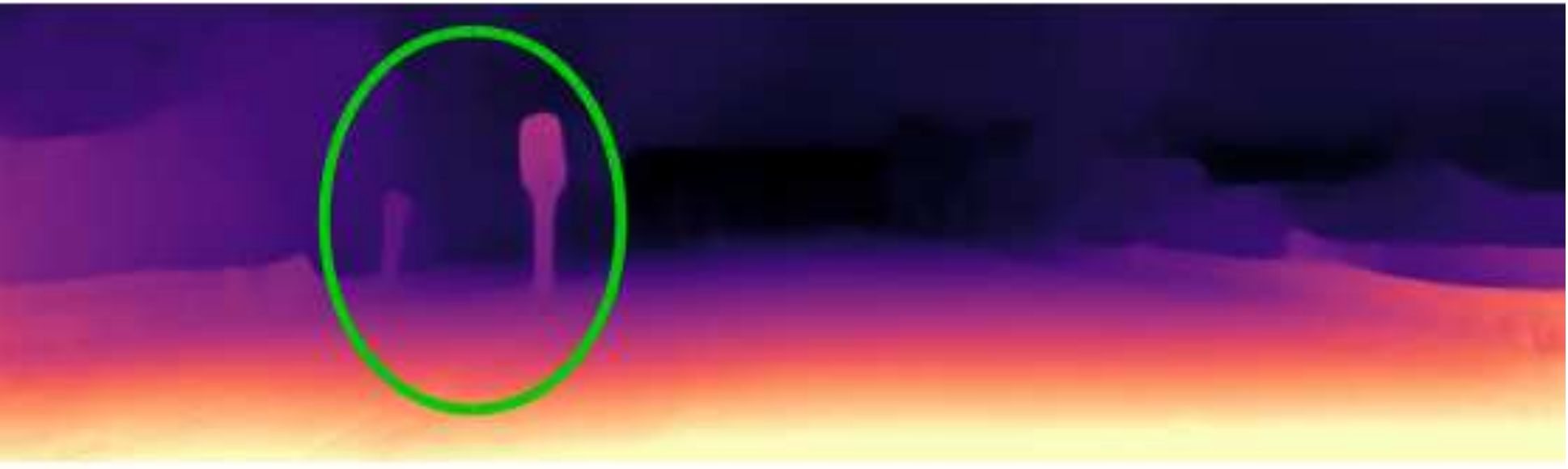} \vspace{-0.15cm}\\
{\scriptsize RGB}   & {\scriptsize Monodepth2-S} & {\scriptsize Monodepth2-M} & {\scriptsize pRGBD-Refined} \\
\end{tabular}
\vspace{-0.3cm}
\caption{Qualitative depth evaluation results on KITTI Raw Eigen's split test set. }
\label{fig:depth-raw-test-qual}
\vspace{-0.5cm}
\end{figure*}
The reason is probably that the aggregated cues from multiple views with wide baseline losses (\eg, our symmetric depth transfer, depth consistency losses) lead to more well-posed depth recovery, and hence even higher accuracy than learning with the pre-calibrated stereo rig with smaller baselines. Further analysis is provided in Sec.~\ref{sec:self-loop-analysis}. Fig.~\ref{fig:depth-raw-test-qual} shows some qualitative results, where our method (pRGBD-Refined) shows visible improvements. Refer supplementary material for more qualitative results.

\noindent \textbf{Results on TUM RGB-D Sequences.} The depth evaluation results on the two TUM \textit{frieburg3} RGB-D sequences is shown in Tab.~\ref{tab:depth-tum}. Our refined depth model (pRGBD-Refined) outperforms pRGBD-Initial/Monodepth2-M in both sequences and all metrics. Refer supplementary material for qualitative results. 

\subsection{Monocular SLAM/Pose Refinement Evaluation }
\label{sec:pose-eval}
In this section, we evaluate pose estimation/refinement on the KITTI Odometry sequences 09 and 10, KITTI Odometry test set sequences 11-21, and two TUM \textit{frieburg3} RGB-D sequences.\\

\begin{table}[h]
    	\centering
    	\scriptsize
    	\setlength{\tabcolsep}{2pt}
    	 \vspace{-0.7cm}
       	 \caption{ Quantitative pose evaluation results on KITTI Odometry validation set. ‘-’ means the result is not available from the paper.}
 \label{tab:pose-odometry-val-quan}
 \vspace{-0.4cm}
 \begin{tabular}{|l|l|ccc|ccc|}
 \hline
 & & & Seq. 09 & & &
 Seq. 10 &
 \\
 &Method & RMSE & Rel Tr & Rel Rot & RMSE & Rel Tr & Rel Rot \\

 \hline
 \parbox[t]{2mm}{\multirow{6}{*}{\rotatebox[origin=c]{90}{Supervised}}}&DeepVO\cite{wang2017deepvo} 
 & - & - & - & - & 8.11 & 0.088\\

 &ESP-VO\cite{wang2018end} 
 & - & - & - & - & 9.77 & 0.102 \\

 &GFS-VO\cite{xue2018guided} 
 & - & - & - & - &  {\ul6.32} &  {\ul0.023} \\

 &GFS-VO-RNN\cite{xue2018guided} 
 & - & - & - & - & 7.44 & 0.032 \\

 &BeyondTracking\cite{xue2019beyond} 
 & - & - & - & - &   \textbf{3.94} &  \textbf{0.017} \\

 &DeepV2D\cite{teed2018deepv2d}
 &  \textbf{79.06} &  \textbf{8.71} &  \textbf{0.037} &  \textbf{48.49} & 12.81 & 0.083 \\

 \hline
 \parbox[t]{2mm}{\multirow{12}{*}{\rotatebox[origin=c]{90}{Self-Supervised}}}&SfMLearner~\cite{zhou2017unsupervised} 
 &  \textbf{24.31} & 8.28 & 0.031 & 20.87 & 12.20 &  \textbf{0.030} \\

 &GeoNet\cite{yin2018geonet} 
 & 158.45 & 28.72 & 0.098 & 43.04 & 23.90 & 0.090 \\

 &Depth-VO\cite{zhan2018unsupervised} 
 & - & 11.93 & 0.039 & - & 12.45 & 0.035 \\

 &vid2depth\cite{mahjourian2018unsupervised} 
 & - & - & - &- & 21.54 & 0.125 \\

 &UnDeepVO\cite{li2018undeepvo} 
 & - &  { {\ul7.01}} & 0.036 & - & 10.63 & 0.046 \\

 &Wang~\etal\cite{wang2019recurrent} 
 & - & 9.88 & 0.034 & - & 12.24 & 0.052 \\

 &CC\cite{ranjan2019competitive} 
 & 29.00 &  \textbf{6.92} &  \textbf{0.018} &  \textbf{13.77} &  {\ul7.97} &  {\ul0.031} \\

 &DeepMatchVO\cite{shen2019icra} 
 &  {\ul27.08} & 9.91 & 0.038 & 24.44 & 12.18 & 0.059 \\

 &Li~\etal\cite{li2019pose} 
 & - & 8.10 &  {\ul0.028} & - & 12.90 & 0.032 \\

 &Monodepth2-M\cite{godard2019digging} 
 & 55.47 & 11.47 & 0.032 &  {\ul20.46} &  \textbf{7.73} & 0.034 \\

 &SC-SfMLearer\cite{bian2019unsupervised}
 & - & 11.2 & 0.034 & - & 10.1 & 0.050 \\
 \hline
 &RGB ORB-SLAM
 &18.34         &  7.42               & \textbf{0.004} &   8.90          & 5.85            & \textbf{0.004} \\
 &pRGBD-Initial
 &   {\ul12.21}         &  {\ul4.26}            & 0.011 &   {\ul8.30}          &  {\ul 5.55}            & 0.017  \\
 &pRGBD-Refined
 &  \textbf{11.97}   & \textbf{4.20}   & {\ul0.010}  &  \textbf{6.35} &  \textbf{4.40}  &  {\ul0.016}  \\
 \hline
 \end{tabular}
  \vspace{-0.4cm}
\end{table}

\noindent \textbf{Results on KITTI Odometry Sequences 09 and 10.} We show the quantitative results on seqs 09 and 10 in Tab.~\ref{tab:pose-odometry-val-quan}. It can be seen that our pRGBD-Initial outperforms RGB ORB-SLAM~\cite{mur2015orb} both in terms of RSME and Rel Tr. 
Our pRGBD-Refined further improves pRGBD-Initial in all metrics, which verifies the effectiveness of our self-improving mechanism in terms of pose estimation. The higher Rel Rot errors of our methods compared to RGB ORB-SLAM could be due to the high uncertainty of CNN-predicted depths for far-away points, which affects our rotation estimation~\cite{hartley2003multiple}. In addition, our methods outperform all the competing supervised and self-supervised methods by a large margin, except for the supervised method of~\cite{xue2019beyond} with lower Rel Tr than ours on sequence 10. Note that we evaluate the camera poses produced by the pose network of Monodepth2-M~\cite{godard2019digging} in Tab.~\ref{tab:pose-odometry-val-quan}, yielding much higher errors than ours.
Fig.~\ref{fig:pose-kitti}(a) shows the camera trajectories estimated for sequence 09 by RGB ORB-SLAM, our pRGBD-Initial, and pRGBD-Refined. It is evident that, although all the methods perform loop closure successfully, our methods generate camera trajectories that align better with the ground truth. \\
\noindent \textbf{Results on KITTI Odometry Test Set.}  The KITTI Odometry leaderboard requires complete camera trajectories of all frames of all the sequences. Since we keep the default setting from ORB-SLAM, causing tracking failures in a few sequences, to facilitate quantitative evaluation on this test set (\ie, sequences 11-21), we use pseudo-ground-truth computed as mentioned in Sec.~\ref{sec:dsets} to evaluate all the competing methods in Tab.~\ref{tab:pose-odometry-test-quan}. 
From the results, RGB ORB-SLAM fails on three challenging sequences due to tracking failures, whereas our pRGBD-Initial fails on two sequences and our pRGBD-Refined fails only on one sequence. Among the sequences where all the competing methods succeed, our pRGBD-Initial reduces the RMSEs of RGB ORB-SLAM by a considerable margin for all sequences except for sequence 19. After our self-improving mechanism, our pRGBD-Refined further boosts the performance, reaching the best results both in terms of RMSE and Rel Tr. Fig.~\ref{fig:pose-kitti}(b) shows qualitative comparisons on sequence 19.\\

\begin{table}[h]
      	\centering
      	\scriptsize
      	\setlength{\tabcolsep}{2pt}
      	\vspace{-0.9cm}
      	    	  \caption{ Quantitative pose evaluation results on KITTI Odometry test set. Since the ground truth for the KITTI Odometry test set is not available we run Stereo ORB-SLAM\cite{mur2017orb} to get the complete camera trajectories and use them as the pseudo ground truth to evaluate.  `X' denotes tracking failure. }
      	    	    \label{tab:pose-odometry-test-quan}
 \vspace{-0.2cm}
      	 \begin{tabular}{|c|c|c|c|c|c|c|c|c|c|}
 \hline
 \multirow{2}{*}{Seq} & \multicolumn{3}{c|}{RGB ORB-SLAM}            & \multicolumn{3}{c|}{pRGBD-Initial}               & \multicolumn{3}{c|}{pRGBD-Refined}             \\ 
                                  & RMSE & Rel Tr & Rel Rot & RMSE & Rel Tr & Rel Rot & RMSE  & Rel Tr & Rel Rot \\ \hline
 11                               & 14.83         & 7.69            & \textbf{0.003}   &  {\ul6.68}          & {\ul3.28}            & 0.016            & \textbf{3.64}   & \textbf{2.96}   & {\ul 0.015}            \\
 13                               &  {\ul6.58}          & {\ul2.39}            & \textbf{0.006}   & 6.83          & 2.52            & 0.008            & \textbf{6.43}   & \textbf{2.31}   & {\ul0.007}            \\ 
 14                               & 4.81          & 5.19            & \textbf{0.004}   &  {\ul4.30}          & {\ul4.14}           & {\ul0.014}            & \textbf{2.15}   & \textbf{3.06}   &{\ul 0.014}            \\ 
 15                               & 3.67          & 1.78            & \textbf{0.004}   &  {\ul2.58}          & {\ul1.61}            & 0.005            & \textbf{2.07}   & \textbf{1.33}   & \textbf{0.004}   \\
 16                               & 6.21          & 2.66            &  \textbf{0.002}   &  {\ul5.78}          & {\ul2.14}            & 0.006            & \textbf{4.65}   & \textbf{1.90}   & {\ul0.004}           \\ 
 18                               & 6.63          & 2.38            &  \textbf{0.002}   &  {\ul5.50}          & {\ul2.30}            & 0.008            & \textbf{4.37}   & \textbf{2.21}   & {\ul0.006}           \\ 
 19                               &  {\ul18.68}         & 4.91            & \textbf{0.002}   & 23.96         & {\ul2.82}            & 0.007            & \textbf{13.85}  &\textbf{2.52}   & {\ul0.006}            \\ 
 20                               & 9.19          & 6.74            & \textbf{0.016}   &  {\ul8.94}          & {\ul5.43}            & 0.027            & \textbf{7.03}   & \textbf{4.50}    & {\ul0.022}            \\
 12                               & X             & X               & X                & X             & X               & X                & \textbf{94.2} & \textbf{32.94}  &  \textbf{0.026}            \\
 17                               & X             & X               & X                & {\ul14.71}        & {\ul8.98}            & \textbf{0.011}            & \textbf{12.23}  & \textbf{7.23}   &  \textbf{0.011}            \\
 21                               & X             & X               & X                & X             & X               & X                & X               & X               & X                \\ \hline
 \end{tabular}
 \vspace{-0.8cm}
\end{table}

\begin{table}[H]
     	\centering
\vspace{-0.7cm}
	\setlength{\tabcolsep}{2pt}
\begin{tabular}{cc}
\hspace{0.1cm}{\includegraphics[width=3.2cm]{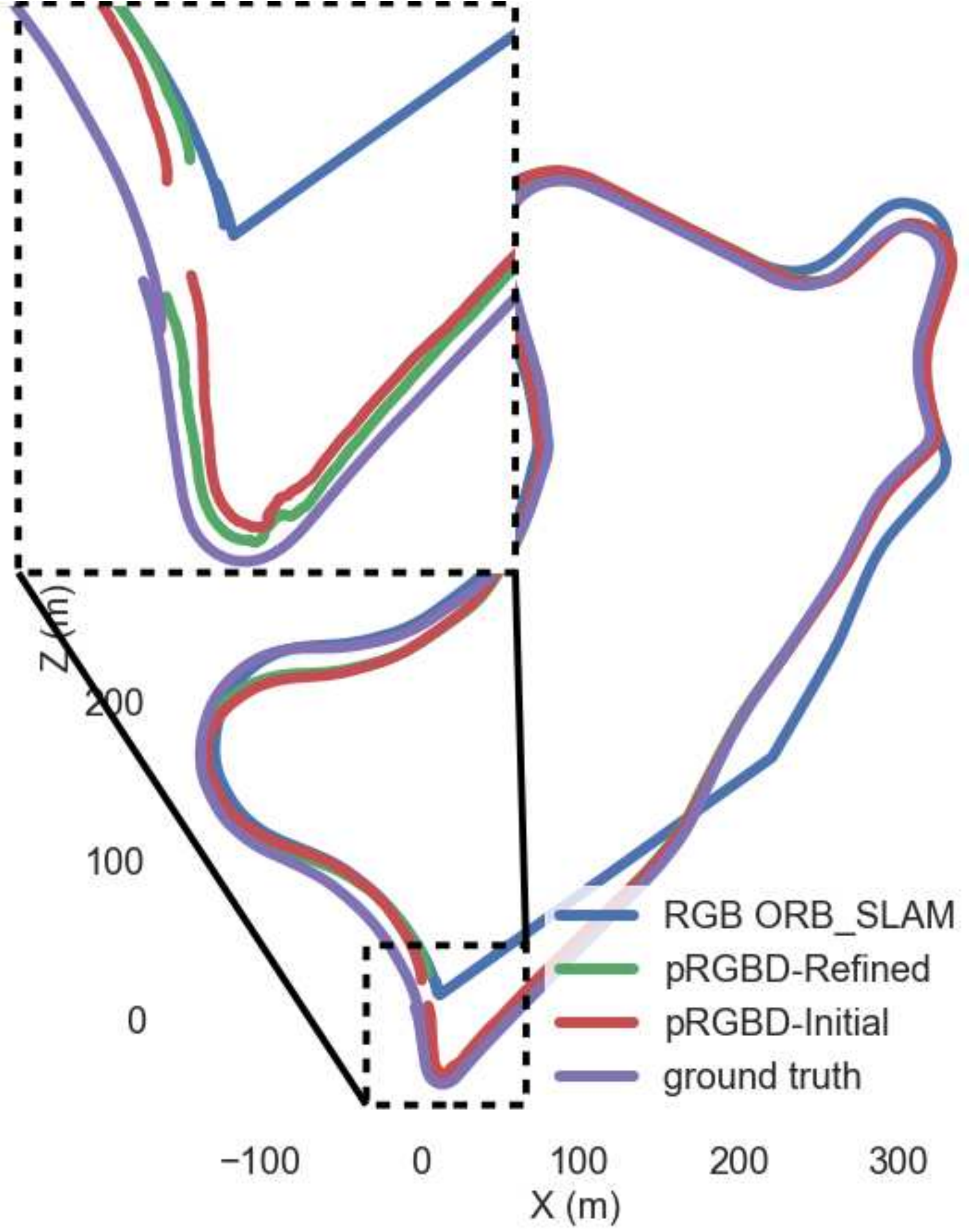}} & \hspace{0.5cm}
\hspace{0.1cm}{\includegraphics[width=3.3cm]{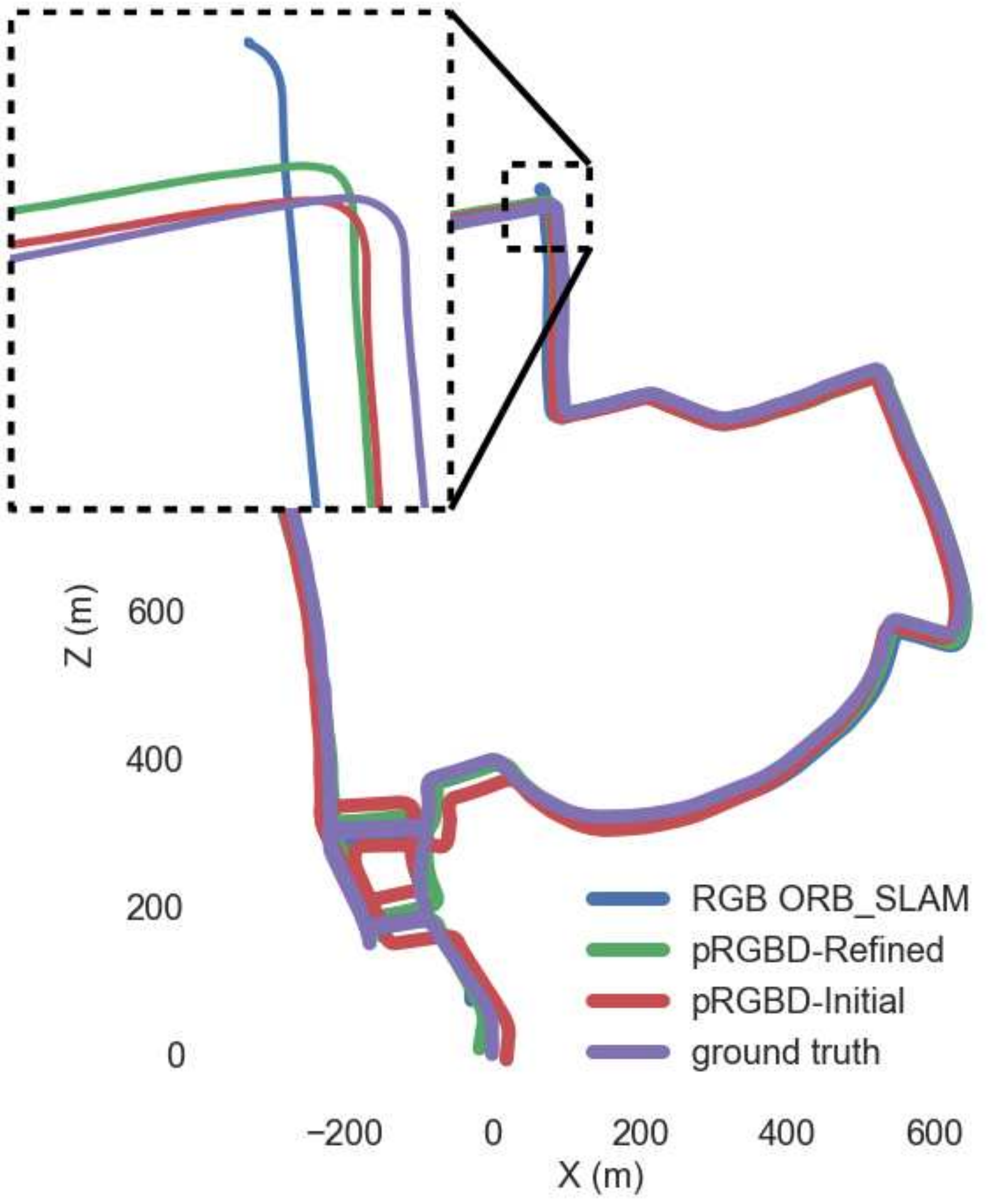}} \\
(a) seq 09 &(b) seq 19 \\
\end{tabular}
\renewcommand{\tablename}{\textbf{Fig.}}
\setcounter{table}{4} 
\hspace{0.5cm}  \caption{ Qualitative pose evaluation results on KITTI sequences.  }
\renewcommand{\tablename}{\textbf{Table}}
\setcounter{table}{3} 
  	\label{fig:pose-kitti}
 \vspace{-0.6cm}
\end{table}
\noindent \textbf{Results on TUM RGB-D Sequences.} Performance of pose refinement step on the two TUM RGB-D sequences is shown in Tab.~\ref{tab:pose-tum}. The result shows increased robustness and accuracy by pRGBD-Refined. In particular, RGB ORB-SLAM fails on walking\_xyz, while pRGBD-Refined succeeds and achieves the best performance on both sequences. Due to space limitation we have moved qualitative results to supplementary material.

\begin{table}[H]
   	\centering
       	\scriptsize
       	\setlength{\tabcolsep}{1pt}
 \vspace{-0.5cm}
       	  	\caption{ Quantitative pose evaluation results on two TUM \textit{frieburg3} RGB-D sequences. Note that RGB ORB-SLAM fail in walking\_xyz sequence. }
 \label{tab:pose-tum}
 \vspace{-0.1cm}
      	 \begin{tabular}{|c|c|c|c|c|c|c|c|c|c|}
 \hline
 \multirow{2}{*}{} & \multicolumn{3}{c|}{RGB SLAM}            & \multicolumn{3}{c|}{pRGBD-Initial}               & \multicolumn{3}{c|}{pRGBD-Refined}             \\ 
                                  & RMSE & RlTr & RlRot & RMSE & RlTr & RlRot & RMSE  & RlTr & RlRot \\ \hline
 W                               &  X         & X            & X   &  {\ul0.23}          & {\ul0.02}            & {\ul0.52}            & \textbf{0.09}   & \textbf{0.01}   & \textbf{0.30}            \\
 L                               &  1.72          & {\ul0.02}            & \textbf{0.32}   & {\ul1.40}          & \textbf{0.01}            & 0.34            & \textbf{0.39}   & \textbf{0.01}   & {\ul0.33}          \\ 
 \hline
 \end{tabular}
\end{table}
\vspace{-1.2cm}
\begin{table}[H]
    	\centering
    	\scriptsize
    	\setlength{\tabcolsep}{2pt}
    	 \caption{ Quantitative depth evaluation results on two TUM \textit{frieburg3} RGB-D sequences. pRGBD-Refined results are after \emph{3 self-improving loops}. }
 \label{tab:depth-tum}
 \begin{tabular}{|l|c|c|c|c|c|c|c|}
\hline
\multicolumn{1}{|c|}{} & \multicolumn{7}{c|}{TUM RGBD Sequences}                                                  \\
                     & \multicolumn{4}{|c}{\cellcolor[HTML]{FFCE93}\textbf{Lower is better}} & \multicolumn{3}{c|}{\cellcolor[HTML]{CBCEFB}\textbf{Higher is better}}   \\
                     \cline{2-8}
\multicolumn{1}{|c|}{Method}               & \multicolumn{1}{c|}{Ab Rel} & \multicolumn{1}{c|}{Sq Rel} & \multicolumn{1}{c|}{RMSE}  & \multicolumn{1}{c|}{RMSElog} &\multicolumn{1}{c|}{ a1}          & \multicolumn{1}{c|}{a2}& \multicolumn{1}{c|}{a3}               \\
\hline
pRGBD-Initial/MonoDepth2-M     & 0.397	&0.848 &	1.090 &	0.719 &	0.483 &	0.722 &	0.862        \\
\hline
pRGBD-Refined    & \textbf{0.307} &\textbf{0.341}&	\textbf{0.743} &	\textbf{0.655} &	\textbf{0.522} &	\textbf{0.766} &	\textbf{0.873}    \\
\hline
\end{tabular}
\end{table}
\vspace{-0.7cm}
\begin{table}[H]
      	\centering
      	\scriptsize
    
	\setlength{\tabcolsep}{2pt}
	\vspace{-0.3cm}
	\caption{Ablation study on $1^{st}$ self-improving loop. The best performance is in \textbf{bold}}.
\label{tab:abl}
\vspace{-0.1cm}
\begin{tabular}{|l|c|c|c|c|c|c|c|}
\hline
\textbf{}                                                      & \multicolumn{4}{c|}{\cellcolor[HTML]{FFCE93}{ \textbf{Lower is better}}} & \multicolumn{3}{c|}{\cellcolor[HTML]{CBCEFB}\textbf{Higher is better}} \\ \hline
\multicolumn{1}{|l|}{{Loss}}                            & {Abs Rel}       & {Sq Rel}      & {RMSE}       & {RMSE log}      & {a1}            & {a2}           & {a3}           \\ \hline
\multicolumn{1}{|l|}{w/o $\mathcal{D}_{c}$}  &  \textbf{0.117} &	0.958	&4.956&	0.194&	0.862&	0.955 &	0.980                \\ \hline
\multicolumn{1}{|l|}{w/o $\mathcal{T}_{c}$}  & 0.118 &	0.955&	4.867&	0.194&	0.872&	0.957&	0.980             \\ \hline
\multicolumn{1}{|l|}{w/o $\mathcal{P}_{c}$}                            &      \textbf{0.117} &	0.942&	4.855&	0.194&	\textbf{0.873}&	\textbf{0.958}&	0.980    \\ \hline         
\multicolumn{1}{|l|}{all losses}                            &       \textbf{ 0.117} &\textbf{	0.931} &	\textbf{4.809} &	\textbf{0.192} &	\textbf{0.873} &	\textbf{0.958} &	\textbf{0.981}      \\ \hline
\end{tabular}
\end{table}

\section{Analysis of Self-Improving Loops}
\label{sec:self-loop-analysis}

\begin{figure}

\vspace{-0.8cm}
\centering
\setlength{\tabcolsep}{-1pt}
\begin{tabular}{c@{\hspace{0.1cm}}c@{\hspace{0.1cm}}c@{\hspace{0.1cm}}c}
\includegraphics[width=3cm]{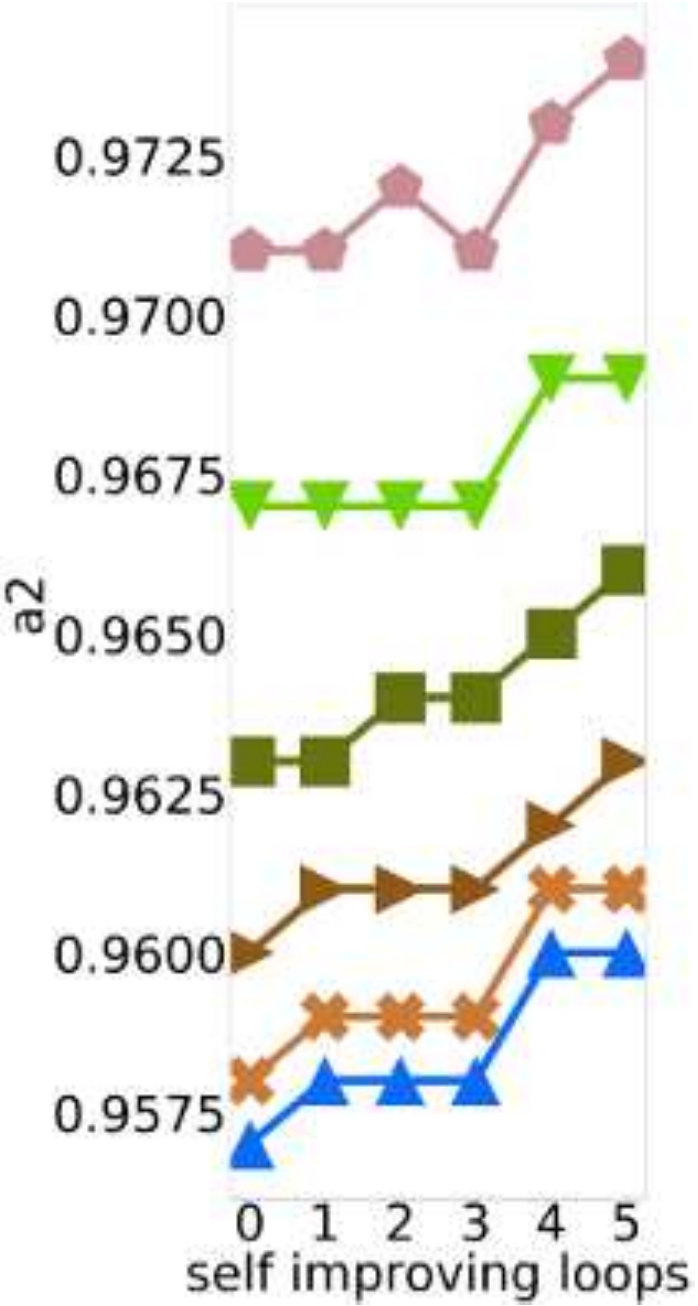} &
\includegraphics[width=2.7cm]{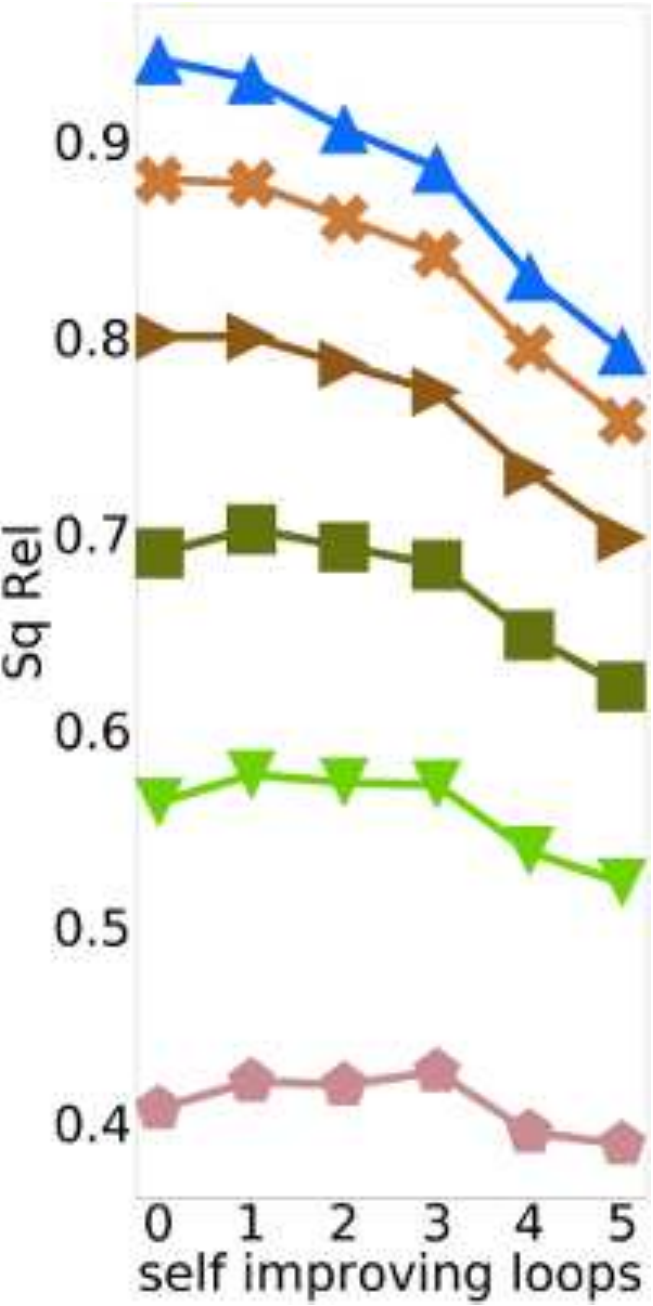}&
\includegraphics[width=2.7cm]{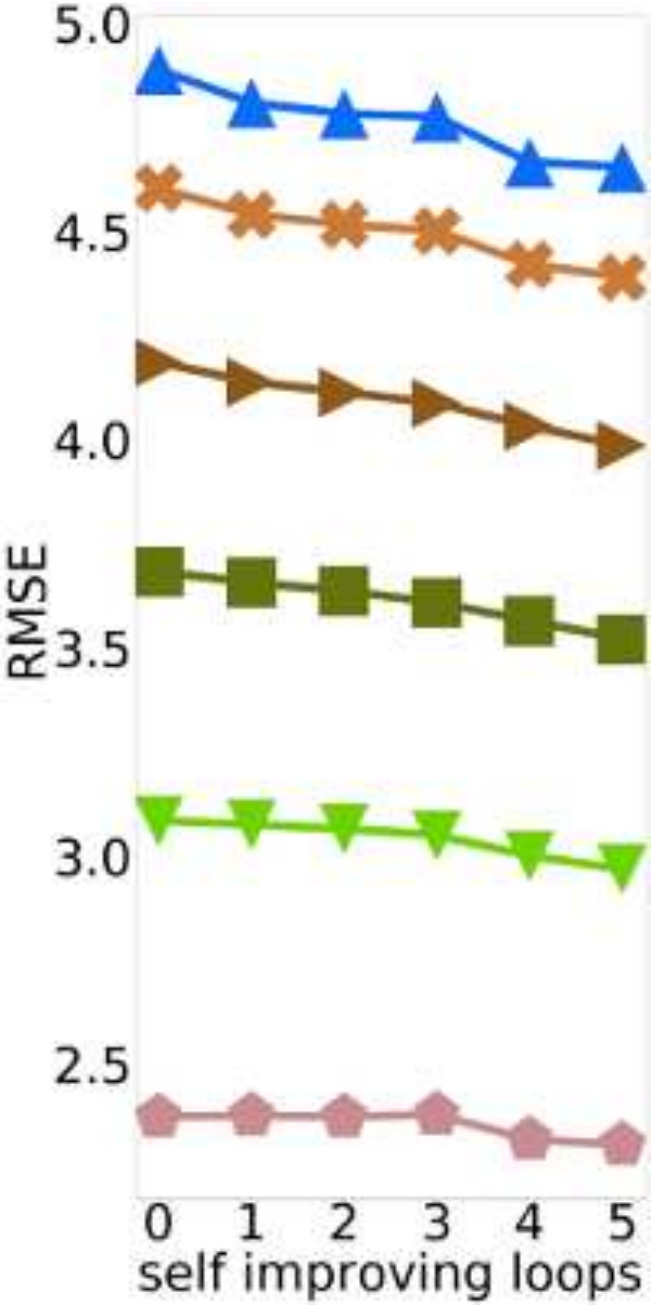}&
\includegraphics[width=3cm]{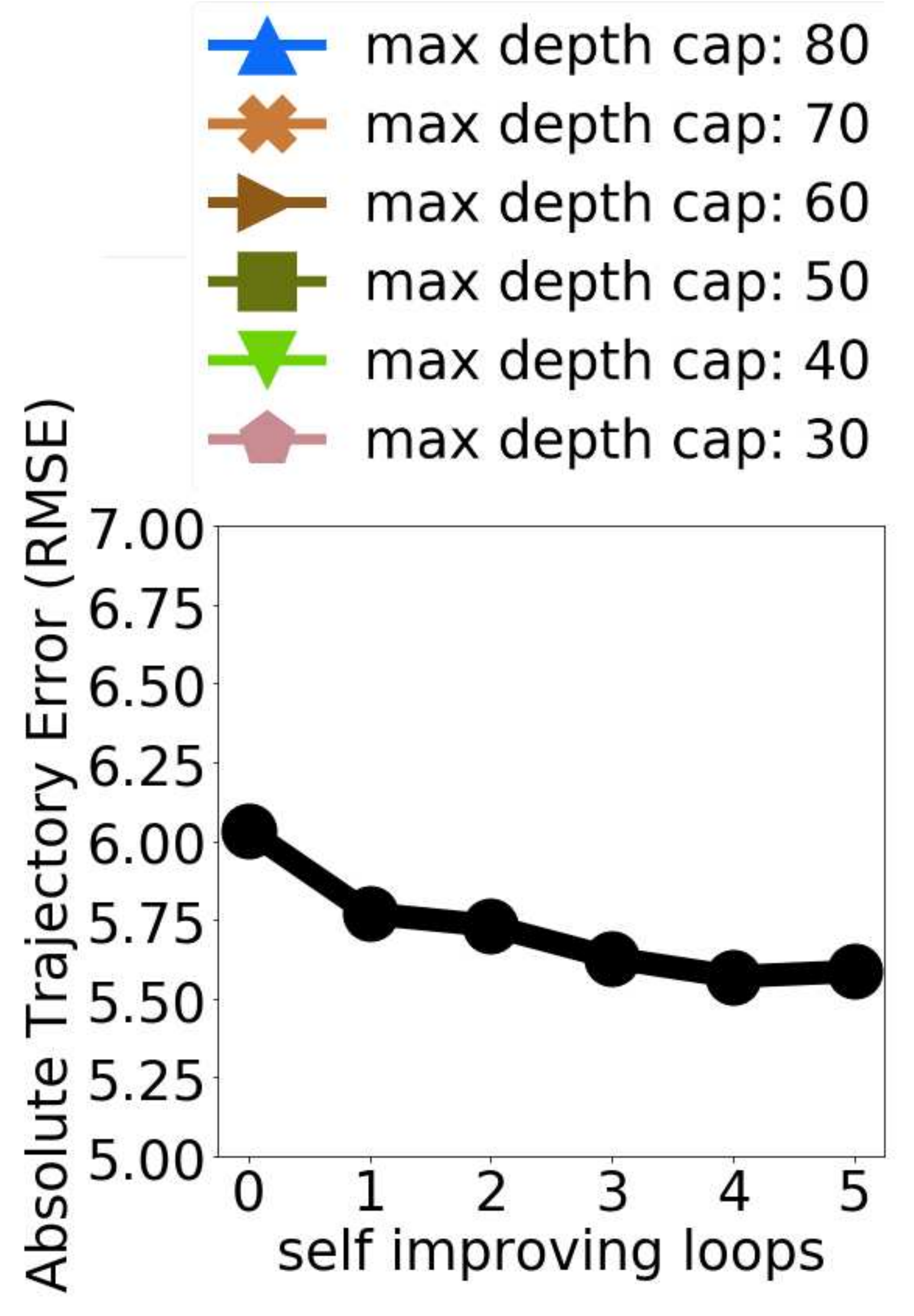}
\\
(a) & (b)&(c) & (d)\\
\end{tabular}
\vspace{-0.3cm}
\setcounter{figure}{5} 
\caption{Depth/pose evaluation metric w.r.t. self-improving loops. Depth evaluation metrics in (a-c) are computed at different max depth caps ranging from 30-80 meters.}
\label{fig:self-loop-analysis}
\vspace{-0.7cm}
\end{figure}

In this section, we analyze the behaviour of three different evaluation metrics for depth estimation: Squared Relative (Sq Rel) error, RMSE error and accuracy metric a2, as defined in Sec. \ref{sec:experiments}. The pose estimation is evaluated using the  absolute trajectory pose error. In Fig.~\ref{fig:self-loop-analysis}, we use the KITTI Eigen split dataset and report these metrics for each iteration of the self-improving loop. 
The evaluation metrics corresponding to the $0^{th}$ self-improving loop are of the pre-trained MonoDepth2-M. We summarize the findings from the plots in Fig.~\ref{fig:self-loop-analysis} as below: 
\begin{itemize}
    \item A comparison of evaluation metrics of farther scene points (\eg max depth 80) with nearby points (\eg max depth 30) at the $0^{th}$ self-improving loop shows that the pre-trained MonoDepth2 performs poorly for farther scene points compared to nearby points. 
    \item In the subsequent self-improving loops, we can see the rate of reduction in the Sq Rel and RMSE error is significant for  farther away points compared to nearby points, \eg,~slope of error curves in Fig.~\ref{fig:self-loop-analysis}(b-c) corresponding to max depth 80 is steeper than that of max depth 30. This validates our hypothesis of including wider baseline losses that help the depth network predict more accurate depth values for farther points. Overall, our joint narrow and wide baseline based learning setup helps improve the depth prediction of both the nearby and farther away points, and outperforms MonoDepth2~\cite{godard2019digging}.
    \item The error plot in Fig. \ref{fig:self-loop-analysis}(d) shows a decrease in pose error with self-improving loops and complements the improvement in depth evaluation metrics as shown in Fig.\ref{fig:self-loop-analysis}(a)-(c). We terminate the self-improvement loop once there is no furhter improvement, \ie, at the $5^{th}$ iteration.
\end{itemize}

\vspace{-0.3cm}
\section{Conclusion}
\label{sec:conclusion}
\vspace{-0.1cm}
In this work, we propose a self-improving framework to couple geometrical and learning based methods for 3D perception. A win-win situation is achieved --- both the monocular SLAM and depth prediction are improved by a significant margin without any additional active depth sensor or ground truth label. 
Currently, our self-improving framework only works in an off-line mode, so developing an on-line real-time self-improving system remains one of our future works. Another avenue for our future works is to move towards more challenging settings, e.g., uncalibrated cameras~\cite{zhuang2019degeneracy} or rolling shutter cameras~\cite{zhuang2019learning}.


\section*{Acknowledgement}
This work was part of L. Tiwari's internship at NEC Labs America, in San Jose. L. Tiwari was supported by Visvesvarya Ph.D. Fellowship. L. Tiwari and S. Anand were also supported by Infosys Center for Artificial Intelligence, IIIT-Delhi. 
\appendix

\section{ Supplementary Material}
This supplementary material is organized as follows. 
We first present depth refinement results on KITTI Odometry sequences in Sec. \ref{sec:depth-ref-odo}. Next, we give a comparison of our pose refinement with state-of-the-art RGB SLAM approaches in Sec. \ref{sec:pose-ref-odo}. We further evaluate pose refinement on KITTI Leaderboard in Sec. \ref{sec:leaderboard}. Additional implementation details and qualitative results of TUM RGB-D experiments are included in Sec. \ref{sec:tum-implementation}. Additional analysis of the self-improving loop with all the 7 depth evaluation metrics is presented in Sec. \ref{sec:additional-plots-analysis}. Some additional qualitative depth evaluation results of KITTI Eigen experiments and pose evaluation results of KITTI Odometry experiments are presented in Sec. \ref{sec:more-results-qual-depth} and Sec. \ref{sec:add-pose_qual} respectively. Finally, we provide some demo videos on KITTI Odometry and TUM RGB-D sequences in Sec.~\ref{sec:video}.


\subsection{Depth Refinement Evaluation on KITTI Odometry}
\label{sec:depth-ref-odo}
We evaluate the depth refinement step of our self-improving pipeline on KITTI Odometry sequences 09 and 10. The first block (\ie MonoDepth2-M vs pRGBD-Refined) of the Tab. \ref{tab:depth_odo_dcnf} shows the improved results after the depth refinement step. We also compare our method with a state-of-the-art depth refinement method DCNF~\cite{yin2017scale}. Note: DCNF~\cite{yin2017scale} uses 
\emph{ground-truth} depths for pre-training the network, while our method uses only \emph{unlabelled} monocular images, and still outperforms DCNF (see second block of the Tab. \ref{tab:depth_odo_dcnf}). The result shows that our self-improving framework with the wide-baseline losses (\ie, symmetric depth transfer and depth consistency losses) improves the depth prediction.
\begin{table}[!h]
\vspace{-0.4cm}
    \centering
    \scriptsize
    \caption{Qualitative depth evaluation on KITTI Odometry sequences 09 and 10. M: self-supervised monocular supervision for fine-tuning. `-' means the result is not available from the paper. Our results are after \emph{5 self-improving loops}. Note: DCNF~\cite{yin2017scale} uses \emph{ground-truth} depths for pre-training.  Best results in each block is in \textbf{bold}.}
    \label{tab:depth_odo_dcnf}
\begin{tabular}{|l|c|c|c|c|c|c|c|c|c|}
\hline
                        &  & Depth & \multicolumn{4}{c}{\cellcolor[HTML]{FFCE93}\textbf{Lower is better}}    & \multicolumn{3}{c|}{\cellcolor[HTML]{CBCEFB}{\color[HTML]{000000} \textbf{Higher is better}}}                                                               \\
 Method                                 &   Train & Cap   & {Abs Rel} & {Sq Rel} & {RMSE}  & {RMSE $\text{log}_{2}$} & {a1} & {a2} &{ a3} \\
                                \hline
MonoDepth2-M~\cite{godard2019digging}              & M    &80 &  0.123 &	0.703 &	4.165 &	0.188&	0.854	&0.956	&0.985                                                 \\
pRGBD-Refined           & M   & 80 & \textbf{0.121}&	\textbf{0.649}	&\textbf{3.995}&	\textbf{0.184}&	\textbf{0.853}&	\textbf{0.960}&	\textbf{0.986}  \\
\hline 
                                DCNF~\cite{yin2017scale}              & M    & 20 &  0.112 &	- &	2.047 &	-  &	-	&-	&-   \\
pRGBD-Refined           & M   &20 &\textbf{0.098 }&	\textbf{0.242} &	\textbf{1.610} &	\textbf{0.145} 	& \textbf{0.906} &	\textbf{0.978} &	\textbf{0.993}  \\
\hline
\end{tabular}
\vspace{-0.7cm}
\end{table}

\subsection{Comparison with State-Of-The-Art SLAM Methods}
\label{sec:pose-ref-odo}
In this section, we compare our pRGBD-Initial and pRGBD-Refined methods against state-of-the-art RGB SLAM methods, \ie, Direct Sparse Odometry (DSO)~\cite{engel2017direct}, Direct Sparse Odometry with Loop Closure (LDSO)~\cite{gao2018ldso}, and Direct Sparse Odometry in Dynamic Environments (DSOD)~\cite{ma2019dsod}. The results are shown in Tab.~\ref{tab:pose-odometry-baselines}. From the results, it is evident that our pRGBD-Refined outperforms all the competing methods in Absolute Trajectory Error (RMSE) and Relative Translation (Rel Tr) Error . While the improvement in Absolute Trajectory Error (RMSE) and Relative Translation (Rel Tr) error is substantial, the performance in Relative Rotation (Rel Rot) is not comparable. The higher Rel Rot errors of our method compared to other RGB ORB-SLAM methods could be due to the high uncertainty of CNN-predicted depths for far-away points, which affects our rotation estimation~\cite{hartley2003multiple}. However, if we compare Rel Rot error of pRGBD-Initial with the pRGBD-Refined, as depth prediction improves (see Tab. \ref{tab:depth_odo_dcnf} MonoDepth2-M/pRGBD-Initial vs pRGBD-Refined) the Rel Rot error also improves (see Tab. \ref{tab:pose-odometry-baselines} ).
    	\begin{table}[]
    	    \centering
    	  
    	 \caption{Comparison with state-of-the-art RGB SLAM methods on KITTI Odometry sequences 09 and 10. Here, ‘-’ means the result is not available from the original paper. $*$ denotes the result is obtained from~\cite{ma2019dsod}.}
 \label{tab:pose-odometry-baselines}
 \begin{tabular}{|l|ccc|ccc|}
 \hline
  & & Seq. 09 & & &
 Seq. 10 &
 \\
 Method & RMSE & Rel Tr & Rel Rot & RMSE & Rel Tr & Rel Rot \\
 \hline
 RGB ORB-SLAM\cite{mur2017orb}
 &18.34         &  7.42               & {\ul0.004} &   8.90          & 5.85            & {\ul0.004} \\
 DSO\cite{engel2017direct}
 &   74.29         &     $72.27^{*}$         & $\textbf{0.002}^{*}$ &   16.32          &  $80.81^{*}$            & $\textbf{0.002}^{*}$  \\
 LDSO\cite{gao2018ldso}
 &  21.64         &  -            & - &   17.36          &  -            & -  \\
 DSOD\cite{ma2019dsod}
 &  -         &  13.85            & \textbf{0.002} &   -          &  13.53            & \textbf{0.002}  \\
pRGBD-Initial
 &   {\ul12.21}         &  {\ul4.26}            & 0.011 &   {\ul8.30}          &  {\ul5.55}            & 0.017  \\
 pRGBD-Refined
 &  \textbf{11.97}   & \textbf{4.20}   & 0.010  &  \textbf{6.35} &  \textbf{4.40}  &  0.016 \\
 \hline
 \end{tabular}
 \vspace{-0.7cm}
 \end{table}

\subsection{KITTI Odometry Leaderboard Results}
\label{sec:leaderboard}
In the main paper, we keep the default setting from ORB-SLAM, which leads to tracking failures of all methods in a few sequences (i.e., see Tab.~3 of the main paper). The KITTI Odometry leaderboard requires the results of all sequences (i.e., sequences 11-21) for evaluation. Therefore, we increase the minimum number of inliers for adding keyframes from 100 to 500 so that our pRGBD-Refined succeeds on all sequences. We report the results of our pRGBD-Refined on the KITTI Odometry leaderboard in Tab.~\ref{tab:leaderboard}. Results show our method outperforms the competing monocular/LiDAR-based methods both in terms of relative translation and rotation errors.

\begin{table}[!htb]
\centering
\caption{Quantitative pose evaluation results on KITTI Odometry leaderboard. Note that we use the estimated trajectories from ORB-SLAM2-S~\cite{mur2017orb} for global scale alignment. The best performance is in \textbf{bold}.}
\begin{tabular}{lcc}
\hline 
Method            & Rel Tr & Rel Rot \\
\hline
ORB-SLAM2-S~\cite{mur2017orb} & 1.70   & 0.0028  \\
\hline
OABA~\cite{frost2016object}             & 20.95  & 0.0135  \\
VISO2-M~\cite{geiger2011stereoscan}          & 11.94  & 0.0234  \\
BLO~\cite{velas2018cnn}             & 9.21  & 0.0163  \\
VISO2-M+GP~\cite{geiger2011stereoscan,song2014robust}        & 7.46   & 0.0245  \\
pRGBD-Refined              & \textbf{6.24}   & \textbf{0.0097}  \\
\hline
\end{tabular}
\label{tab:leaderboard}
\vspace{-0.4cm}
\end{table}
       	
\subsection{Experiments on TUM RGB-D Sequences}
\label{sec:tum-implementation}
\subsubsection{Implementation Details} 

We pre-train/fine-tune the depth network on image resolution $480 \times 320$. For pre-training, we set the learning rate to $10^{-4}$ initially, reduce it to $10^{-5}$ after 20 epochs, and train for 30 epochs. For fine-tuning, we extract camera poses, 2D keypoints and the associated depths from keyframes while running RGB-D ORB-SLAM on the training sequences. We fine-tune the depth network with the fixed learning rate of $10^{-6}$. We use the following 6 sequences for pre-training/fine-tuning: 1. {fr3/long\_office\_household}, 2. {fr3/long\_office\_household\_validation}, 3. {fr3/sitting\_xyz}, 4. {fr3/structure\_texture\_far}, 5. {fr3/structure\_texture\_near}, 6. {fr3/teddy}, and the following 2 sequences for testing: 1. {fr3/walking\_xyz}, 2. {fr3/large\_cabinet\_validation}. Note that these are the only 8 sequences with provided {rectified} images among the entire TUM RGB-D dataset.

\subsubsection{Qualitative Results}
\begin{figure*}[h]
\vspace{-0.6cm}
\centering
\setlength{\tabcolsep}{0.5pt}
\begin{tabular}{cc}
\includegraphics[width=5cm]{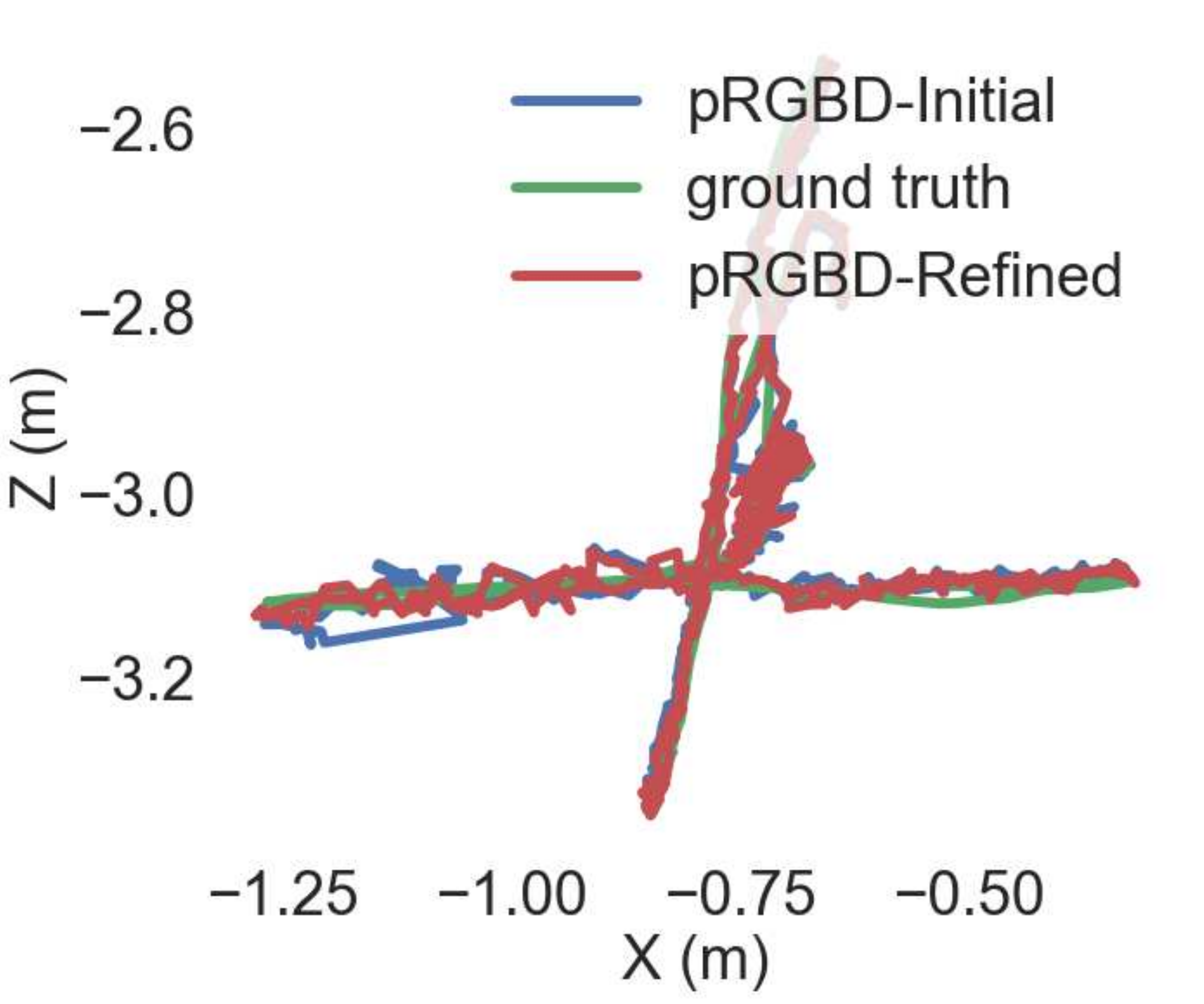} &
 \includegraphics[width=5cm]{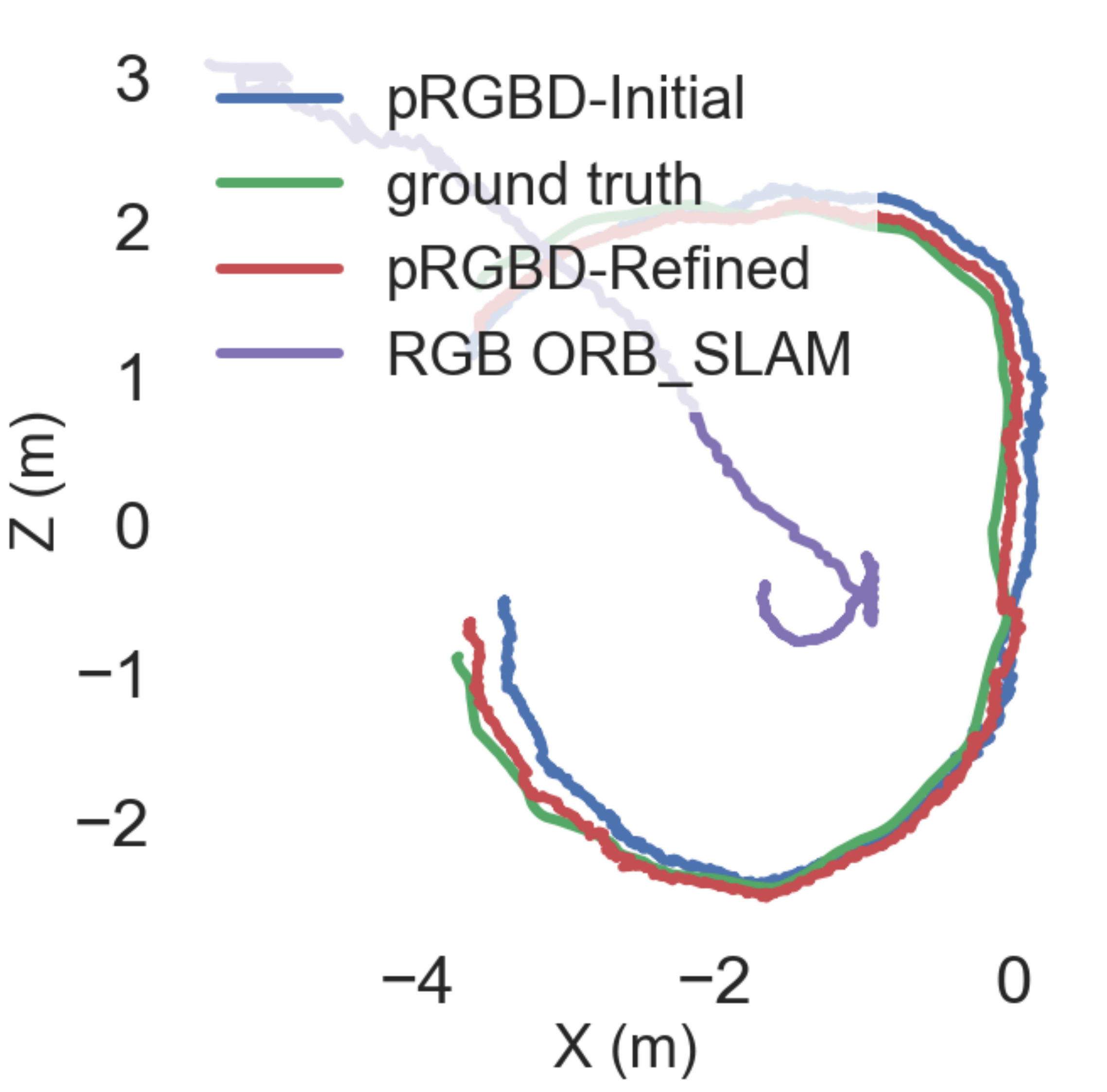}  \vspace{-0.15cm}\\
{\scriptsize (a) fr3/walking\_xyz}   & {\scriptsize (b)fr3/large\_cabinet\_validation}  \\
\end{tabular}
\vspace{-0.3cm}
\caption{Qualitative pose evaluation results on TUM RGB-D sequences. Note that RGB ORB-SLAM fails in (a).}
\label{figtab:pose-tum}
\vspace{-0.5cm}
\end{figure*}

\begin{figure*}[h]
\centering
\setlength{\tabcolsep}{0.5pt}
\begin{tabular}{ccc}
\includegraphics[width=3cm]{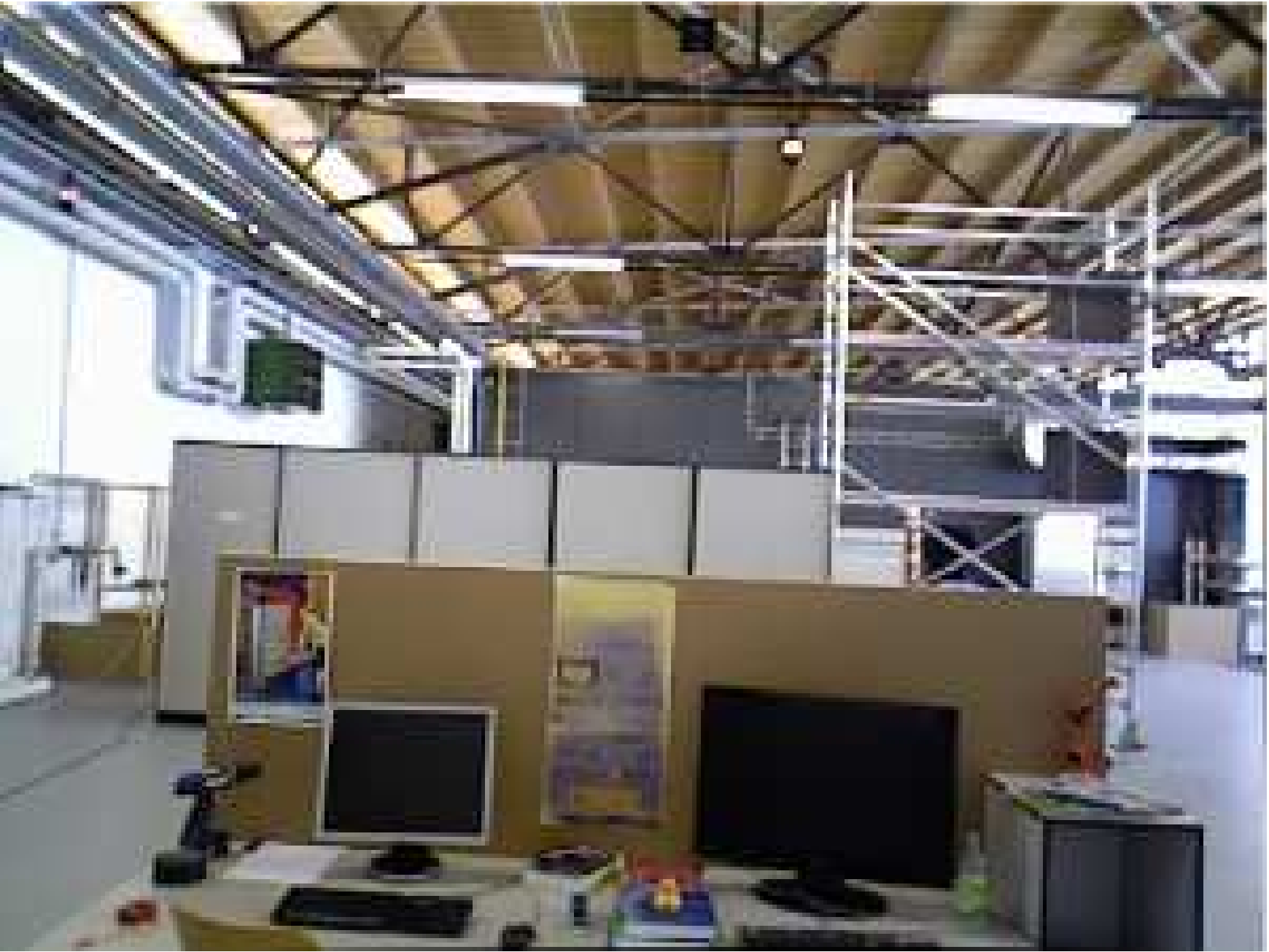} & 
\includegraphics[width=3.5cm]{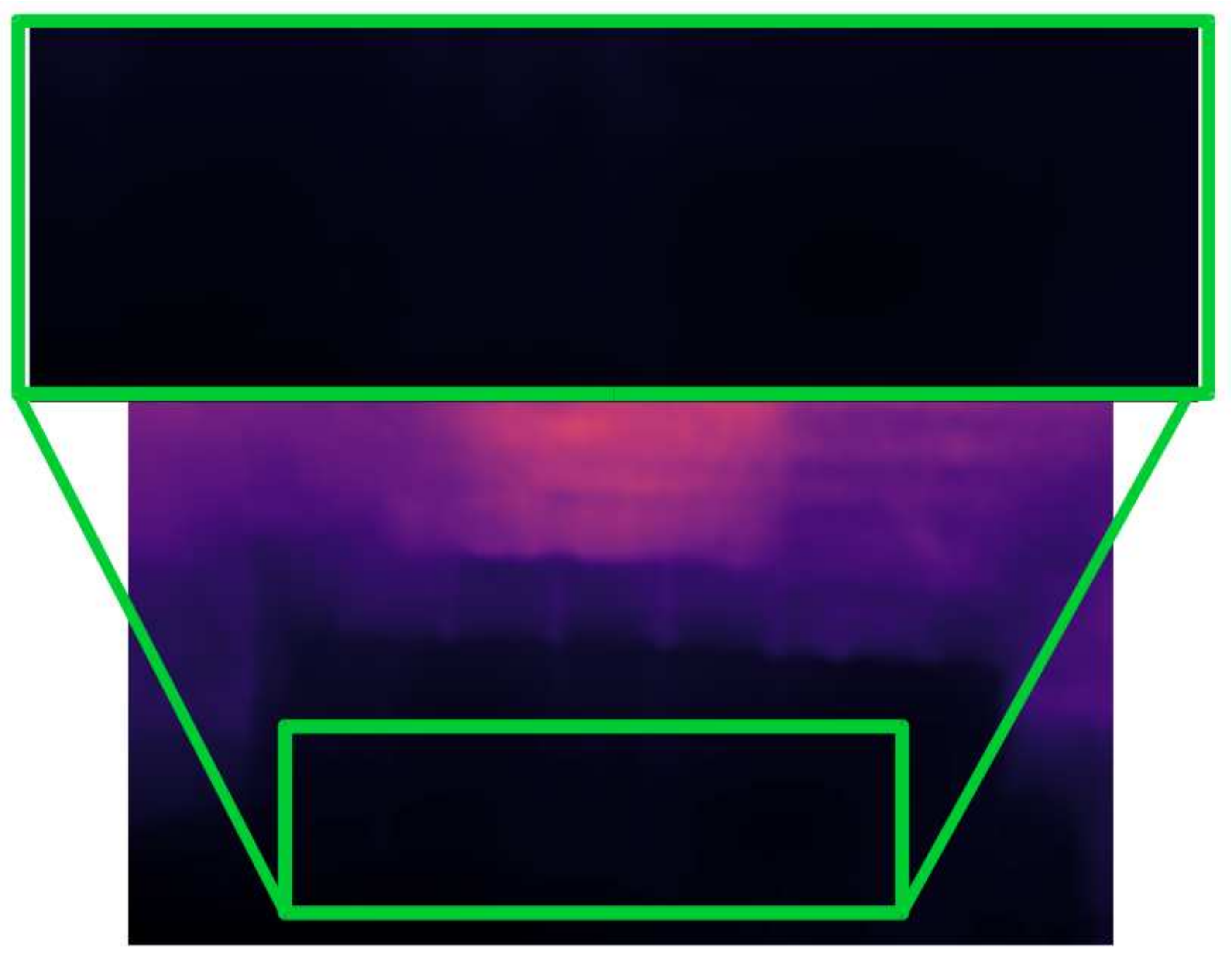} &
\includegraphics[width=3.5cm]{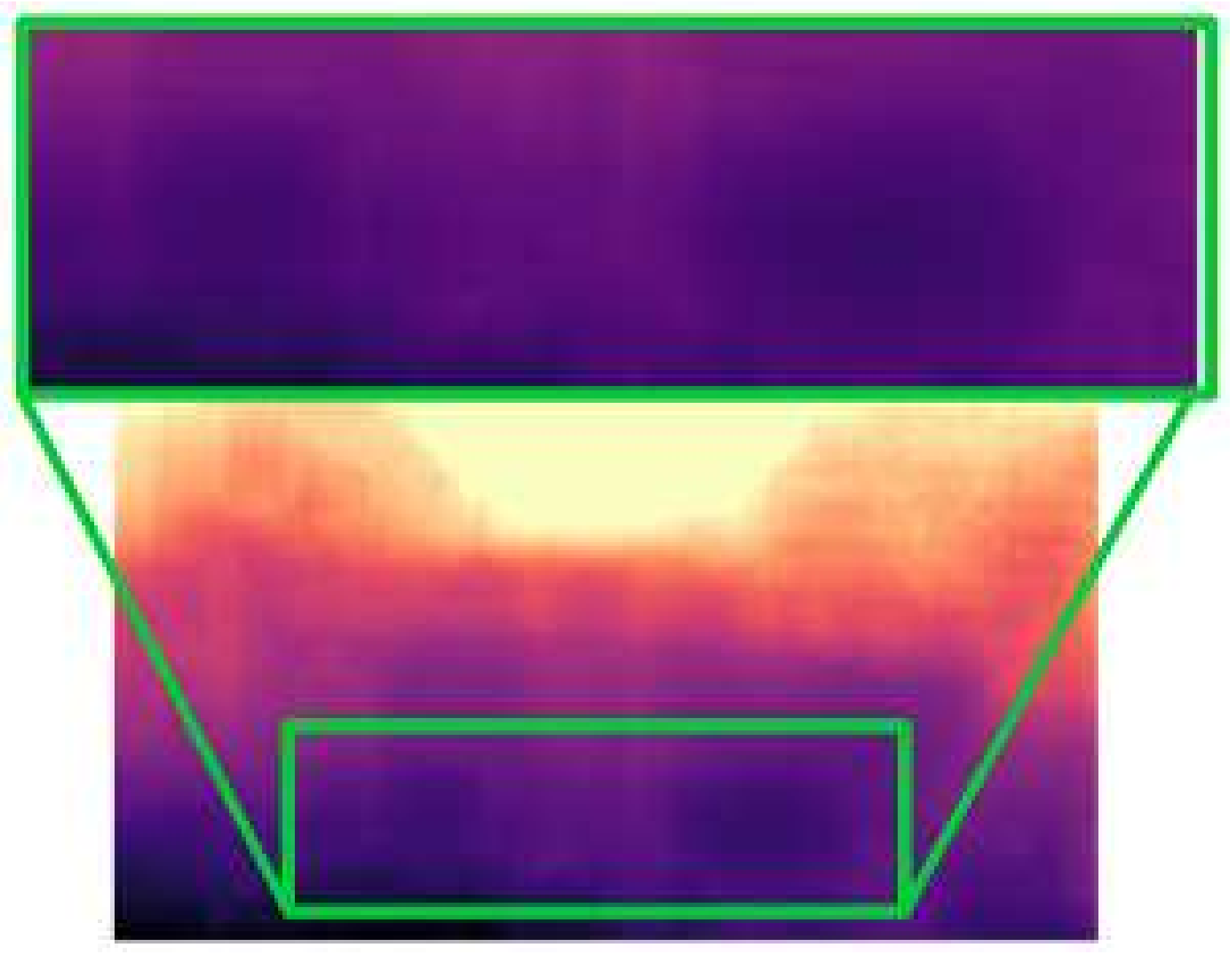} \\
{\scriptsize (a) RGB }   & {\scriptsize  (b) pRGBD-Initial} & {\scriptsize  (c) pRGBD-Refined}  \\
\end{tabular}
\vspace{-0.3cm}
\caption{Qualitative depth evaluation results on TUM RGB-D sequences. }
 \label{figtab:depth-tum}
\end{figure*}

Fig. \ref{figtab:pose-tum}(a) and Fig. \ref{figtab:pose-tum}(b) shows qualitative pose evaluation results on test sequences \emph{walking\_xyz} and \emph{large\_cabinet\_validation} respectively.
The results, show the increased robustness and accuracy by pRGBD-Refined.   In  particular,  RGB ORB-SLAM  fails  on walking\_xyz, while pRGBD-Refined succeeds and achieves the best performance on both sequences. Some qualitative depth refinement results are presented in Fig. \ref{figtab:depth-tum}. It can be seen that the disparity between the depth values of nearby and farther scene points become clearer, \eg, see depth around the two monitors.

\subsection{Additional Plots of Self-Improving Loop Analysis}
\label{sec:additional-plots-analysis}
In the main paper, we have shown behaviours of 3 depth evaluation metrics named as (Sq. Rel), (RMSE) and (a2). In this section we present behaviours of all $7$ metrics and pose evaluation metrics. Our analysis in the Sec. 5 of the main paper holds true with respect to all the 7 depth evaluation metrics.
\vspace{-0.7cm}
\begin{figure}[H]
    \centering
\setlength{\tabcolsep}{-1pt}
\begin{tabular}{c@{\hspace{0.1cm}}c@{\hspace{0.1cm}}c@{\hspace{0.1cm}}c}
\includegraphics[width=0.24\linewidth]{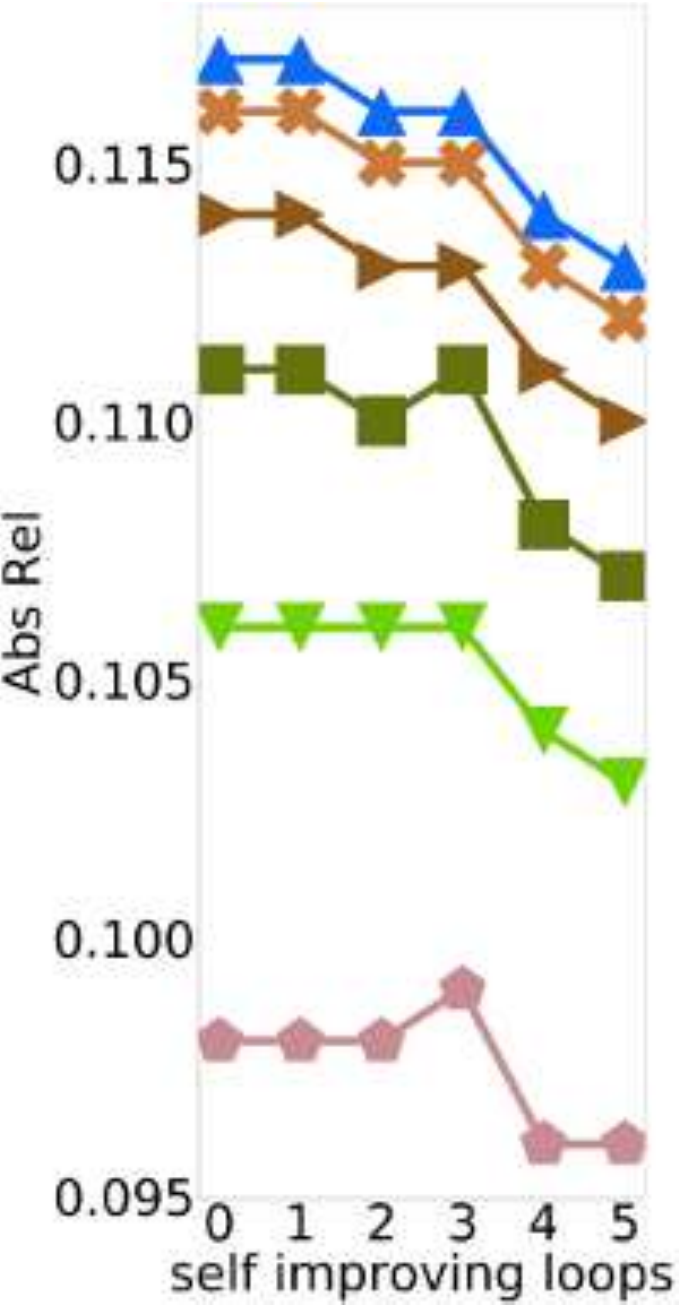} &
\includegraphics[width=0.23\linewidth]{sq_rel_new.pdf}&
\includegraphics[width=0.23\linewidth]{rmse_new.pdf}&
\includegraphics[width=0.24\linewidth]{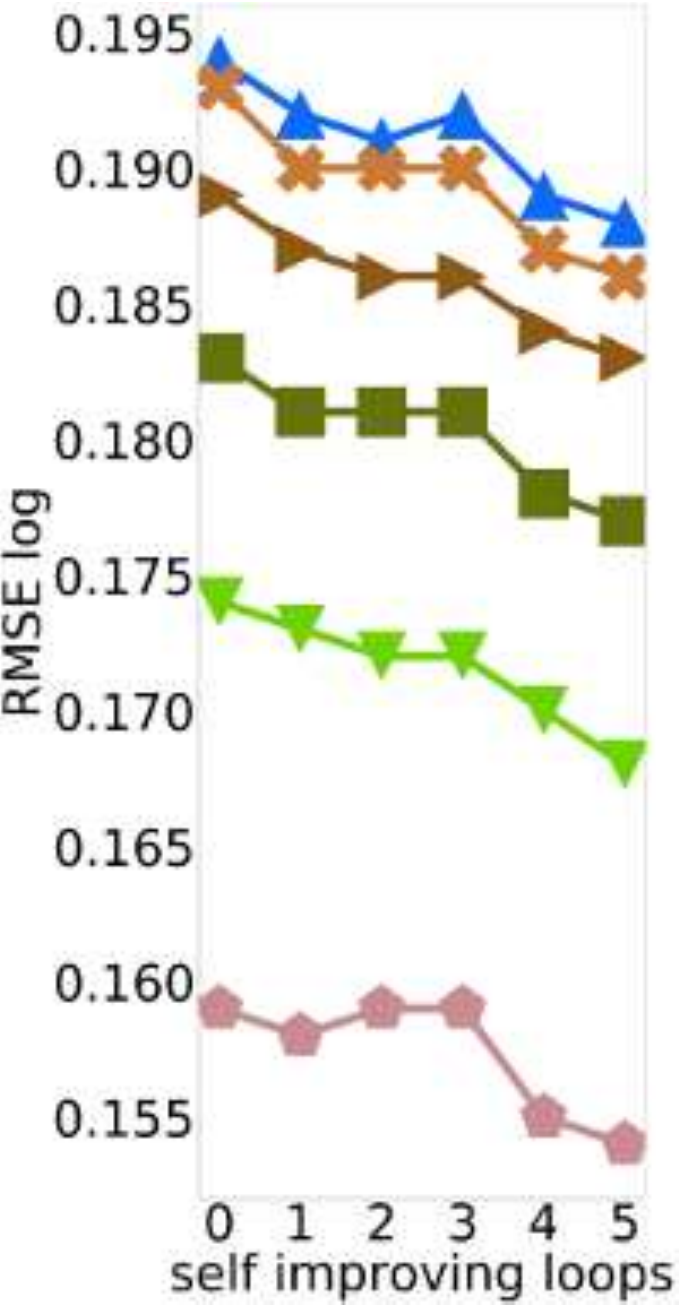} 
\\
(a) & (b)&(c) & (d)\\
\includegraphics[width=0.23\linewidth]{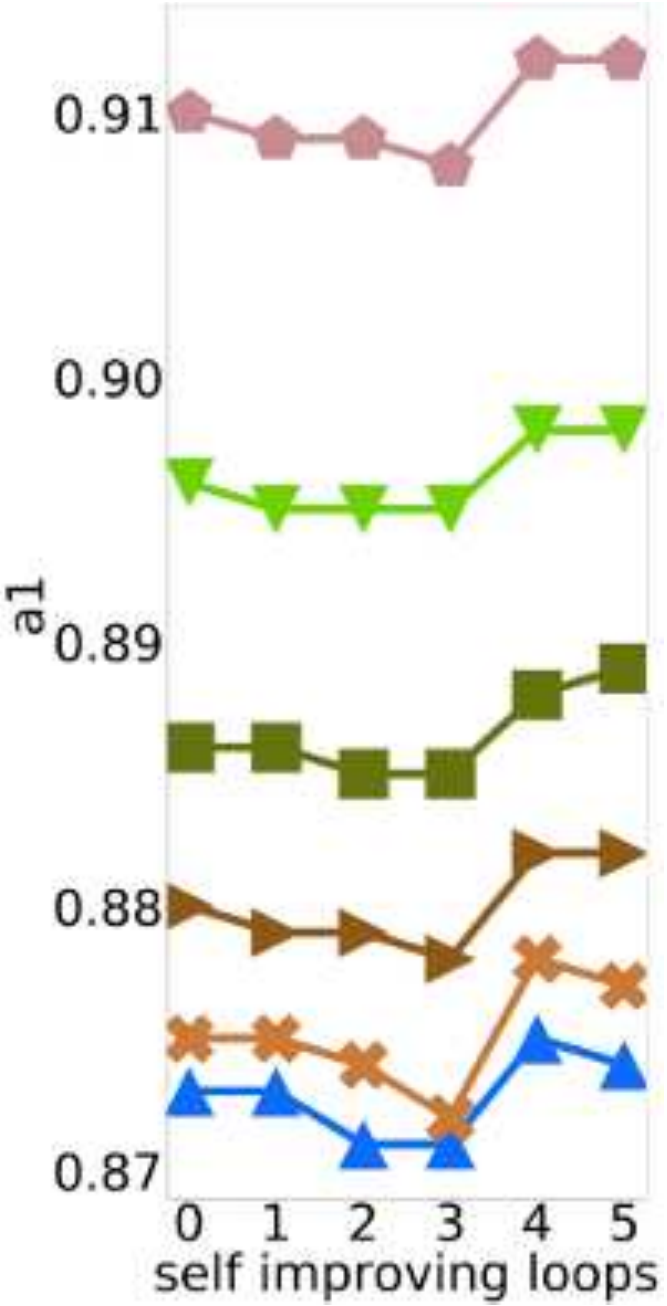}&
\includegraphics[width=0.24\linewidth]{a2_new.pdf}&
\includegraphics[width=0.23\linewidth]{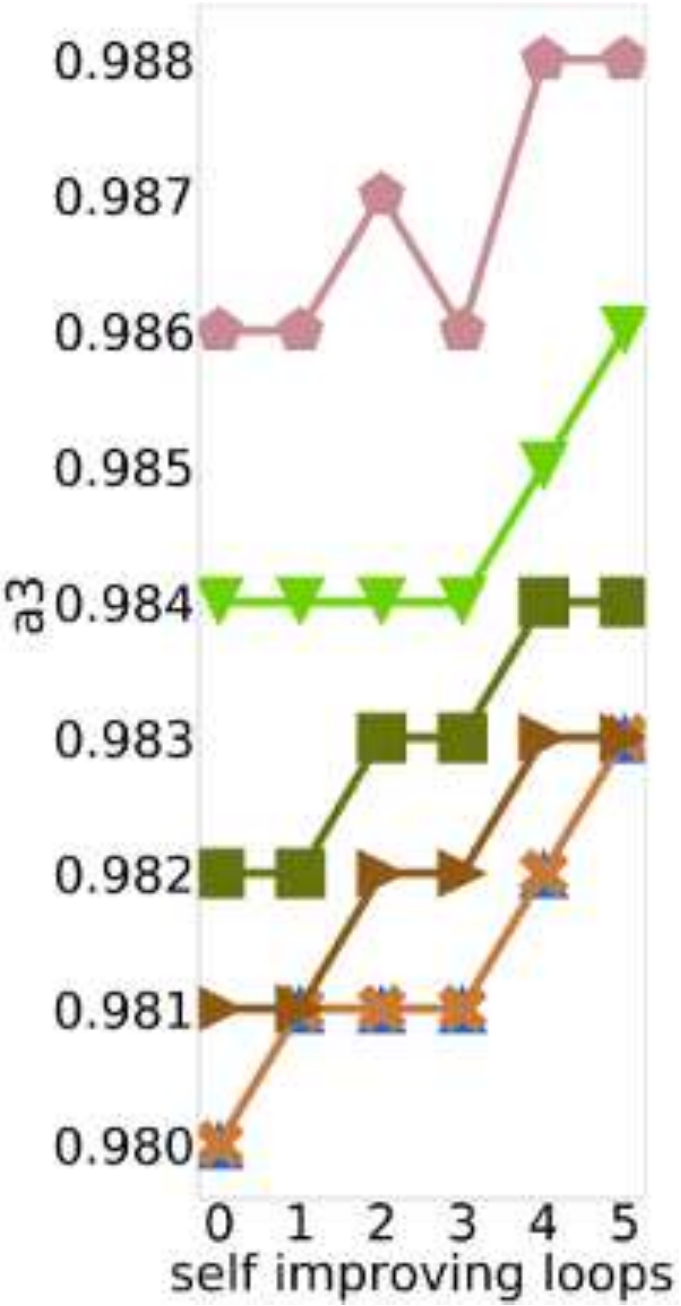}&
\includegraphics[width=0.28\linewidth]{pose_legend_new2.pdf}
\\
(e) & (f)& (g) & (h)\\
\end{tabular}
\caption{Depth/Pose evaluation metrics w.r.t. self-improving loops. (a). Absolute Relative (Abs Rel) (\textit{lower is better}) (b). Squared Relative (Sq Rel) (\textit{lower is better}) (c). RMSE (\textit{lower is better}) (d). RMSE Log (\textit{lower is better}), (e). a1 (\textit{higher is better}), (f). a2 (\textit{higher is better}), (g). a3 (\textit{higher is better}) and (h) Absolute Trajectory Error (RMSE) (\textit{lower is better}). Depth evaluation metrics in (a-g) are computed at different max depth caps ranging from 30-80 meters.}
\label{fig:self-loop-analysis-add}
\end{figure}

\subsection{Additional Depth Refinement Qualitative Results}
\label{sec:more-results-qual-depth}
\begin{figure*}[!htb]
\centering
\setlength{\tabcolsep}{0.05pt}
\begin{tabular}{cccc}
\includegraphics[width=4.1cm]{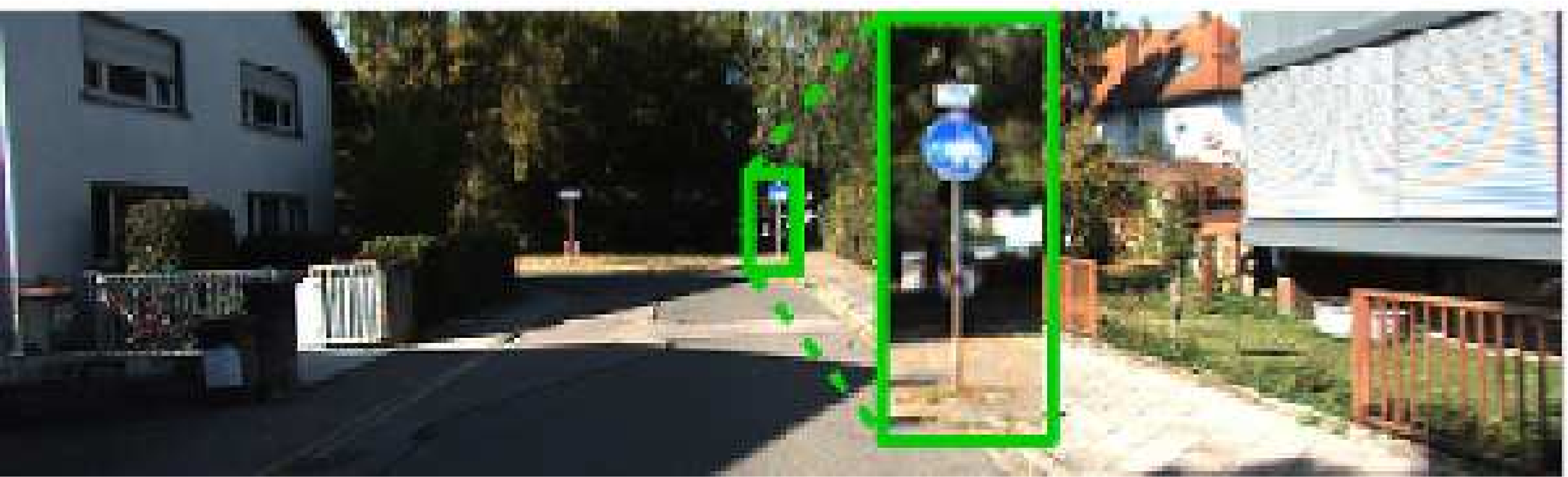} &
\includegraphics[width=4.1cm]{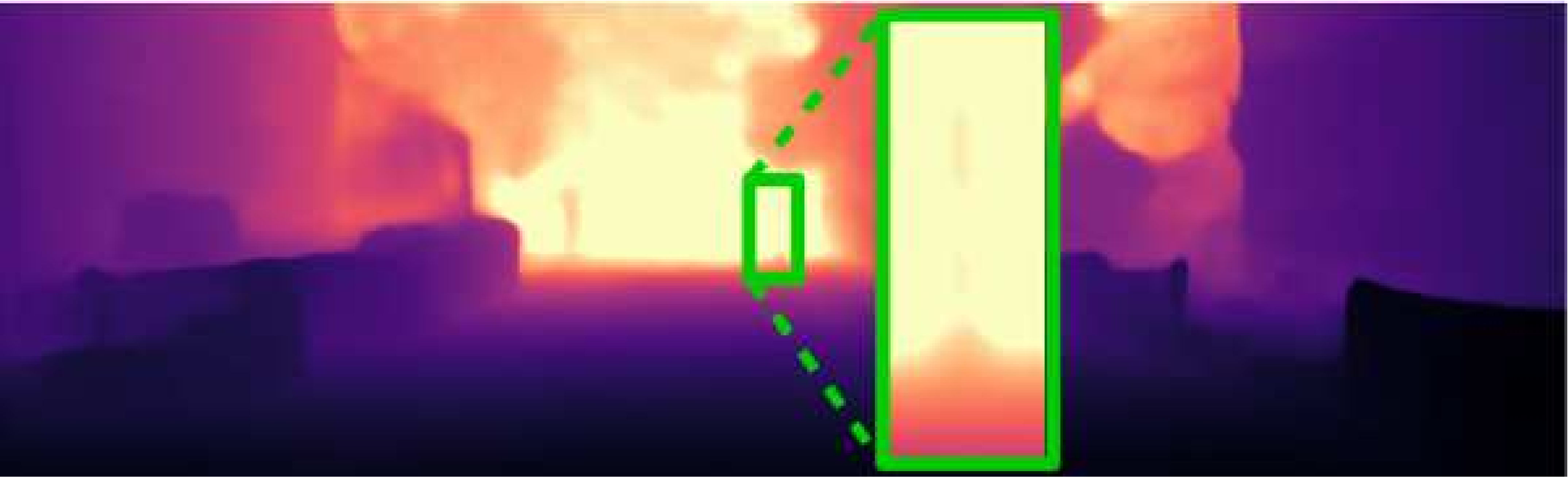} &
\includegraphics[width=4.1cm]{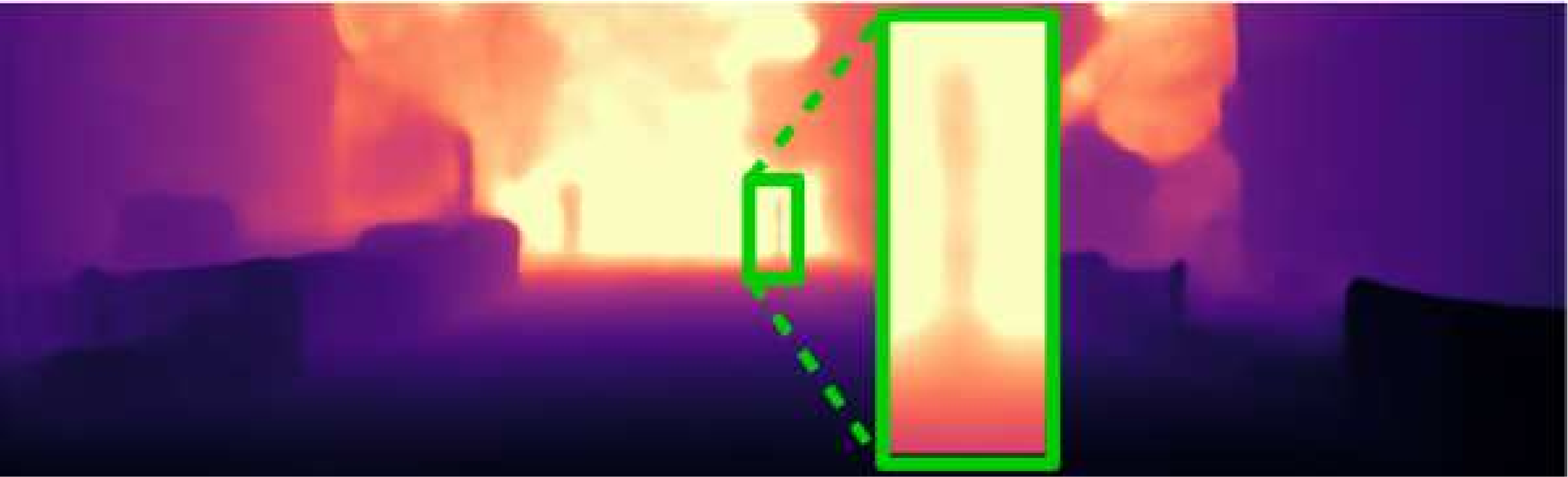} 
\vspace{-0.1cm}\\ 
\includegraphics[width=4.1cm]{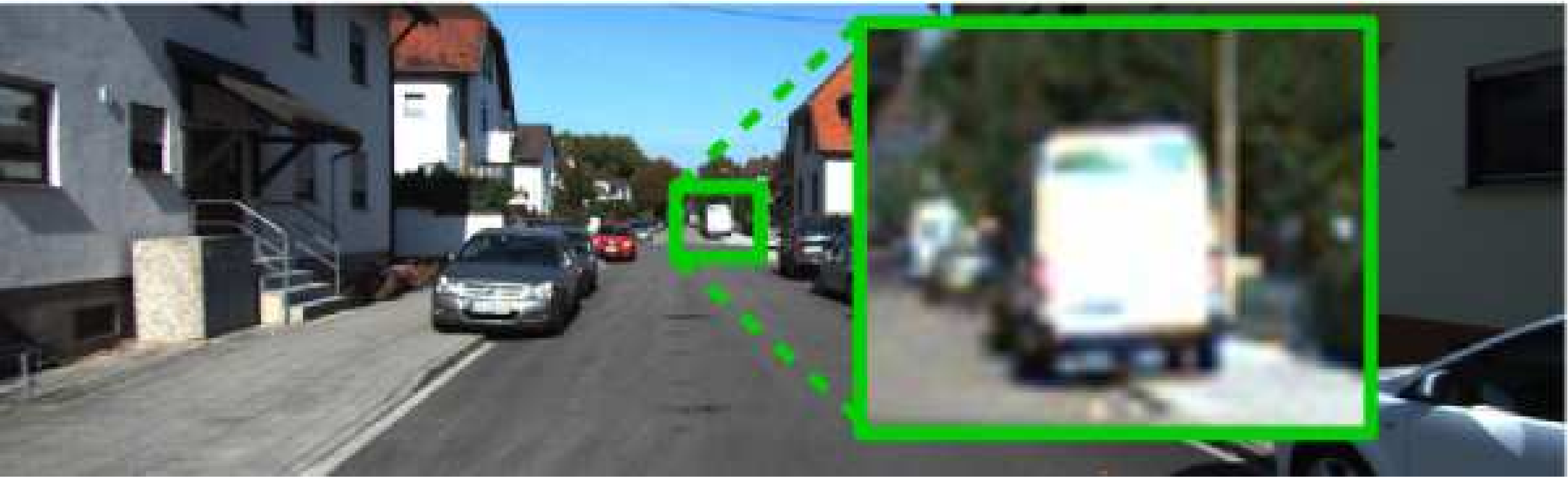} &
\includegraphics[width=4.1cm]{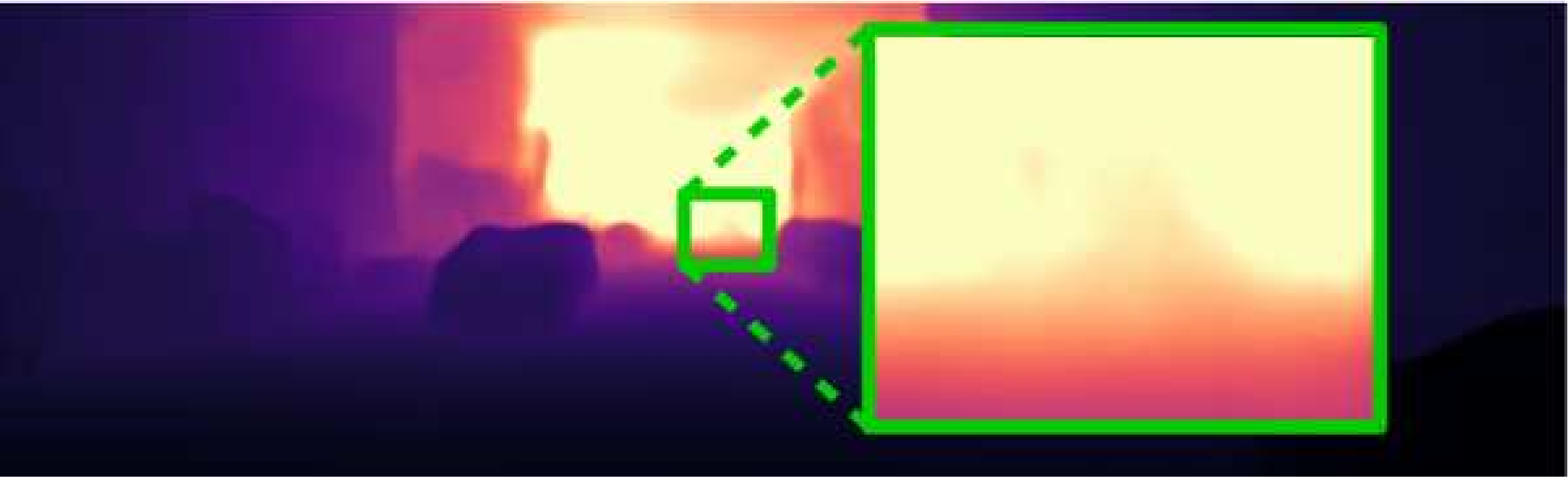} &
\includegraphics[width=4.1cm]{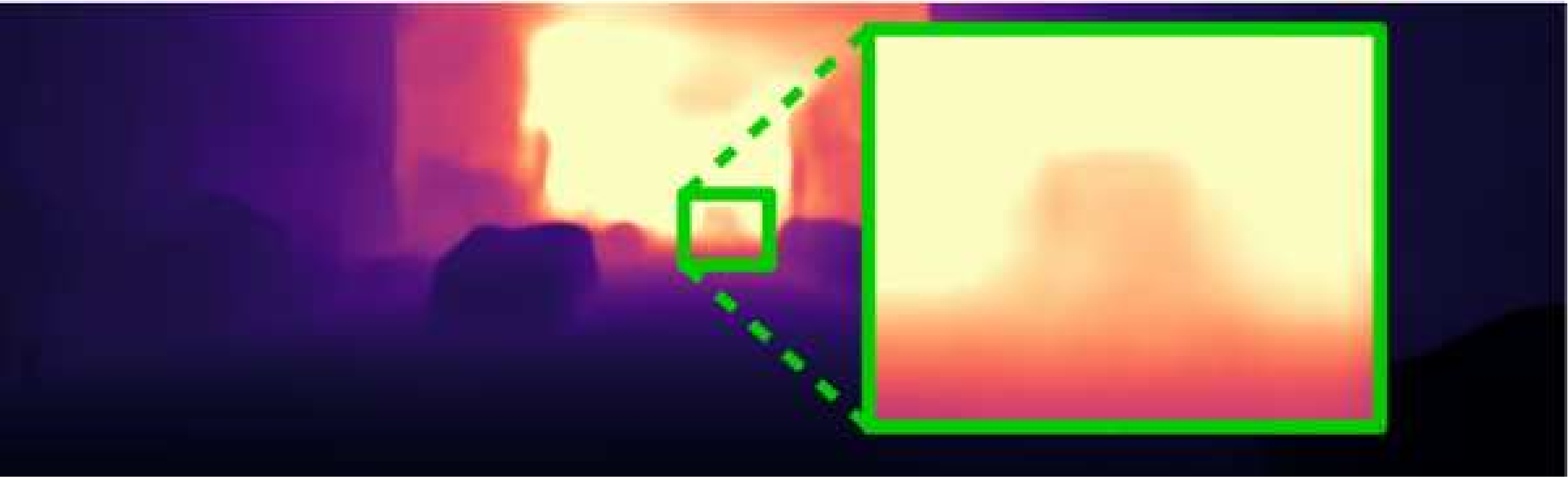} 
\vspace{-0.1cm} \\
{\scriptsize RGB}   & {\scriptsize Monodepth2-M} & {\scriptsize pRGBD-Refined} \\
\end{tabular}
\vspace{-0.3cm}
\caption{Qualitative depth evaluation results on KITTI Odometry test set. Improvement in depth prediction of farther scene points. }
\label{fig:depth-odo-test-far}
\end{figure*}

\begin{figure*}[!htb]
\centering
\setlength{\tabcolsep}{0.05pt}
\begin{tabular}{cccc}
\includegraphics[width=4cm]{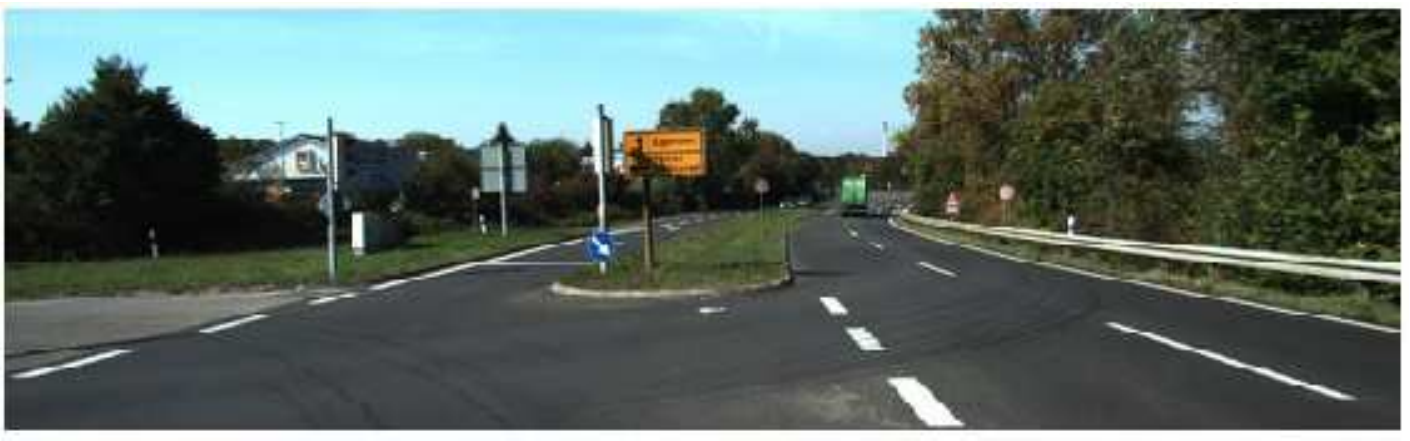} &
\includegraphics[width=4cm]{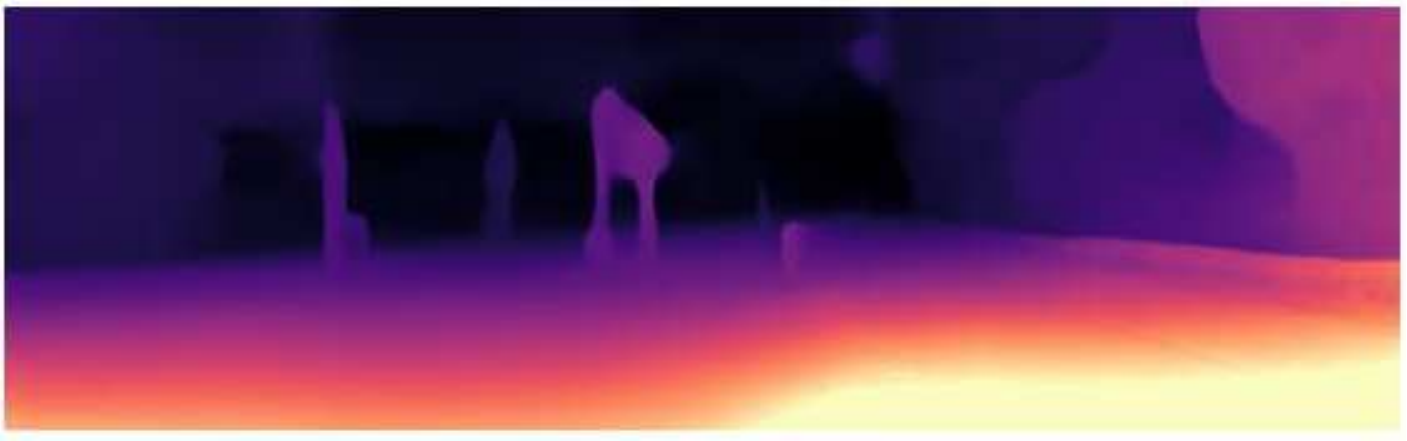} &
\includegraphics[width=4cm]{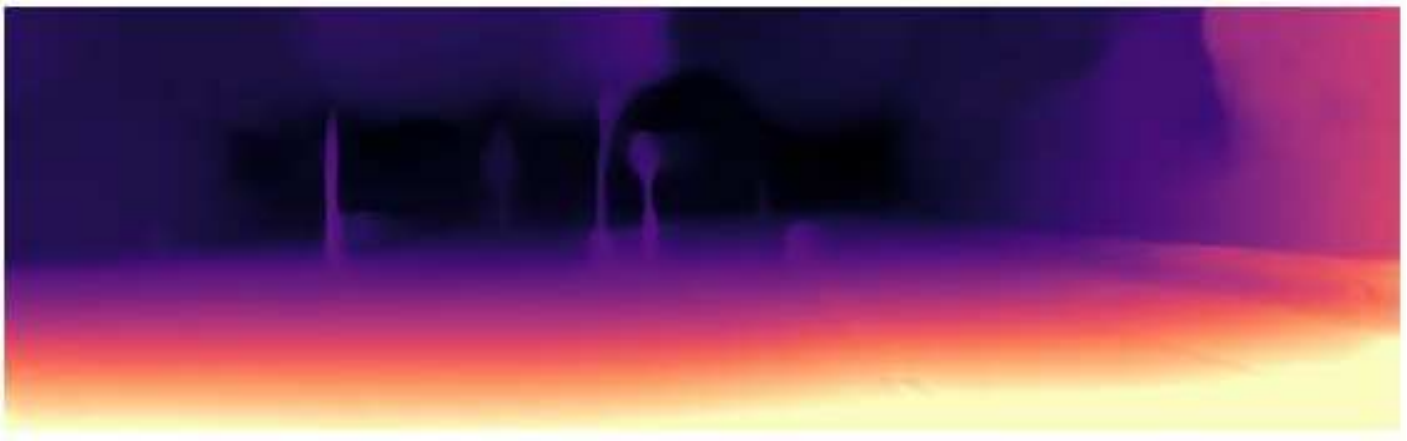} 
\vspace{-0.1cm}\\ 
\includegraphics[width=4cm]{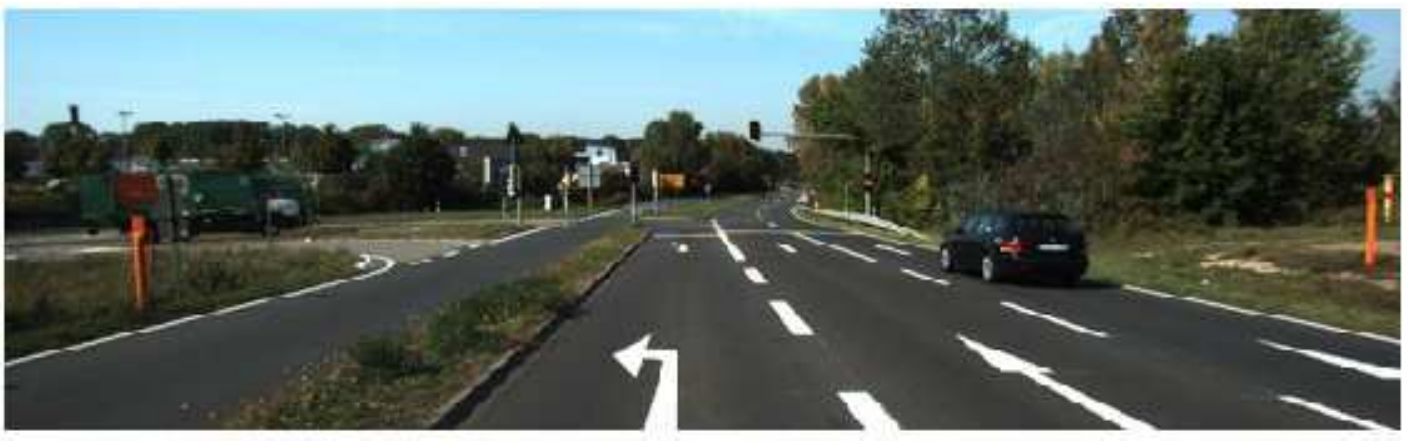} &
\includegraphics[width=4cm]{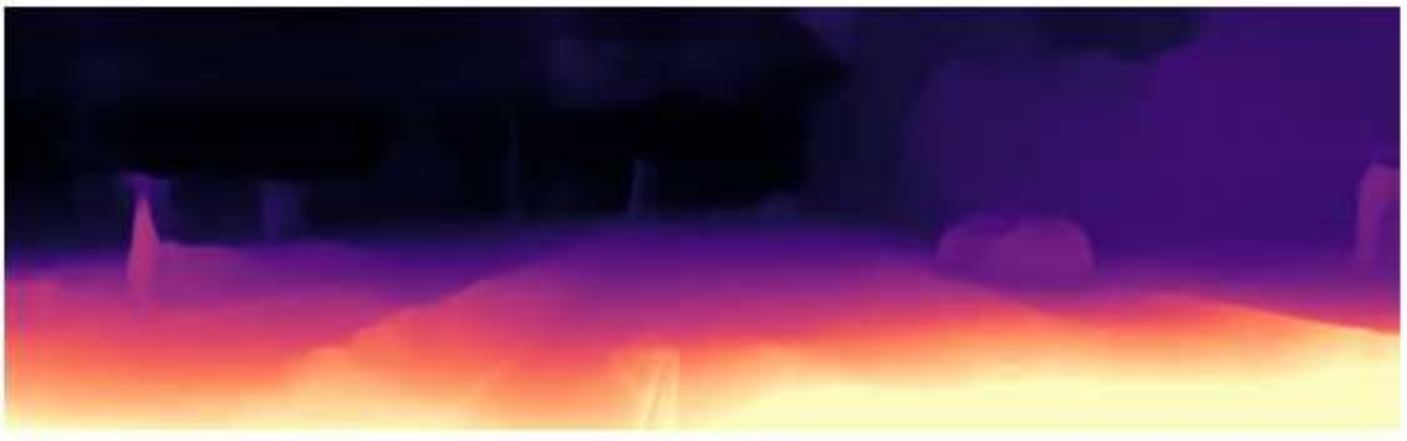} &
\includegraphics[width=4cm]{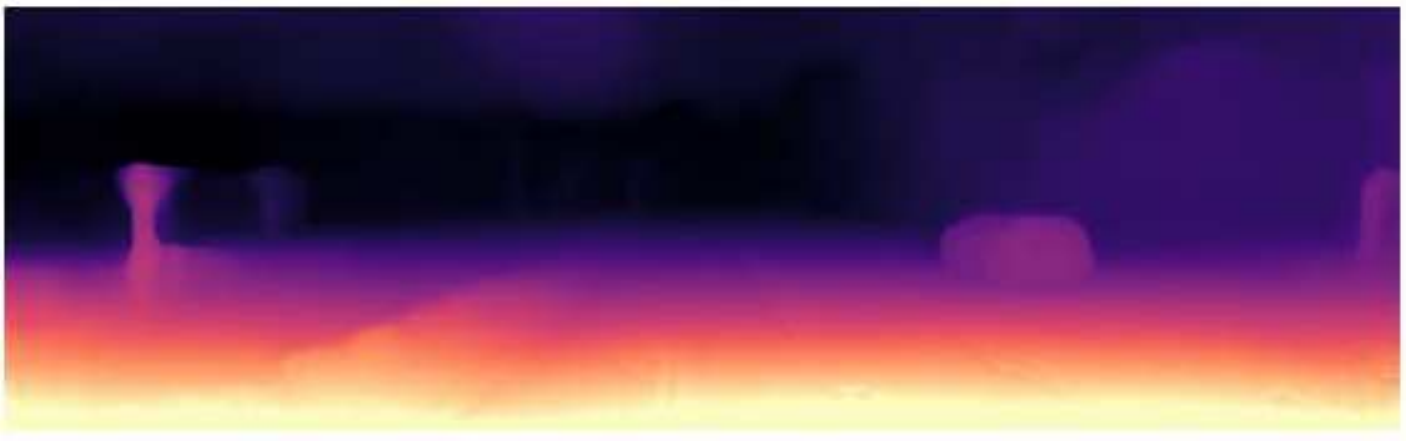} 
\vspace{-0.1cm} \\
\includegraphics[width=4cm]{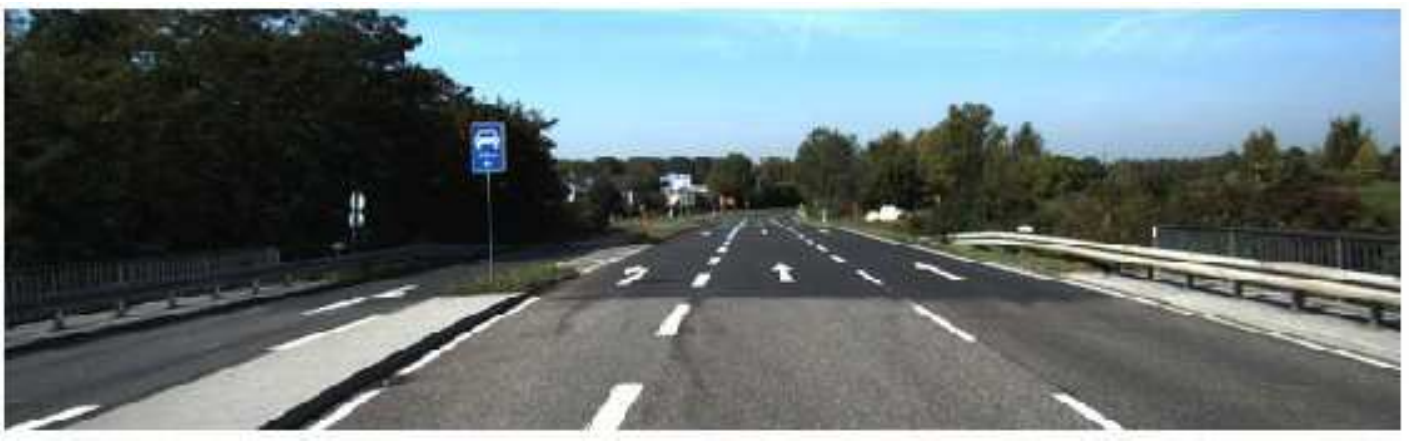} &
\includegraphics[width=4cm]{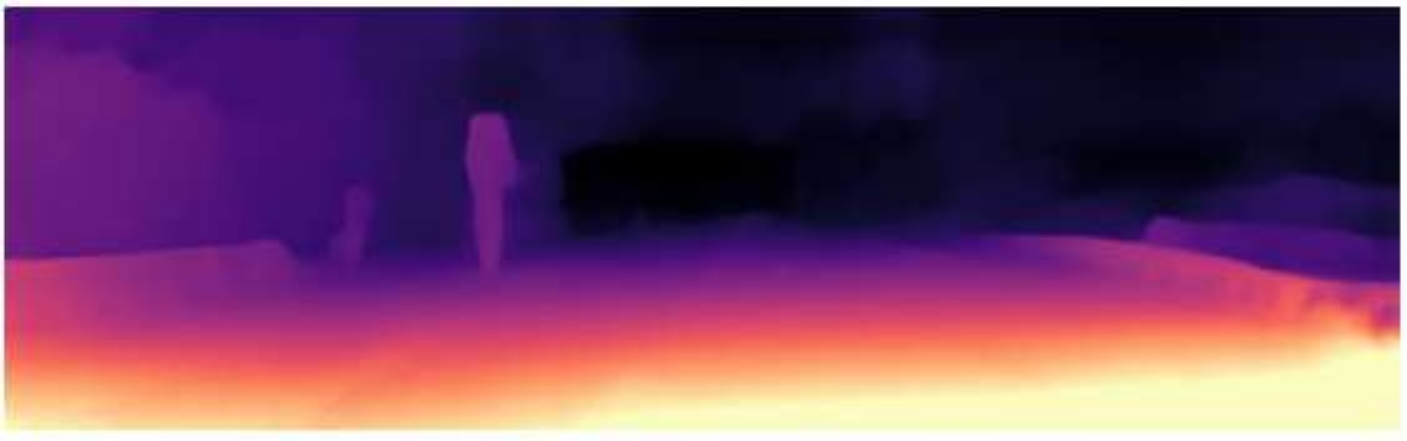} &
\includegraphics[width=4cm]{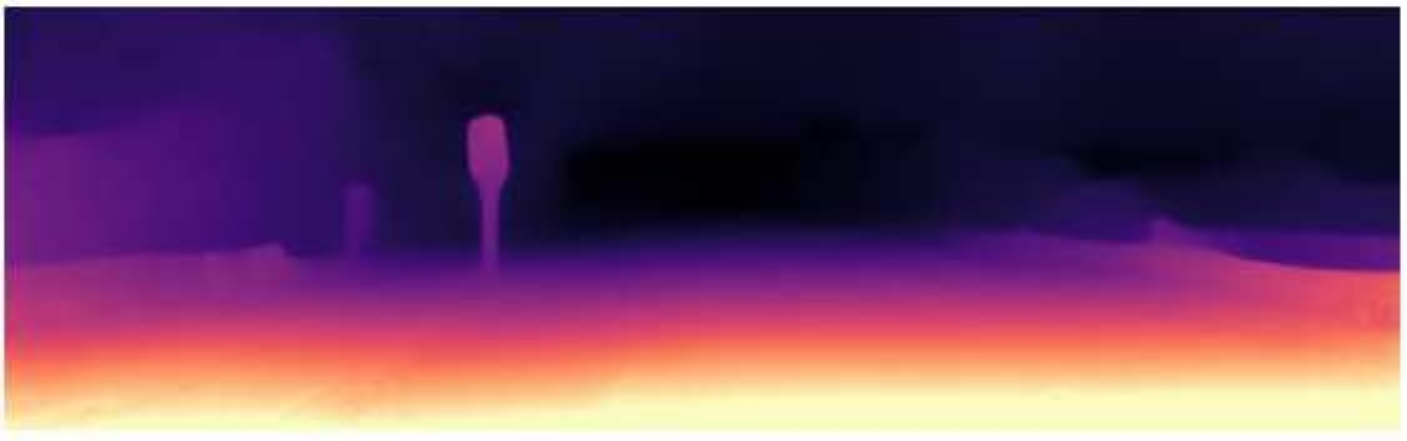}
\vspace{-0.15cm}\\
\includegraphics[width=4cm]{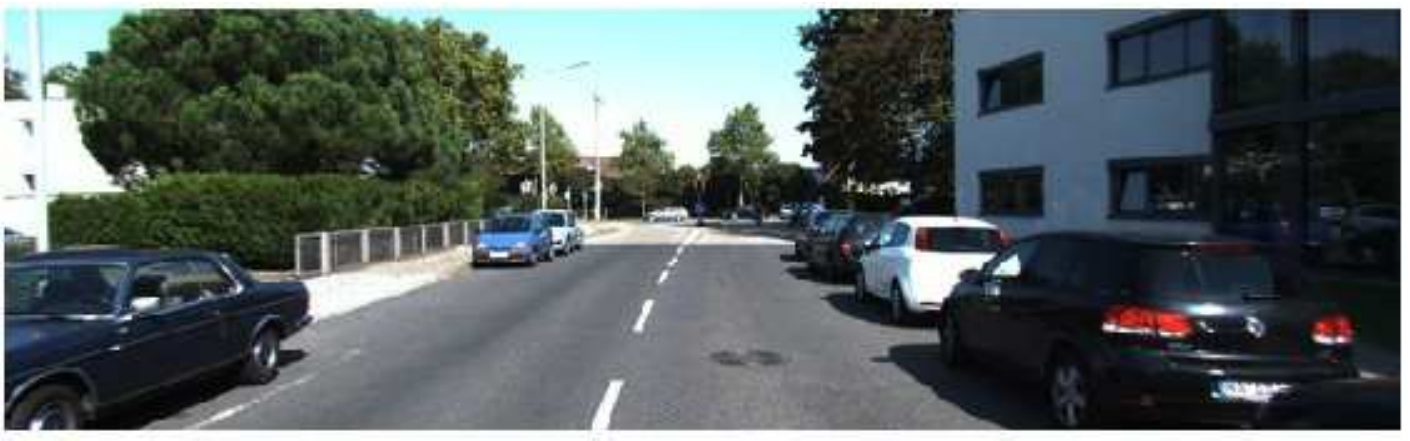} &
\includegraphics[width=4cm]{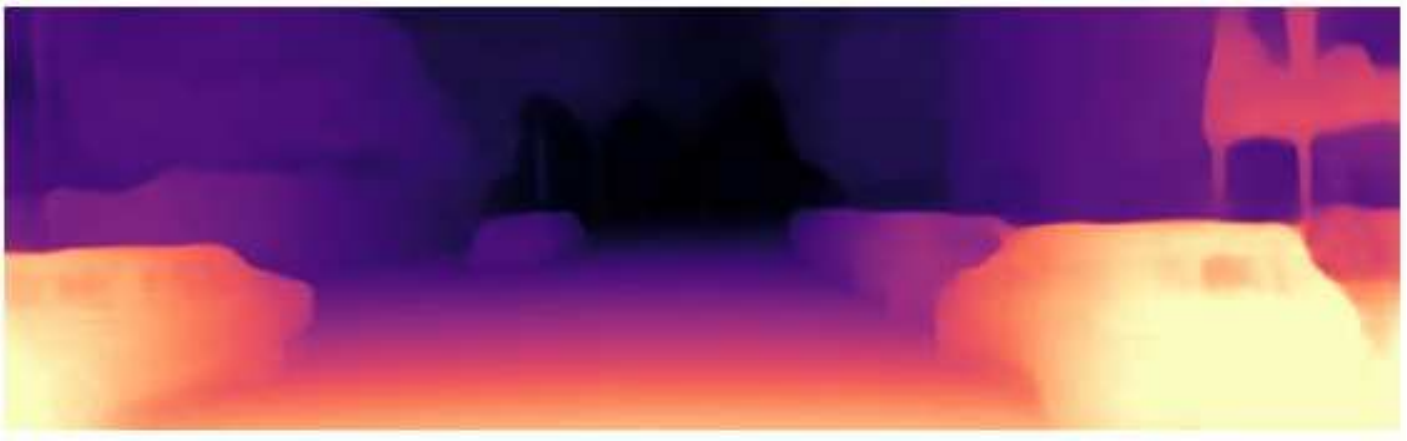} &
\includegraphics[width=4cm]{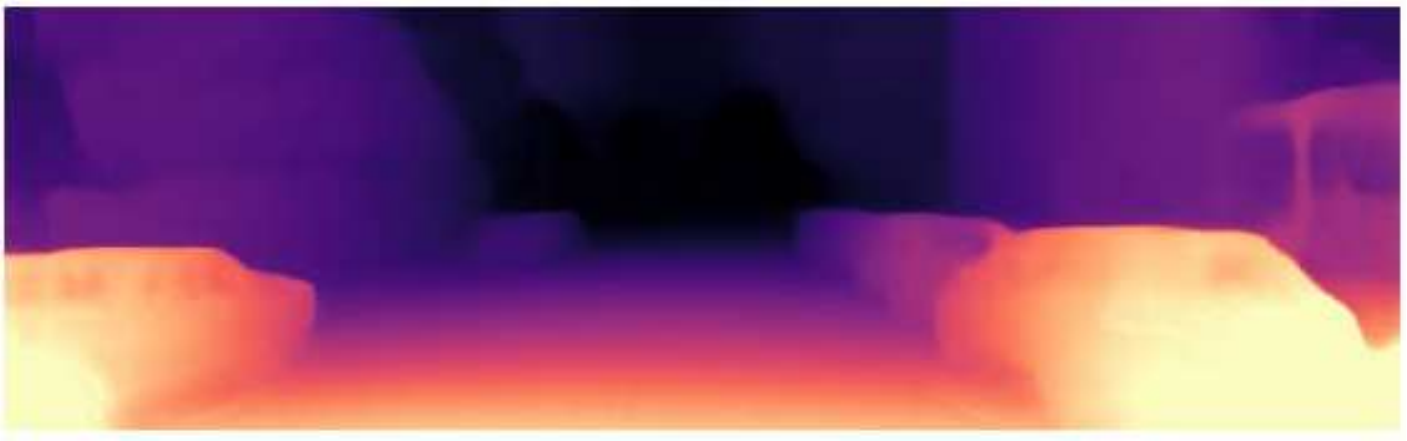}
\vspace{-0.15cm}\\
\includegraphics[width=4cm]{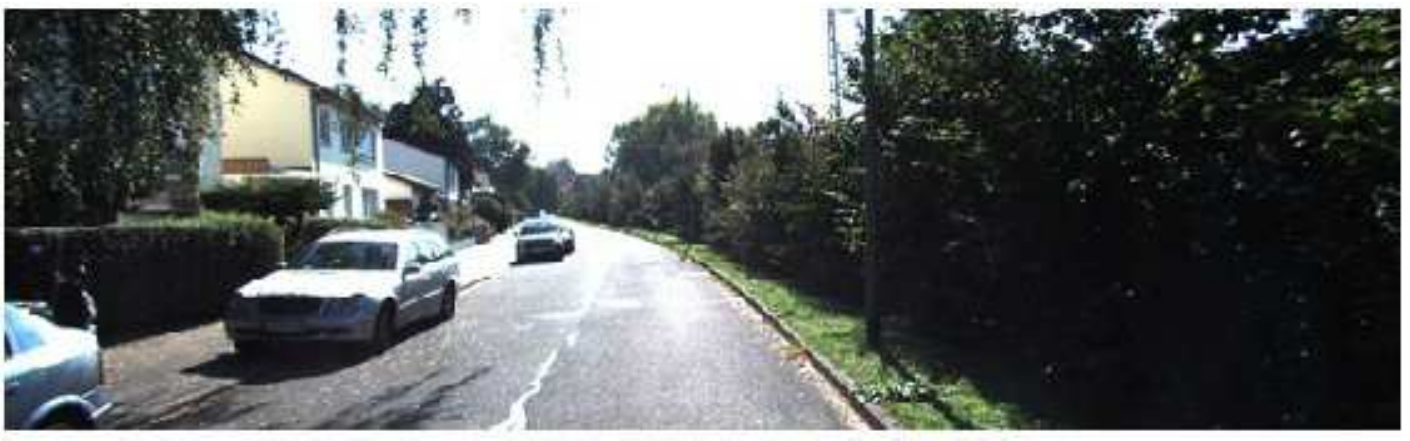} &
\includegraphics[width=4cm]{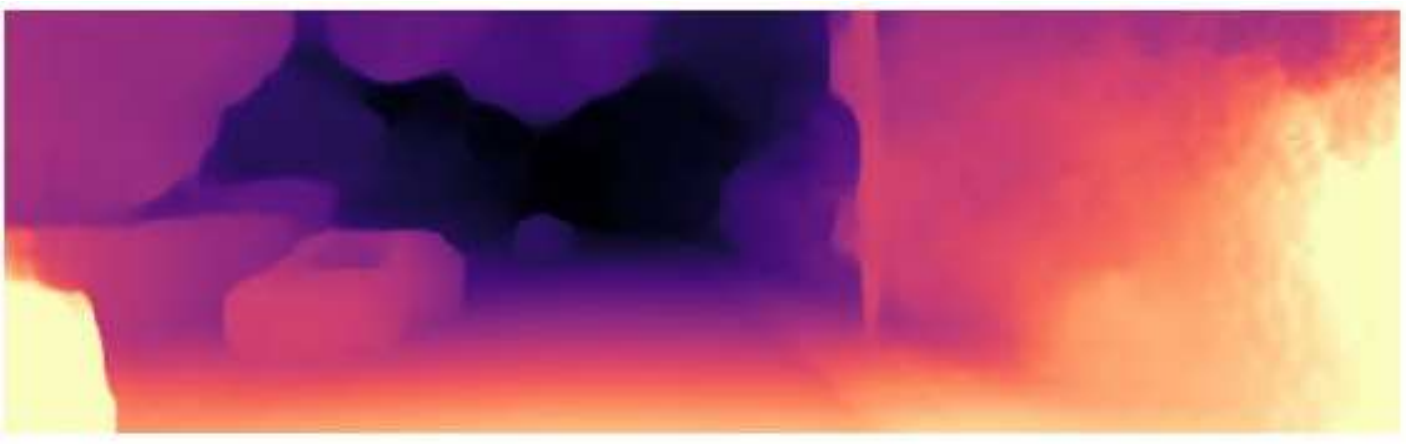} &
\includegraphics[width=4cm]{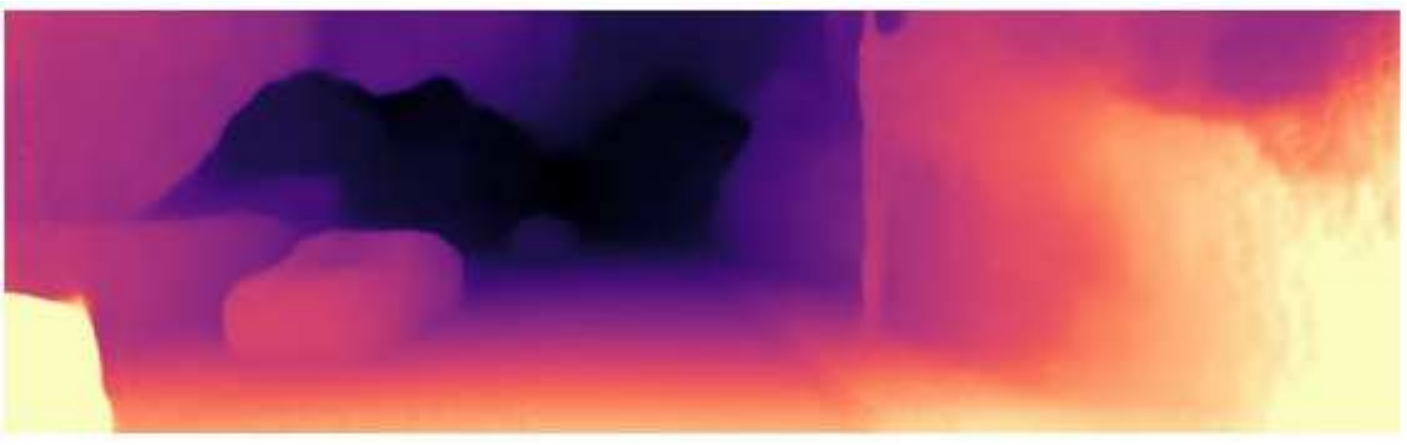}
\vspace{-0.15cm}\\
\includegraphics[width=4cm]{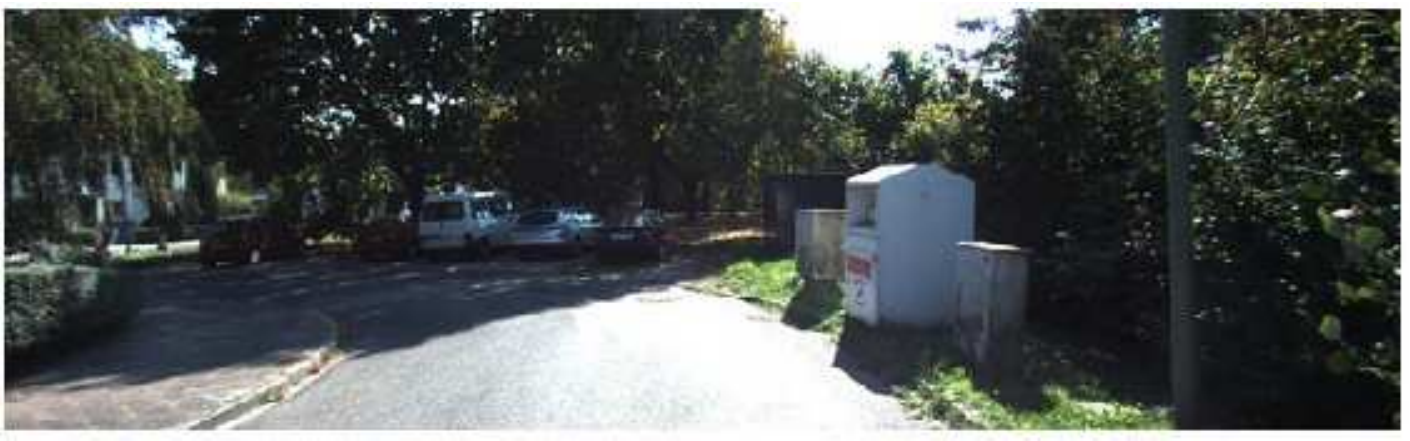} &
\includegraphics[width=4cm]{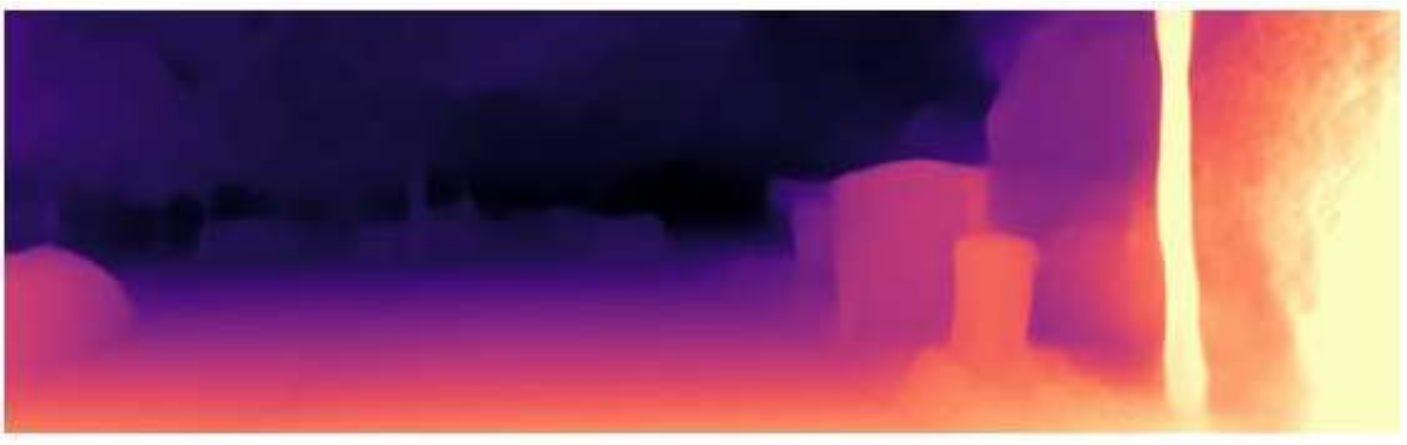} &
\includegraphics[width=4cm]{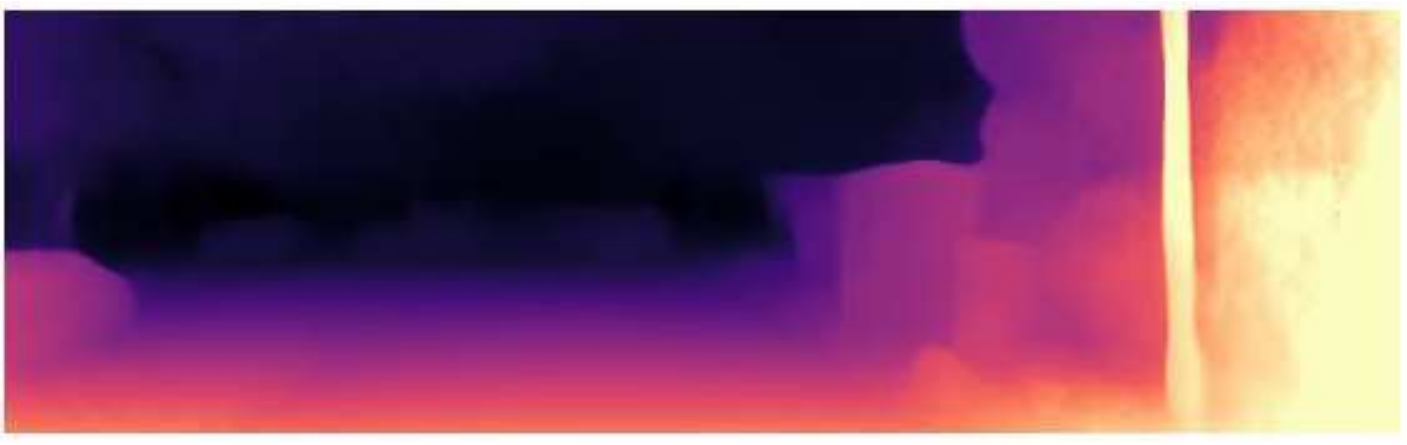}
\vspace{-0.15cm}\\
\includegraphics[width=4cm]{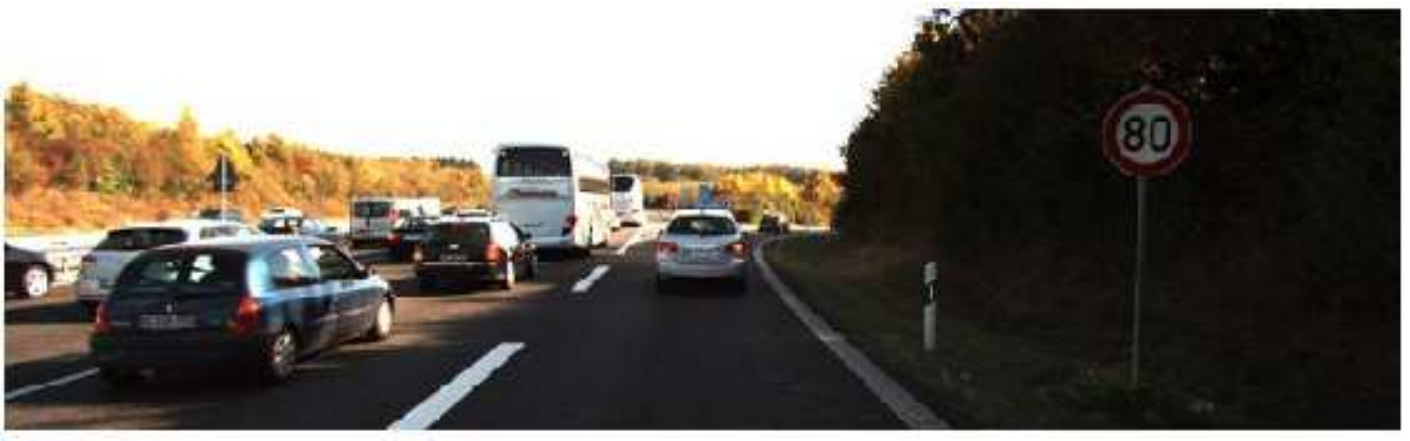} &
\includegraphics[width=4cm]{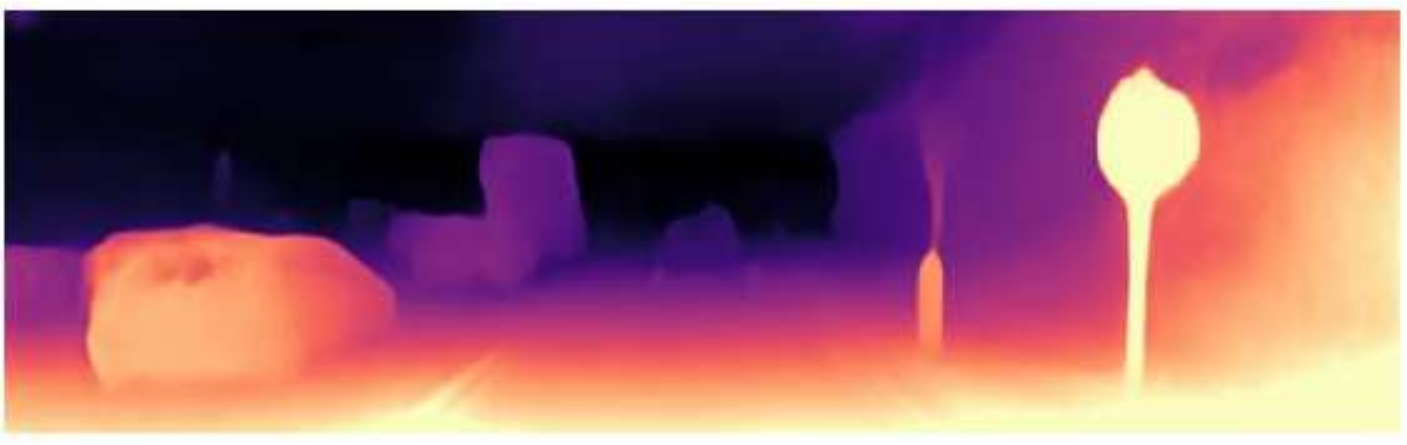} &
\includegraphics[width=4cm]{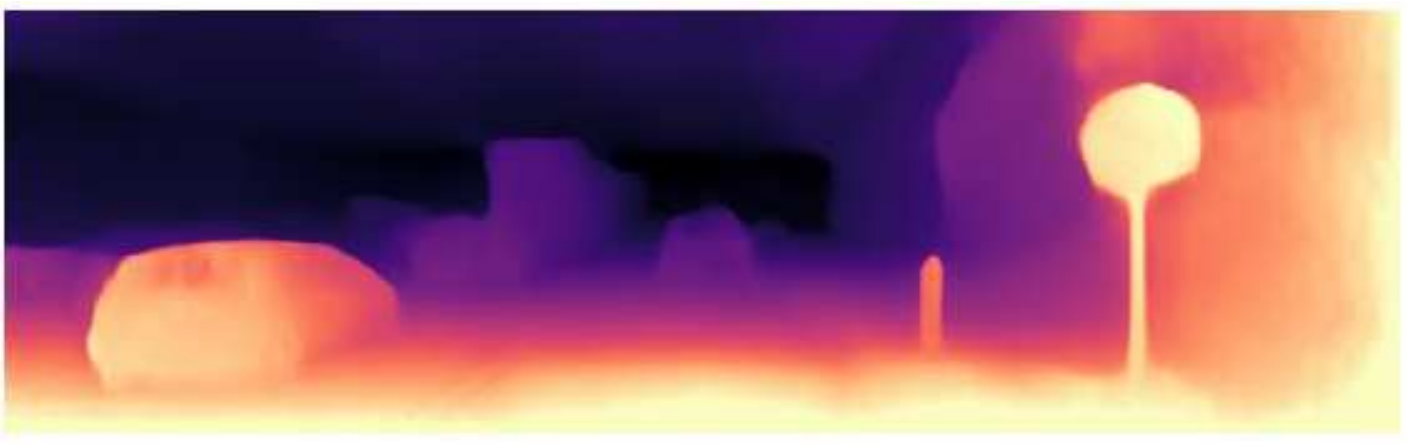}
\vspace{-0.15cm}\\
\includegraphics[width=4cm]{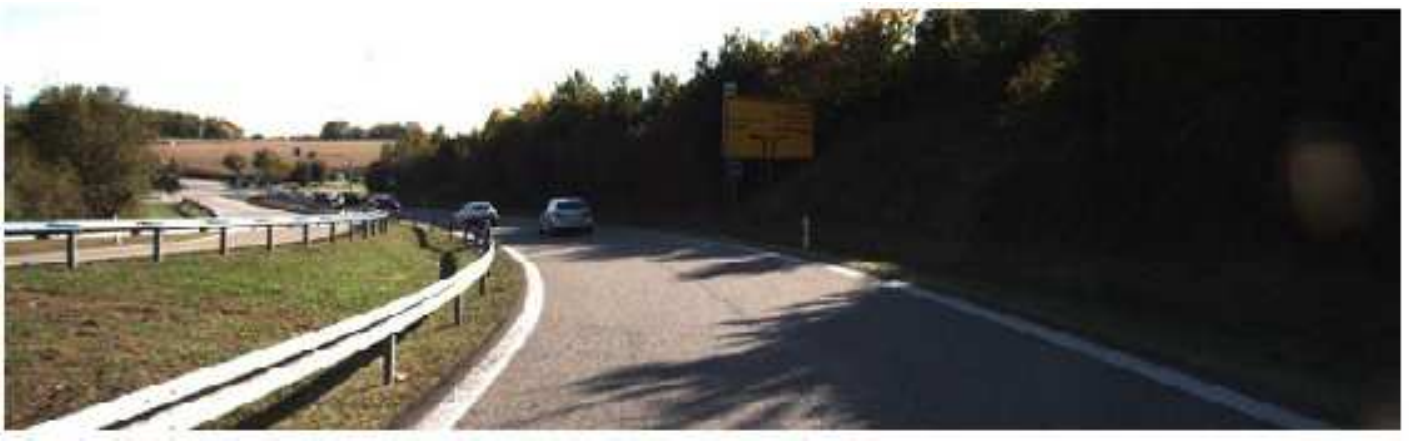} &
\includegraphics[width=4cm]{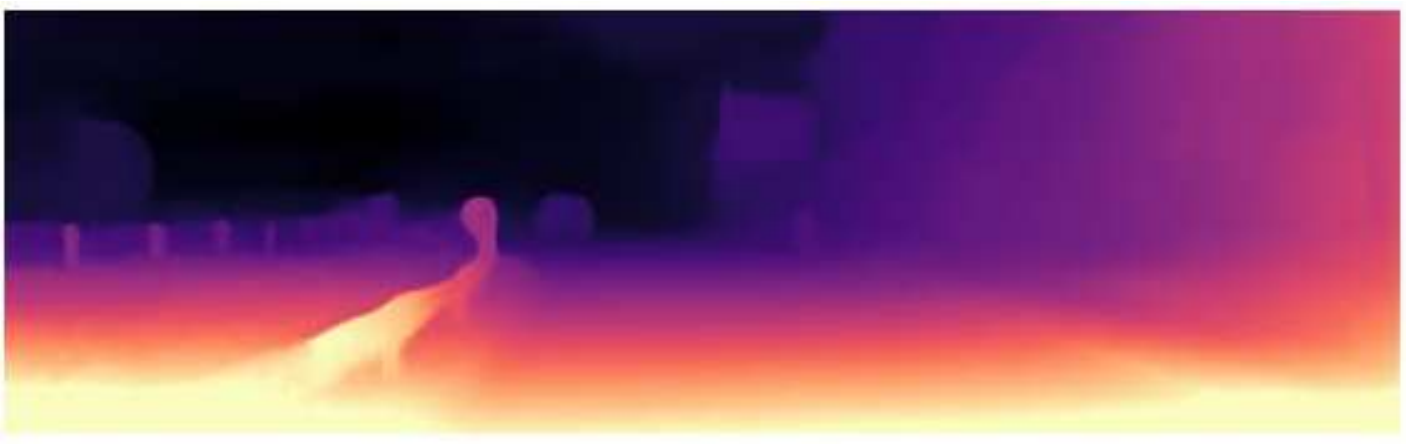} &
\includegraphics[width=4cm]{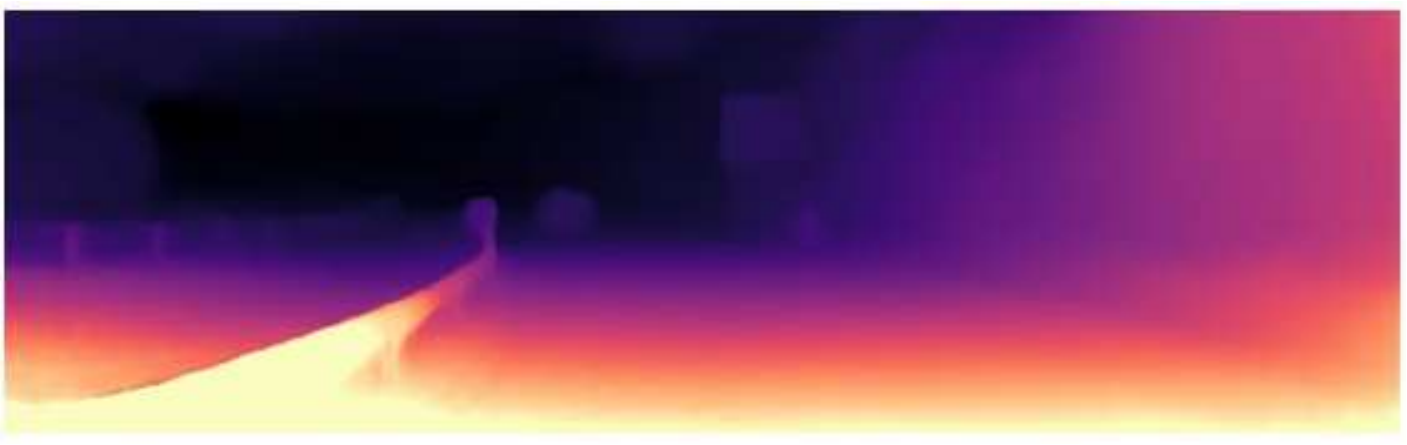}
\vspace{-0.15cm}\\
{\scriptsize RGB}   & {\scriptsize MonoDepth2-M} & {\scriptsize pRGBD-Refined} \\
\end{tabular}
\vspace{-0.3cm}
\caption{Qualitative depth evaluation results on KITTI Raw Eigen split test set. MonoDepth2-M: MonoDepth2 trained using monocular images,  }
\label{fig:depth-raw-test-mix}
\end{figure*}

Fig. \ref{fig:depth-odo-test-far} shows some visual improvements in depth predictions of farther scene points.  Fig. \ref{fig:depth-raw-test-mix} shows some additional qualitative results, where pRGBD-Refined shows visible improvements at occlusion boundaries and thin objects.
The reason for the improvements is the aggregated cues from multiple views with wider baselines (e.g., our depth transfer and depth consistency losses) lead to more well-posed depth recovery.

\subsection{Additional Pose Refinement Qualitative Results}
\label{sec:add-pose_qual}
\begin{figure*}[!htb]
\centering
\setlength{\tabcolsep}{2pt}
\begin{tabular}{ccc}
\includegraphics[width=3.3cm]{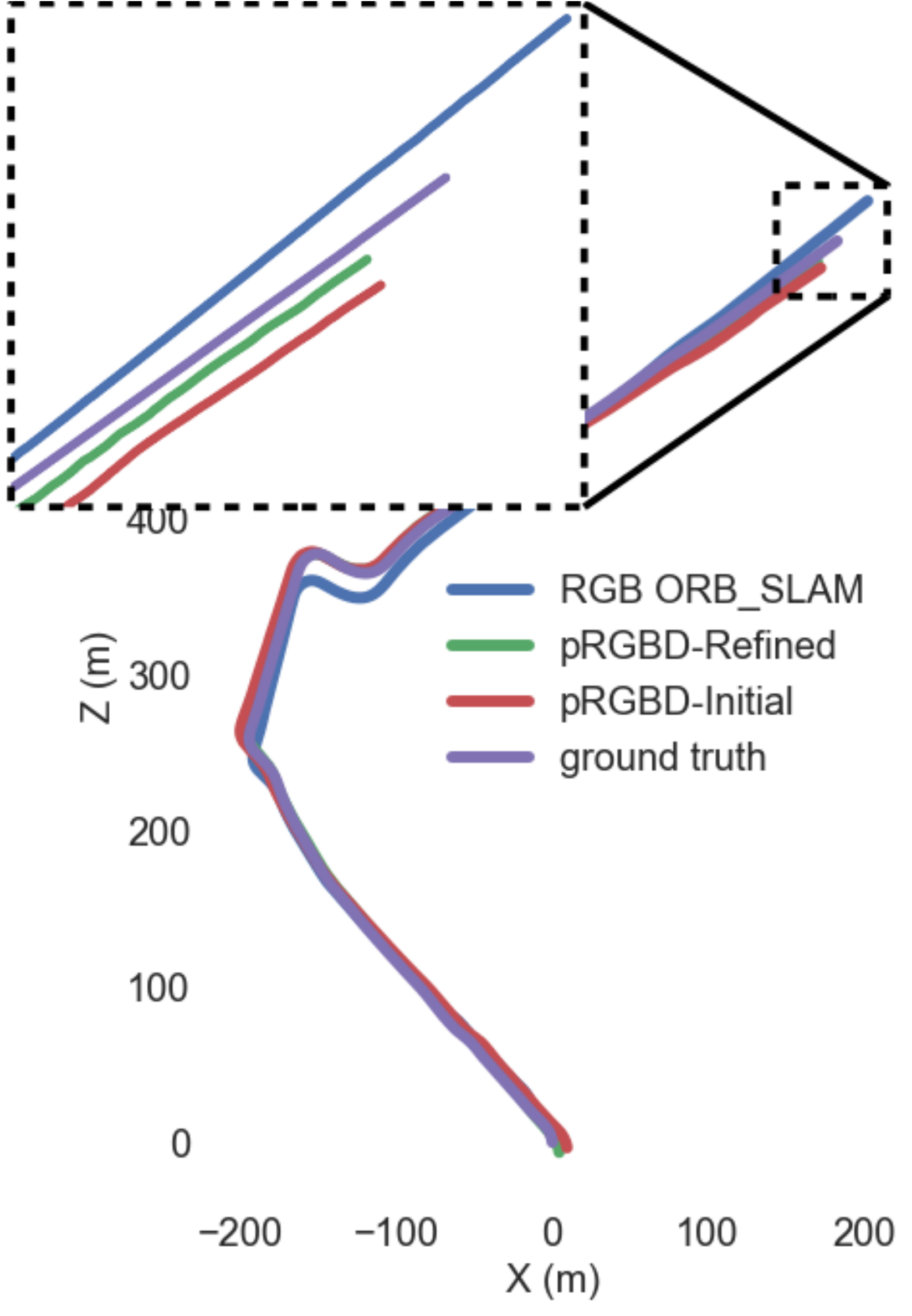} &
\includegraphics[width=4cm]{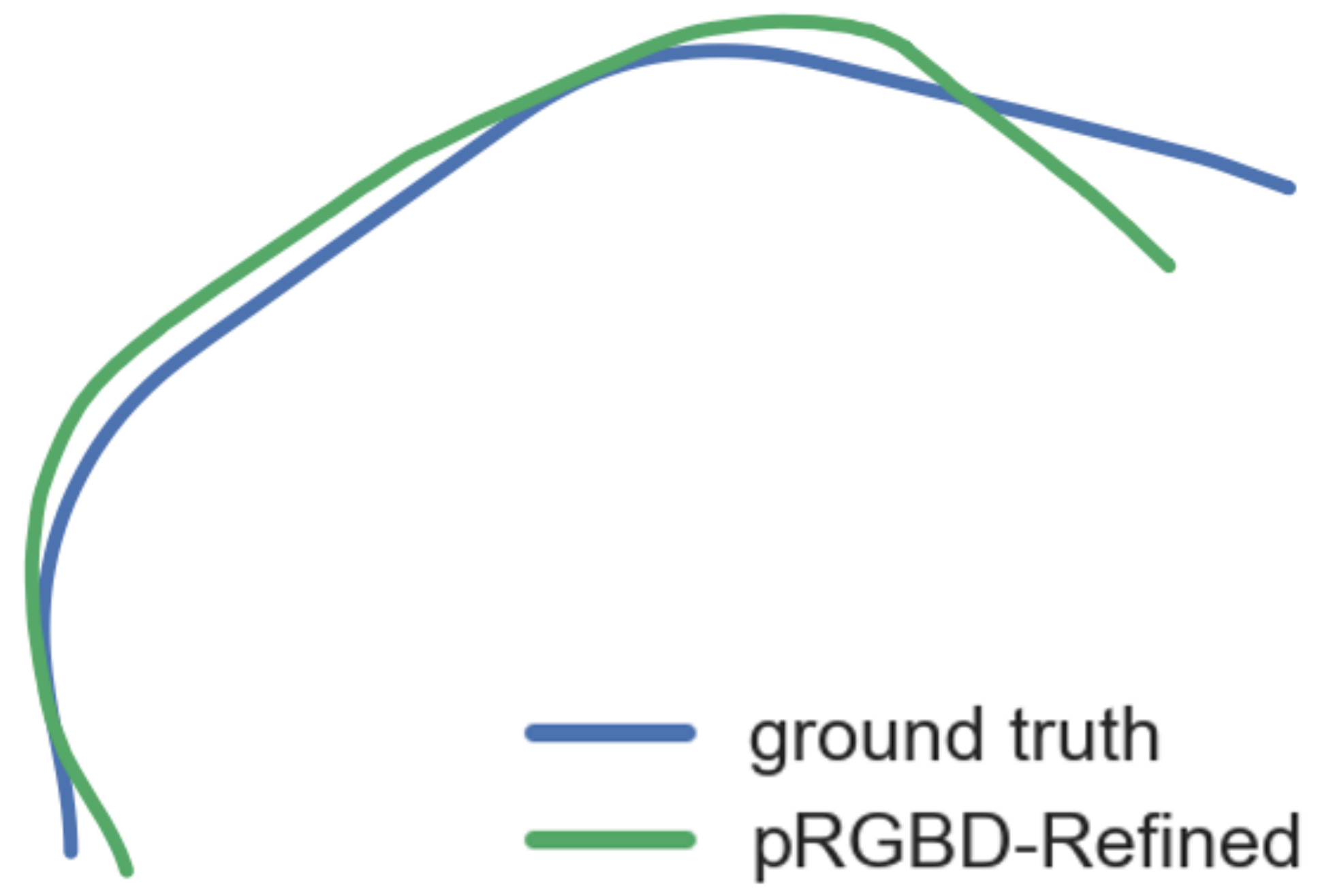} &
\includegraphics[width=4cm]{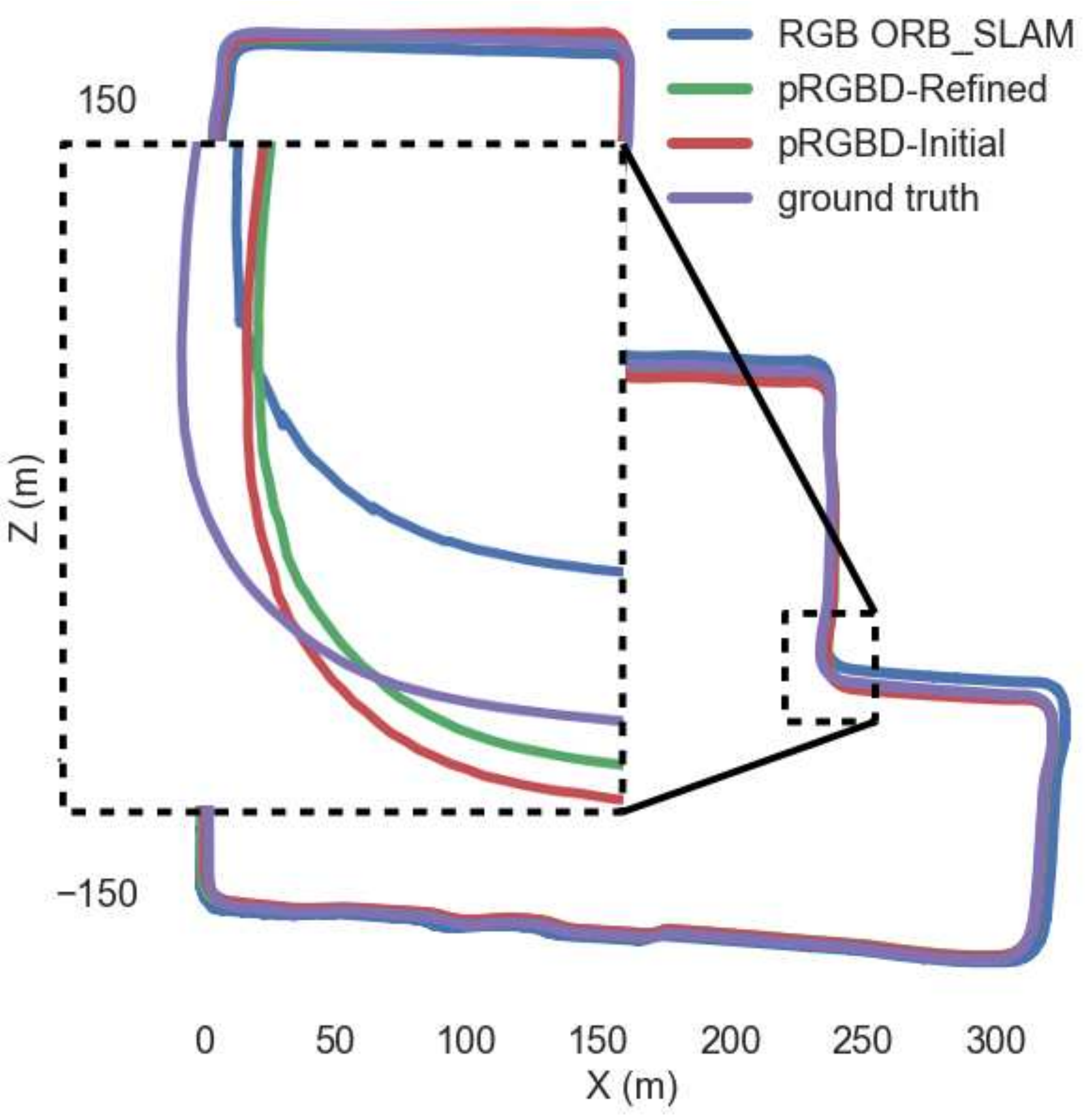}  \\
(a) Seq 11 & (b) Seq 12 & (a) Seq 15 \\
\end{tabular}
\caption{Qualitative pose evaluation results on KITTI Odometry sequences. Note that both RGB ORB-SLAM and pRGBD-Initial fail in (b).}
\label{fig:add-pose-odometry-test-qual}
\end{figure*}

Some additional pose refinement qualitative results are shown in Fig. \ref{fig:add-pose-odometry-test-qual}. In all the three sequences our pRGBD-Refined aligned well with the ground-truth trajectory. Note that both RGB ORB-SLAM and our pRGBD-Initial fail on sequence 12, whereas our pRGBD-Refined succeeds, showing the enhanced robustness by our self-improving framework. 

\subsection{Demo Videos}
\label{sec:video}
We include example videos on sequences 11 and 19 of KITTI Odometry (i.e., \textbf{http://tiny.cc/pRGBD\_KITTI\_11}  and  \textbf{http://tiny.cc/pRGBD\_KITTI\_19} , respectively) and sequence {fr3/large\_cabinet\_ validation} of TUM RGB-D (i.e., \textbf{http://tiny.cc/pRGBD\_TUM\_LCV}). In particular, we illustrate the improvements in depth prediction at frames 140, 352 of \textit{pRGBD\_KITTI\_11}, frames 1652, 3248, 3529 of \textit{pRGBD\_KITTI\_19}, and frames 153, 678 of \textit{pRGBD\_TUM\_LCV}. 
In addition, we highlight the failure of RGB ORB-SLAM at frame 2985 of \textit{pRGBD\_KITTI\_19}.



\clearpage
\bibliographystyle{splncs04}
\bibliography{reference.bib}

\end{document}